\documentclass[10pt]{article}
\usepackage[accepted]{rlj}

\usepackage{basic}
\usepackage{math}
\disablehyphenation

\setlist[itemize]{nosep, leftmargin=4mm}
\setlist[enumerate]{nosep}
\renewcommand{\labelitemi}{\rule[.2ex]{.8ex}{.8ex}}

\usepackage[titles]{tocloft}
\usepackage[noend]{algpseudocode}

\tikzstyle{state}=[draw, circle, minimum size=.8cm, line width=1pt]
\tikzstyle{action}=[->, line width=1pt]

\newcommand{\List}[1]{\bracks*{#1}}
\newcommand{\emptylist}{\List{\;}}

\newcommand{\States}{S}
\newcommand{\Actions}{A}
\newcommand{\Rewards}{R}
\newcommand{\Statistics}{T}
\newcommand{\StateActions}{{\States \times \Actions}}

\newcommand{\eR}{{\overline{\R}}} 

\newcommand{\policy}{\pi}
\newcommand{\transition}{\mathrm{p}}
\newcommand{\reward}{\mathrm{r}}
\newcommand{\statistic}{\tau}
\newcommand{\advantage}{\alpha}

\newcommand{\terminal}{\omega}
\newcommand{\horizon}{\Omega}

\DeclareMathOperator{\step}{step}
\DeclareMathOperator{\gen}{gen}

\newcommand{\update}{\mathbin{\triangleright}}
\DeclareMathOperator{\post}{post}
\DeclareMathOperator*{\agg}{agg}

\DeclareMathOperator{\dadd}{+_\gamma}
\DeclareMathOperator{\dsum}{{\fsum}_\gamma}
\DeclareMathOperator{\dmax}{{\max}_\gamma}
\DeclareMathOperator{\dmin}{{\min}_\gamma}

\newcommand{\svalue}{\mathrm{v}} 
\newcommand{\qvalue}{\mathrm{q}} 
\newcommand{\Bellman}{\mathcal{B}} 

\newcommand{\pair}{{\angles{\id_\States, \policy}}}
\newcommand{\ptro}{{\policy, \transition, \reward, \terminal}}
\newcommand{\aggiu}{{\agg}_{\init, \update}}

\newcommand{\scases}[2]{
\begin{cases}
#1 & s \in \States_\terminal
\\
#2 & s \notin \States_\terminal
\end{cases}
}

\newcommand{\Probability}[1]{\mathbb{P}#1}

\newcommand{\actorP}{\theta}
\newcommand{\criticP}{\phi}

\newcommand{\modified}[1]{{\color{\yellow} #1}}

\theoremstyle{plain}
\newtheorem{theorem}{Theorem}[section]

\newtheorem{lemma}[theorem]{Lemma}
\newtheorem{corollary}[theorem]{Corollary}
\theoremstyle{definition}
\newtheorem{definition}[theorem]{Definition}

\theoremstyle{remark}
\newtheorem{remark}{Remark}

\title{Recursive Reward Aggregation}
\setrunningtitle{Recursive Reward Aggregation}

\author{%
Yuting Tang\textsuperscript{1, 2}
\quad
Yivan Zhang\textsuperscript{1, 2}
\quad
Johannes Ackermann\textsuperscript{1, 2}
\\
Yu-Jie Zhang\textsuperscript{2}
\quad
Soichiro Nishimori\textsuperscript{1, 2}
\quad
Masashi Sugiyama\textsuperscript{2, 1}
}

\emails{%
tang@ms.k.u-tokyo.ac.jp
\quad
yivan.zhang@k.u-tokyo.ac.jp
}

\affiliations{
\textsuperscript{1}%
\textbf{The University of Tokyo, Japan}
\hspace{2em}
\textsuperscript{2}%
\textbf{RIKEN AIP, Japan}
}

\newcommand{\codelink}{\url{https://github.com/Tang-Yuting/recursive-reward-aggregation}}

\summary{
In reinforcement learning (RL), agents typically learn desired behaviors by maximizing the (discounted) sum of rewards, making the design of reward functions crucial for aligning the agent behavior with specific objectives.
However, because rewards often carry intrinsic meanings tied to the task, modifying them can be challenging and may introduce complex trade-offs in real-world scenarios.
In this work, rather than modifying the reward function itself, we propose leveraging different reward aggregation functions to achieve different behaviors.
By introducing an algebraic perspective on Markov decision processes (MDPs), we show that the Bellman equations naturally emerge from the recursive generation and aggregation of rewards.
This perspective enables the generalization of the standard discounted sum to other recursive aggregation functions, such as discounted max and Sharpe ratio.
We empirically evaluate our approach across diverse environments using value-based and actor-critic algorithms, demonstrating its effectiveness in optimizing a wide range of objectives.
Furthermore, we apply our method to a real-world portfolio optimization task, showcasing its potential for practical deployment in decision-making applications where objectives cannot easily be expressed as the discounted sum of rewards.
}

\contribution{
We provide an algebraic perspective on the recursive structure of MDPs based on fusion.
}
{
The algebra of recursive functions \citep{meijer1991functional, bird1997algebra, hutton1999tutorial} is a well-studied topic in functional programming.
The fusion technique, explored by \citet{hinze2010theory}, has been applied to dynamic programming \citep{bellman1966dynamic, de1994categories, bertsekas2022abstract}.
In the context of RL, the recursive structure of the discounted sum of rewards was studied by \citet{hedges2022value}.
Our diagrammatic representation of recursive reward generation and aggregation processes is inspired by \citet{gavranovic2022space}.
}

\contribution{
We generalize the Bellman equations and Bellman operators for the standard discounted sum to other recursive aggregation functions, providing greater flexibility in goal specification.
}
{
The problem of alternative reward aggregations is not entirely new.
Prior works have explored objectives such as optimizing the maximum \citep{quah2006maximum, gottipati2020maximum, veviurko2024max}, minimum \citep{cui2023reinforcement}, top-k \citep{wang2020planning}, and Sharpe ratio \citep{nagele2024tackling} of rewards.
Specifically, the method proposed by \citet{cui2023reinforcement} is a special case of our framework, where the recursive structure is on the original reward space, and the update function is order-preserving.
}

\contribution{
We extend existing RL algorithms by incorporating the generalized Bellman operators and empirically demonstrate their effectiveness across various tasks.
}
{
While our method modifies the Bellman operators within the base RL algorithms, the fundamental structures of Q-learning \citep{watkins1989learning, watkins1992q}, PPO \citep{schulman2017proximal}, and TD3 \citep{fujimoto2018addressing} remain unchanged.
}

\begin{document}

\setlength{\abovedisplayskip}{\lineskip}
\setlength{\belowdisplayskip}{\lineskip}

\makeCover
\maketitle


\begin{abstract}
In reinforcement learning (RL), aligning agent behavior with specific objectives typically requires careful design of the reward function, which can be challenging when the desired objectives are complex.
In this work, we propose an alternative approach for flexible behavior alignment that eliminates the need to modify the reward function by selecting appropriate reward aggregation functions.
By introducing an algebraic perspective on Markov decision processes (MDPs), we show that the Bellman equations naturally emerge from the recursive generation and aggregation of rewards, allowing for the generalization of the standard discounted sum to other recursive aggregations, such as discounted max and Sharpe ratio.
Our approach applies to both deterministic and stochastic settings and integrates seamlessly with value-based and actor-critic algorithms.
Experimental results demonstrate that our approach effectively optimizes diverse objectives, highlighting its versatility and potential for real-world applications.

\end{abstract}

\section{Introduction}

Reinforcement learning (RL) formalizes sequential decision-making as interaction between an agent and an environment modeled by a Markov decision process (MDP).
In standard RL, the objective is to \emph{maximize the discounted sum of rewards} obtained through interaction \citep{sutton1998introduction, bowling2023settling}.
This formulation has been widely adopted across various domains, including games \citep{mnih2015human,silver2018general, guss2019minerl}, autonomous driving \citep{kiran2021deep}, and stock trading \citep{wu2020adaptive, kabbani2022deep, liu2024dynamic}.



While the discounted sum is standard in RL, many important objectives cannot be expressed in this form.
For example, in tasks where stability matters, minimizing some measure of \emph{variability} in rewards is as important as maximizing expected returns \citep{sobel1982variance, tamar2012policy}.
In finance, the \emph{Sharpe ratio} \citep{sharpe1966mutual} evaluates risk-adjusted returns by penalizing high volatility, requiring optimization beyond simple returns.
Other objectives include
(i) maximizing the \emph{peak performance} in drug discovery to identify the most effective compounds \citep{quah2006maximum, gottipati2020maximum},
(ii) maximizing the \emph{worst-case outcome} in safety-critical domains like self-driving \citep{wang2020planning} or \emph{bottleneck objective} in network routing \citep{cui2023reinforcement}, or
(iii) maximizing the \emph{average reward} in continuing tasks where future and immediate rewards are equally important \citep{schwartz1993reinforcement, mahadevan1996average}.
These cases call for alternative reward aggregation beyond the discounted sum.



Can we simply \emph{modify the reward function} to accommodate these objectives?
This is a natural idea, and prior work has explored shaping or redesigning the rewards to reflect alternative criteria \citep{moody1998performance, ng1999policy, moody2001learning, nagele2024tackling}.
However, this approach often requires expanding the state space to encode long-term objectives \citep{mannor2011mean, wang2020planning}, which can change the effective optimization landscape.
Moreover, manually redesigning a reward function that induces the desired behavior is notoriously difficult \citep{leike2017ai, hadfield2017off, zhu2020safe}, which can lead to reward hacking or goal misalignment \citep{amodei2016concrete, christiano2017deep, di2022goal, ji2023ai}.



\begin{figure}
\centering
\begin{adjustbox}{scale=.65}
\begin{tikzpicture}[shorten >=2pt, shorten <=2pt]
\node (a) [state] at (0, 0) {};
\node (b) [state] at (1, 1) {};
\node (c) [state] at (3, 1) {};
\node (d) [state] at (4, 0) {};
\node (e) [state] at (2, 0) {};
\node (f) [state] at (2, -1) {};
\draw [action] (-1, 0) -- (a);
\draw [action, \red] (a) -- (b) node [midway, above left] {$1$};
\draw [action, \red] (b) -- (c) node [midway, above] {$3$};
\draw [action, \red] (c) -- (d) node [midway, above right] {$5$};
\draw [action, \blue] (a) -- (e) node [midway, above] {$4$};
\draw [action, \blue] (e) -- (d) node [midway, above] {$4$};
\draw [action, \yellow] (a) -- (f) node [midway, below left] {$0$};
\draw [action, \yellow] (f) -- (d) node [midway, below right] {$6$};
\draw [action] (d) -- (5, 0);
\end{tikzpicture}
\end{adjustbox}
\begin{tabular}{c cccccc}
\toprule
& $\fsum$ $^\uparrow$
& $\mean$ $^\uparrow$
& $\max$ $^\uparrow$
& $\min$ $^\uparrow$
& $\operatorname{top}\mhyphen2$ $^\uparrow$
& $\range$ $_\downarrow$
\\
\midrule
$\mathcolor{\red}{\policy_1}$
& $\underline{9}$ & $3$ & $5$ & $1$ & $3$ & $4$
\\
$\mathcolor{\blue}{\policy_2}$
& $8$ & $\underline{4}$ & $4$ & $\underline{4}$ & $\underline{4}$ & $\underline{0}$
\\
$\mathcolor{\yellow}{\policy_3}$
& $6$ & $3$ & $\underline{6}$ & $0$ & $0$ & $6$
\\
\bottomrule
\end{tabular}
\caption[Different aggregation functions lead to different policy preferences]{%
Illustration of three deterministic policies in a simple environment, shown as colored paths with their rewards on edges.
The table on the right shows the aggregated rewards for each policy, with the \underline{optimal scores} (higher $^\uparrow$ or lower $_\downarrow$) underlined.
We can observe that different aggregation functions lead to different policy preferences.
}
\vspace{-1.5em}
\label{fig:aggregation}
\end{figure}
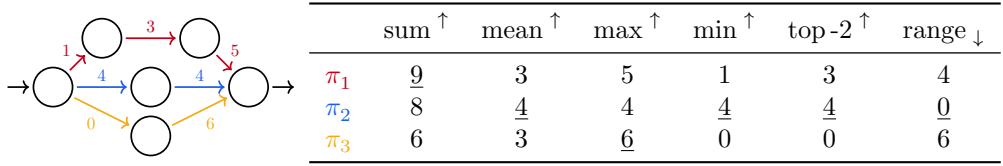

In this work, we propose a simple yet general alternative: \emph{optimize different reward aggregations} directly, while keeping the state space and reward function fixed.
This shifts the focus from what to reward to how to evaluate reward sequences, enabling greater flexibility without increased structural complexity.
Choosing the right aggregation is essential because it defines the optimization objective.
As shown in \cref{fig:aggregation}, even in a toy environment, different aggregation functions such as sum, mean, or max can lead to different policy preferences.
This highlights the need for a general framework that can express and optimize such objectives in a unified way.



\paragraph{Intuition}

In order to optimize reward aggregations directly, our key insight is that many aggregation functions, including the standard discounted sum, can be computed \emph{recursively}, one reward at a time.
For example, the discounted sum aggregation $\dsum$ with a discount factor $\gamma$ satisfies
\begin{equation}
\mathcolor{\red}{\dsum} [r_1, r_2, r_3, \dots]
\defeq
r_1 + \gamma \cdot \mathcolor{\red}{\dsum} [r_2, r_3, \dots]
= r_1 + \gamma r_2 + \gamma^2 r_3 + \dots,
\end{equation}
while the discounted maximum aggregation $\max$ obeys
\begin{equation}
\mathcolor{\red}{\dmax} [r_1, r_2, r_3, \dots]
\defeq
\max(r_1, \gamma \cdot \mathcolor{\red}{\dmax} [r_2, r_3, \dots])
= \max\set{r_1, \gamma r_2, \gamma^2 r_3, \dots}.
\end{equation}
This shared structure suggests a unifying algebraic view: each aggregation \emph{folds} a reward sequence using an update rule and an initial value.
Such recursions naturally induce Bellman-like equations, enabling direct optimization of diverse objectives using standard RL machinery.
Concretely, we use a technique known as \emph{algebra fusion} \citep{hinze2010theory} to derive Bellman-style updates for a wide range of recursive reward aggregations, integrating them into standard RL algorithms without altering the state space or reward function.



\paragraph{Related work}

Several studies have extended Bellman-style updates to optimize non-cumulative objectives.
An early method by \citet{quah2006maximum} defined a value function for expected discounted maximum rewards, but it lacked rigorous justification and incorrectly interchanged expectation with maximum \citep{gottipati2020maximum}, effectively optimizing the discounted maximum of expected rewards.
Later approaches addressed this issue by augmenting the state space with auxiliary variables \citep{veviurko2024max}.
Beyond maximum objectives, \citet{wang2020planning, cui2023reinforcement} analyzed broader classes of objectives, including minimum, harmonic mean, and top-$k$, but their approaches either required symmetry (precluding discounting), were limited to deterministic systems, or implicitly used order-preserving properties.
In contrast, we derive value functions from first principles using algebraic fusion \citep{hinze2010theory}, supporting recursive aggregations involving multi-dimensional statistics (e.g., for range, mean, and variance), while unifying deterministic and stochastic cases.


\paragraph{Contributions}

In this paper, we introduce an \emph{algebraic perspective} on the MDP model, showing that the Bellman equations naturally emerge from the \emph{recursive} generation and aggregation of rewards (\cref{sec:fusion}).
This perspective allows us to generalize the standard discounted sum to other recursive aggregation functions, such as discounted max and Sharpe ratio (\cref{sec:aggregation}), while unifying deterministic and stochastic settings within the same framework (\cref{sec:stochastic}).
We provide theoretical justification for our approach, which enables the optimization of various objectives beyond cumulative rewards while maintaining computational efficiency.
Finally, we validate the effectiveness of our method in both discrete and continuous environments across various recursive reward aggregation functions, showcasing its flexibility and scalability in handling diverse reward structures (\cref{sec:experiments}).\footnote{Code: \codelink.}

\section{An algebraic perspective on Bellman equations}
\label{sec:fusion}
In this section, we introduce the standard MDP model \citep{puterman1994markov} for sequential decision-making problems from an algebraic perspective.
Using a technique known as \emph{fusion} in algebra and functional programming \citep{meijer1991functional, hinze2010theory}, we show that the \emph{Bellman equations} \citep{bellman1966dynamic} naturally arise from the \emph{recursive} generation and aggregation of rewards.
This perspective reveals opportunities for generalizing to alternative reward aggregation functions.

In this section, we focus on the standard \emph{discounted sum} and \emph{deterministic} transitions and policies.
We study other \emph{recursive aggregations} in \cref{sec:aggregation} and \emph{stochastic} transitions and policies in \cref{sec:stochastic}.


\subsection{Preliminaries}

\paragraph{Notation}

In this section, $\States$ is the set of \emph{states}, $\Actions$ is the set of \emph{actions}, and $\Rewards$ is the set of \emph{rewards}, which can be finite or infinite.
The dynamics of the environment is given by a (deterministic) \emph{transition function} $\transition: \StateActions \to \States$.
An agent interacts with the environment by following a (deterministic) \emph{policy} $\policy: \States \to \Actions$ that maps states to actions.
A \emph{reward function} $\reward: \StateActions \to \Rewards$ assigns a reward to each state-action pair.
Furthermore, we assume that there is an \emph{initial state} $s_0 \in \States$ and a subset $\States_\terminal \subset \States$ of \emph{terminal states}, whose indicator function is $\terminal$.
The \emph{horizon} $\horizon$ of the task can be fixed or varying, depending on the terminal condition $\terminal$.

Moreover, $\singleton$ denotes a \emph{singleton} (any set with a single element $*$).
$\List{\Rewards}$ denotes the set of \emph{finite lists} of rewards, defined using the \emph{empty list function} $\nil: \singleton \to \List{\Rewards}$, which represents the empty list $\emptylist$, and the \emph{list constructor function} $\cons: \Rewards \times \List{\Rewards} \to \List{\Rewards}$, which prepends an element to a list.
We have $\cons(r, \emptylist) = [r]$ and $\cons(r_t, [r_{t+1}, \dots, r_\horizon]) = [r_t, r_{t+1}, \dots, r_\horizon]$, which we abbreviate as $r_{t:\horizon}$.


\paragraph{Composite functions}

Let us introduce some composite functions that are useful for defining the recursive generation of states, actions, and rewards.
Given a policy $\policy: \States \to \Actions$, the \emph{pairing function} $\pair: \States \to \StateActions = \anon{s}{s, \policy(s)}$ keeps a copy of the current state $s \in \States$ and outputs the next action $\policy(s) \in \Actions$.%
\footnote{%
For a set $C$, $\id_C: C \to C$ is the \emph{identity function} mapping an element $c \in C$ to itself.
For two functions $f: C \to A$ and $g: C \to B$, their \emph{pairing} $\angles{f, g}: C \to A \times B$ is the unique function that applies these two functions to the same input, mapping an input $c \in C$ to a pair $(f(c), g(c)) \in A \times B$ of outputs.
}%
\footnote{%
We write $\mathrm{name}: \mathrm{domain} \to \mathrm{codomain} = \anon{\mathrm{input}}{\mathrm{output}}$, assigning an \emph{anonymous function} $\anon{\mathrm{input}}{\mathrm{output}}$ to a named, typed function $\mathrm{name}: \mathrm{domain} \to \mathrm{codomain}$, following \citet{petrov2020compositional}.
}
Then, pre-composing this function with the transition function $\transition: \StateActions \to \States$ and the reward function $\reward: \StateActions \to \Rewards$ yields two \emph{policy-dependent} functions as follows.
We use the subscripts $\policy$ to explicitly indicate the dependence on the policy $\policy$:
\begin{itemize}
\item \emph{state transition}
$
\transition_\policy: \States \to \States
\defeq
\transition \compL {\pair}
=
\anon{s}{\transition(s, \policy(s))}
$
and
\item \emph{state reward function}
$
\reward_\policy: \States \to \Rewards
\defeq
\reward \compL {\pair}
=
\anon{s}{\reward(s, \policy(s))}
$.
\end{itemize}


\subsection{Recursive generation of rewards}

Using the state transition $\transition_\policy$ and reward function $\reward_\policy$, we can generate states and rewards step by step:
\begin{equation}
\label{eq:state_step}
\step_\ptro: \States \to \singleton + \Rewards \times \States
\defeq
\anon{s}{\scases{*}{\reward_\policy(s), \transition_\policy(s)}}.
\end{equation}
Let us take a closer look at this step function.
The codomain, $\singleton + \Rewards \times \States$, is the \emph{disjoint union} ($+$) of a \emph{singleton} $\singleton$, representing termination, and the \emph{Cartesian product} $\Rewards \times \States$ of rewards and states.
At each step, the step function either halts by returning the \emph{termination signal} $*$ if the current state $s$ is terminal or continues by returning a pair of the reward $\reward_\policy(s) \in \Rewards$ and the next state $\transition_\policy(s) \in \States$, both determined by the policy $\policy$.

\begin{remark}[Terminal condition]
By incorporating the terminal condition $\terminal$ into the step function, we can describe both \emph{episodic} and \emph{continuing} tasks for any reward aggregation, without relying on a special absorbing state and the unit of the aggregation function, e.g., $0$ for the discounted sum function.
See also \citet[Section~3.4]{sutton1998introduction}.
\end{remark}

Starting from an initial state, by recursively applying this step function and collecting the results, we can obtain a sequence of rewards:
\begin{definition}[Recursive generation]
\label{def:recursive_generation}
Given a policy $\policy$, a transition function $\transition$, a reward function $\reward$, and a terminal condition $\terminal$, a \emph{recursive reward generation function} $\gen_\ptro: \States \to \List{\Rewards}$ is defined as follows:
\begin{equation}
\label{eq:state_generation}
\mathcolor{\blue}{\gen_\ptro}: \States \to \List{\Rewards}
\defeq
\anon{s}{\scases{\emptylist}{
\cons(\reward_\policy(s), \mathcolor{\blue}{\gen_\ptro}(\transition_\policy(s)))
}}.
\end{equation}
\end{definition}


\subsection{Recursive aggregation of rewards}

Given a sequence of rewards, we can aggregate them into a single value using an aggregation function.
In the standard MDP setting, the \emph{discounted sum} $\dsum: \List{\Rewards} \to \Rewards \defeq \anon{r_{1:\horizon}}{\sum_{t=1}^\horizon \gamma^{t-1} r_t}$ of rewards is a standard choice, where $\gamma \in [0, 1]$ is a \emph{discount factor}.

Note that the discounted sum function can be expressed as a recursive function:
\begin{equation}
\label{eq:discounted_sum}
\mathcolor{\red}{\dsum}: \List{\Rewards} \to \Rewards
\defeq
\bracks*{
\begin{array}{@{}ccl@{}}
\emptylist & \mapsto & 0
\\
r_{t:\horizon} & \mapsto & r_t + \gamma \cdot \mathcolor{\red}{\dsum}(r_{t+1:\horizon})
\end{array}
}.
\end{equation}
In other words, the discounted sum function is uniquely defined by two functions: the base case $0 \in \Rewards$ and the recursive case \say{discounted addition} $\dadd: \Rewards \times \Rewards \to \Rewards \defeq \anon{a, b}{a + \gamma \cdot b}$.
This recursive structure has been used, explicitly or implicitly, in prior work on alternative objectives \citep{quah2006maximum, hedges2022value, cui2023reinforcement, veviurko2024max}.
In \cref{sec:aggregation}, we show that many other reward aggregations also admit similar recursive definitions.


\subsection{Bellman equation for the state value function}

We have introduced the recursive generation and aggregation of rewards in a standard MDP model.
The generation function $\gen_\ptro: \States \to \List{\Rewards}$ is the \emph{producer} of rewards, and the discounted sum function $\dsum: \List{\Rewards} \to \Rewards$ is the \emph{consumer} of rewards.
By composing these two recursive functions, we obtain a \emph{state value function} $\svalue_\policy: \States \to \Rewards$, which can also be calculated recursively:
\begin{equation}
\label{eq:state_value_discounted_sum}
\mathcolor{\yellow}{\svalue_\policy}: \States \to \Rewards
\defeq
\dsum \compL \gen_\ptro
=
\anon{s}{\scases{0}{
\reward_\policy(s) + \gamma \cdot \mathcolor{\yellow}{\svalue_\policy}(\transition_\policy(s))
}}.
\end{equation}
This recursive calculation of the state value function $\svalue_\policy: \States \to \Rewards$ is known as the \emph{Bellman equation} \citep{bellman1966dynamic}, which expresses the value of a state $s$ under a policy $\policy$ as the sum of the immediate reward $\reward_\policy(s)$ and the discounted value of the next state $\transition_\policy(s)$.

\begin{remark}[State-action recursion]
We can define the state-action transition/step/generation functions and derive a Bellman equation for the \emph{state-action value function} $\qvalue_\policy: \StateActions \to \Rewards$ in a similar way, which is omitted here for brevity and discussed in \cref{app:state-action}.
\end{remark}

\begin{remark}[Algebra fusion]
For readers familiar with algebra and functional programming, we point out that the Bellman equation emerges as a consequence of the \emph{fusion law} for recursive coalgebras \citetext{\citealp[Section~4]{hinze2010theory}; \citealp[Section~10]{yang2022fantastic}}, shown in the following diagram:%
\footnote{%
For two functions $f: A \to C$ and $g: B \to C$, their \emph{copairing} $\bracks{f, g}: A + B \to C$ is the unique function defined by cases, mapping an input $x \in A + B$ to $f(x)$ if $x \in A$, to $g(x)$ if $x \in B$.
}
\begin{equation}
\label{diag:fusion_state_value_discounted_sum}
\begin{tikzcd}[column sep=0em]
&[.5em]
\singleton + \Rewards \times \States
\arrow[r, "\id_\singleton + \id_\Rewards \times \mathcolor{\blue}{\gen_\ptro}", unique morphism]
\arrow[rr, "\id_\singleton + \id_\Rewards \times \mathcolor{\yellow}{\svalue_\policy}", bend left=60, looseness=.2]
&[9em]
\singleton + \Rewards \times \List{\Rewards}
\arrow[r, "\id_\singleton + \id_\Rewards \times \mathcolor{\red}{\dsum}", unique morphism]
\arrow[d, "\bracks{\nil, \cons}"']
&[7.5em]
\singleton + \Rewards \times \Rewards
\arrow[d, "\bracks{0, \dadd}"']
\\
\singleton
\arrow[r, "s_0"]
&
\States
\arrow[u, "\step_\ptro"]
\arrow[r, "\mathcolor{\blue}{\gen_\ptro}", unique morphism]
\arrow[rr, "\mathcolor{\yellow}{\svalue_\policy}", bend right=60, looseness=.2]
&
\List{\Rewards}
\arrow[r, "\mathcolor{\red}{\dsum}", unique morphism]
&
\Rewards
\end{tikzcd}
\end{equation}
The left square is the recursive definition of the \textcolor{\blue}{generation function} in \cref{eq:state_generation}, and the right square is the recursive definition of the \textcolor{\red}{discounted sum function} in \cref{eq:discounted_sum}.
Consequently, the whole rectangle is the Bellman equation for the \textcolor{\yellow}{state value function} in \cref{eq:state_value_discounted_sum}.
See \cref{app:algebra} for more details.
\end{remark}

\section{Recursive reward aggregation functions}
\label{sec:aggregation}
In this section, we generalize the discounted sum function in \cref{eq:discounted_sum} to other recursive reward aggregation functions that summarize a sequence of rewards into a single value.
Our primary goal is to derive a generalized Bellman equation extending \cref{eq:state_value_discounted_sum} and provide theoretical insights for efficient policy evaluation and optimization with recursive reward aggregation.


\begin{table}
\centering
\vspace{-1em}
\caption{Recursive aggregation functions}
\label{tab:recursive_aggregation_functions}
\begin{adjustbox}{width=\linewidth}
\begin{tabular}{lllll}
\toprule
& definition
& initial value of statistic(s)
& update function
& post-processing
\\
& ${\post} \compL \aggiu: \List{\Rewards} \to \Rewards$
& $\init \in \Statistics$
& $\update: \Rewards \times \Statistics \to \Statistics$
& $\post: \Statistics \to \Rewards$
\\
\midrule
discounted sum
& $r_1 + \gamma r_2 + \dots + \gamma^{t-1} r_t$
& discounted sum $s$: $0 \in \R$
& $\dadd \defeq \anon{r, s}{r + \gamma \cdot s}$
& $\id_\R$
\\
\midrule
discounted min
& $\min\set{r_1, \gamma r_2, \dots, \gamma^{t-1} r_t}$
& discounted min $n$: $\infty \in \eR$
& $\dmin \defeq \anon{r, n}{\min(r, \gamma \cdot n)}$
& $\id_\eR$
\\
\midrule
discounted max
& $\max\set{r_1, \gamma r_2, \dots, \gamma^{t-1} r_t}$
& discounted max $m$: $-\infty \in \eR$
& $\dmax \defeq \anon{r, m}{\max(r, \gamma \cdot m)}$
& $\id_\eR$
\\
\midrule
log-sum-exp
& $\log\parens*{e^{r_1} + e^{r_2} + \dots + e^{r_t}}$
& log-sum-exp $m$: $-\infty \in \eR$
& $\anon{r, m}{\log\parens*{e^r + e^m}}$
& $\id_\eR$
\\
\midrule
range
& $\max(r_{1:t}) - \min(r_{1:t})$
&
\begin{tabular}{@{}l@{}}
max $m$\\ min $n$
\end{tabular}
$\begin{bmatrix}-\infty\\ \infty\end{bmatrix} \in \eR^2$
& $\anon{r, \begin{bmatrix}m\\ n\end{bmatrix}}{\begin{bmatrix}\max(r, m)\\ \min(r, n)\end{bmatrix}}$
& $\anon{\begin{bmatrix}m\\ n\end{bmatrix}}{m - n}$
\\
\midrule
mean
& $\overline{r} \defeq \frac1t \sum_{i=1}^t r_i$
& 
\begin{tabular}{@{}l@{}}
length $n$\\ sum $s$
\end{tabular}
$\begin{bmatrix}0\\ 0\end{bmatrix} \in \begin{bmatrix}\N\\ \R\end{bmatrix}$
& $\anon{r, \begin{bmatrix}n\\ s\end{bmatrix}}{\begin{bmatrix}n + 1\\ s + r\end{bmatrix}}$
& $\anon{\begin{bmatrix}n\\ s\end{bmatrix}}{\frac{s}{n}}$
\\
\addlinespace[2pt]
\cmidrule(lr){3-5}
&&
\begin{tabular}{@{}l@{}}
length $n$\\ mean $m$
\end{tabular}
$\begin{bmatrix}0\\ 0\end{bmatrix} \in \begin{bmatrix}\N\\ \R\end{bmatrix}$
& $\anon{r, \begin{bmatrix}n\\ m\end{bmatrix}}{\begin{bmatrix}n + 1\\ \frac{n \cdot m + r}{n + 1}\end{bmatrix}}$
& $\anon{\begin{bmatrix}n\\ m\end{bmatrix}}{m}$
\\
\midrule
variance
&
$\textstyle \frac1t \sum_{i=1}^t (r_i - \overline{r})^2
= \overline{r^2} - \overline{r}^2$
&
\begin{tabular}{@{}l@{}}
length $n$\\ sum $s$\\ sum square $q$
\end{tabular}
$\begin{bmatrix}0\\ 0\\ 0\end{bmatrix} \in \begin{bmatrix}\N\\ \R\\ \R_{\geq 0}\end{bmatrix}$
& $\anon{r, \begin{bmatrix}n\\ s\\ q\end{bmatrix}}{\begin{bmatrix}n + 1\\ s + r\\ q + r^2\end{bmatrix}}$
& $\anon{\begin{bmatrix}n\\ s\\ q\end{bmatrix}}{\frac{q}{n} - \pfrac{s}{n}^2}$
\\
\addlinespace[2pt]
\cmidrule(lr){3-5}
&&
\begin{tabular}{@{}l@{}}
length $n$\\ mean $m$\\ variance $v$
\end{tabular}
$\begin{bmatrix}0\\ 0\\0\end{bmatrix} \in \begin{bmatrix}\N\\ \R\\ \R_{\geq 0}\end{bmatrix}$
& $\anon{r, \begin{bmatrix}n\\ m\\ v\end{bmatrix}}{\begin{bmatrix}n + 1\\ \frac{n \cdot m + r}{n + 1}\\ v + \frac{n(r-m)^2 - (n+1)v}{(n+1)^2}\end{bmatrix}}$
& $\anon{\begin{bmatrix}n\\ m\\ v\end{bmatrix}}{v}$
\\
\midrule
top-$k$
& $k$-th largest in $r_{1:t}$
&
\begin{tabular}{@{}l}
top-$k$\\ buffer
\end{tabular}
\begin{tabular}{@{}c}
top-$1$\\ top-$2$\\ $\vdots$
\end{tabular}
$\begin{bmatrix}-\infty\\ -\infty\\ \vdots\end{bmatrix} \in \eR^k$
& $\anon{r, b}{\begin{cases}\mathrm{insert}(r, b) & r > \min b \\ b & r \leq \min b \end{cases}}$
& $\anon{b}{\min b}$
\\
\bottomrule
\end{tabular}
\end{adjustbox}
\end{table}


\subsection{Bellman equation for the state statistic function}

First, we observe that many aggregation functions are inherently recursive.
However, the recursive structure does not always operate directly within the original space.
For instance, we can calculate the arithmetic mean by calculating both the sum and the length recursively and then dividing the sum by the length.
Based on this observation, we propose the following definition:
\begin{definition}[Recursive aggregation]
\label{def:recursive_aggregation}
Let $\Statistics$ be a set of \emph{statistics}.
Given an \emph{initial value} $\init \in \Statistics$, an \emph{update function} $\update: \Rewards \times \Statistics \to \Statistics$, and a \emph{post-processing function} $\post: \Statistics \to \Rewards$, a \emph{recursive statistic aggregation function} $\aggiu: \List{\Rewards} \to \Statistics$ of is defined as follows:
\begin{equation}
\mathcolor{\red}{\aggiu}: \List{\Rewards} \to \Statistics
\defeq
\bracks*{
\begin{array}{@{}ccl@{}}
\emptylist & \mapsto & \init
\\
r_{t:\horizon} & \mapsto & r_t \update \mathcolor{\red}{\aggiu}(r_{t+1:\horizon})
\end{array}
},
\end{equation}
and a \emph{recursive reward aggregation function} ${\post} \compL \aggiu: \List{\Rewards} \to \Rewards$ is the composition of this function with the post-processing function $\post: \Statistics \to \Rewards$, shown in the following diagram:
\begin{equation}
\begin{tikzcd}[column sep=0em]
\singleton + \Rewards \times \List{\Rewards}
\arrow[r, "\id_\singleton + \id_\Rewards \times \mathcolor{\red}{\aggiu}", unique morphism]
\arrow[d, "\bracks{\nil, \cons}"']
&[10em]
\singleton + \Rewards \times \Statistics
\arrow[d, "\bracks{\init, \update}"']
\\
\List{\Rewards}
\arrow[r, "\mathcolor{\red}{\aggiu}", unique morphism]
&
\Statistics
\arrow[r, "\post"]
&[2em]
\Rewards
\end{tikzcd}
\end{equation}
\end{definition}

By substituting the discounted sum function with a general recursive reward aggregation function, we can generalize the Bellman equation in \cref{eq:state_value_discounted_sum} as follows:
\begin{restatable}[Bellman equation for the state statistic function]{theorem}{BellmanEquationState}
\label{thm:bellman_equation_state}
Given a recursive reward generation function $\gen_\ptro$ (\cref{def:recursive_generation}) and a recursive statistic aggregation function $\aggiu$ (\cref{def:recursive_aggregation}), their composition, called the state statistic function $\statistic_\policy: \States \to \Statistics$, satisfies
\begin{equation}
\label{eq:state_statistic}
\mathcolor{\yellow}{\statistic_\policy}: \States \to \Statistics
\defeq
{\aggiu} \compL {\gen_\ptro}
=
\anon{s}{\scases{\init}{
\reward_\policy(s) \update \mathcolor{\yellow}{\statistic_\policy}(\transition_\policy(s))
}}.
\end{equation}
\end{restatable}

\begin{definition}[Value function]
The \emph{state value function} $\svalue_\policy: \States \to \Rewards \defeq {\post} \compL \statistic_\policy$ is the composition of the state statistic function $\statistic_\policy: \States \to \Statistics$ with the post-processing function $\post: \Statistics \to \Rewards$.
\end{definition}

\begin{figure}
\centering
\vspace{-1em}
\begin{adjustbox}{scale=0.75}
\begin{tikzpicture}
\coordinate (i) at (0, 0);
\coordinate (r) at (3, 0);
\coordinate (j) at (6, 0);
\coordinate (d) at (8, 0);
\coordinate (k) at (10, 0);

\coordinate (s) at (0, 4);
\coordinate (t) at (0, 0);

\node (Si) at (i |- s) {$\States$};
\node (Ti) at (i |- t) {$\Statistics$};
\node (R) at (r |-, 1) {$\Rewards$};
\node (Sj) at (j |- s) {$\States$};
\node (Tj) at (j |- t) {$\Statistics$};
\node (Sd) at (d |- s) {$\dots$};
\node (Td) at (d |- t) {$\dots$};
\node (Sk) at (k |- s) {$\singleton$};
\node (Tk) at (k |- t) {$\Statistics$};

\node (copyS) [diagonal] at (2, |- s) {};
\node (copySa) [diagonal] at (2, 3) {};
\node (copyA) [diagonal] at (4, 3) {};
\node (copyTi) [diagonal] at (1, 0) {};
\node (copyTj) [diagonal] at (7, 0) {};
\draw (Si) [->] to (copyS);
\draw (copyS) [->] to (copySa);

\node (policy) [morphism, minimum width=.6cm, minimum height=.6cm] at (3, 3) {$\policy$};
\draw (copySa) [->] to (policy);
\draw (policy) [->] to (copyA);

\node (transition) [morphism, minimum width=.6cm, minimum height=.6cm] at (4, |- s) {$\transition$};
\draw (copyS) [->] to (transition);
\draw (copyA) [->] to (transition);
\draw (transition) [->] to (Sj);

\node (reward) [morphism, minimum width=2.6cm, minimum height=.6cm] at (r |-, 2) {$\reward$};
\draw (copySa) [->] to (copySa |- reward.north);
\draw (copyA) [->] to (copyA |- reward.north);
\draw (reward) [->] to (R);

\node (update) [morphism, minimum width=.6cm, minimum height=.6cm] at (r |- t) {$\update$};
\draw (R) [->] to (update);
\draw (Tj) [->] to (update);
\draw (update) [->] to (copyTi);

\draw (Sj) [->] to (Sd);
\draw (Sd) [->] to (Sk);
\draw (Tk) [->] to (Td);
\draw (Td) [->] to (copyTj);

\node (init) [morphism, minimum width=1cm, minimum height=.6cm] at ($(Sk)!0.5!(Tk)$) {$\init$};
\draw (Sk) [->] to (init);
\draw (init) [->] to (Tk);

\node [draw, rectangle, dashed, minimum width=3cm, minimum height=3cm, color=\blue, line width=1.5pt] at (3, 3) {};
\node [draw, rectangle, dashed, minimum width=3cm, minimum height=1cm, color=\red, line width=1.5pt] at (3, 0) {};
\node [draw, rectangle, dashed, minimum width=1.5cm, minimum height=5cm, line width=1.5pt] at (init) {};

\node [right=0 of Si.north east, black!42] {$s_t$};
\node [right=0 of Sj.north east, black!42] {$s_{t+1}$};
\node [right=0 of Ti.north east, black!42] {$\tau_t$};
\node [right=0 of Tj.north east, black!42] {$\tau_{t+1}$};
\node [right=0 of R.north, black!42] {$r_{t+1}$};

\node [right=0 of policy.north east, black!42] {$a_t$};

\node (posti) [morphism, minimum width=1cm, minimum height=.6cm] at (copyTi |-, -1) {$\post$};
\node (Ri) at (i |-, -1) {$\Rewards$};
\draw (copyTi) [->] to (Ti);
\draw (copyTi) [->] to (posti);
\draw (posti) [->] to (Ri);

\node (postj) [morphism, minimum width=1cm, minimum height=.6cm] at (copyTj |-, -1) {$\post$};
\node (Rj) at (j |-, -1) {$\Rewards$};
\draw (copyTj) [->] to (Tj);
\draw (copyTj) [->] to (postj);
\draw (postj) [->] to (Rj);

\node (statistici) [morphism, minimum width=.6cm, minimum height=.6cm] at (i |-, 2) {$\mathcolor{\yellow}{\statistic_\policy}$};
\draw (Si) [->, dashed] to (statistici);
\draw (statistici) [->, dashed] to (Ti);

\node (statisticj) [morphism, minimum width=.6cm, minimum height=.6cm] at (j |-, 2) {$\mathcolor{\yellow}{\statistic_\policy}$};
\draw (Sj) [->, dashed] to (statisticj);
\draw (statisticj) [->, dashed] to (Tj);

\node (value) [morphism, minimum width=.6cm, minimum height=.6cm] at (-1.5, 2) {$\svalue_\policy$};
\draw (Si) [->, dashed, out=180, in=90] to (value);
\draw (value) [->, dashed, out=-90, in=180] to (Ri);

\end{tikzpicture}
\end{adjustbox}
\caption[State statistic bidirectional process]{%
By combining the recursive generation and aggregation of rewards, we can express the \textcolor{\yellow}{state statistic function} $\mathcolor{\yellow}{\statistic_\policy}: \States \to \Statistics$ as a composition of \emph{bidirectional processes}.
The \textcolor{\blue}{forward process} $\States \to \Rewards \times \States$, parameterized by a policy $\policy$, takes a state $s_t \in \States$ and generates a reward $r_{t+1} \in \Rewards$ and the next state $s_{t+1} \in \States$.
The \textcolor{\red}{backward process} $\Rewards \times \Statistics \to \Statistics$ takes a statistic $\tau_{t+1} \in \Statistics$ from the future and updates it with the previously generated reward $r_{t+1} \in \Rewards$ to produce the current statistic $\tau_t \in \Statistics$.
These bidirectional processes continue until a terminal state is reached, at which point its statistic is assigned the initial value $\init \in \Statistics$.
See \cref{app:algebra} for more details.
}
\label{fig:optic_state_value}
\end{figure}
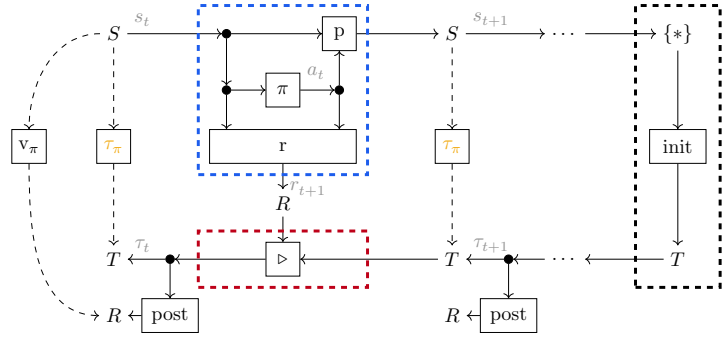

While prior work such as \citet{quah2006maximum} defined the recursive structure of the value function directly, our approach derives it from the recursive structure of the reward generation and aggregation processes.
Examples of recursive reward aggregation functions are provided in \cref{tab:recursive_aggregation_functions}.
An illustration of the recursive structure is given in \cref{fig:optic_state_value}.


\subsection{Policy evaluation: Iterative statistic function estimation}
\label{ssec:policy_evaluation}

Next, we consider how to estimate the state statistic function $\statistic_\policy: \States \to \Statistics$ for an arbitrary policy $\policy$, known as the \emph{policy evaluation} problem \citep[Sections~4.1 and 11.4]{sutton1998introduction}.
We introduce a generalized \emph{Bellman operator} and prove the uniqueness of its fixed points under certain conditions.
This result enables iterative statistic/value function estimation used in \emph{policy iteration} and modern \emph{actor-critic} methods \citep{barto1983neuronlike, mnih2016asynchronous, haarnoja2018soft, fujimoto2018addressing}.
Concretely, the Bellman operator is defined as follows:
\begin{restatable}[Bellman operator]{definition}{BellmanOperatorState}
\label{def:bellman_operator_state}
Given a policy $\policy$, a transition function $\transition$, a reward function $\reward$, a terminal condition $\terminal$, and a recursive statistic aggregation function $\aggiu$ (\cref{def:recursive_aggregation}), the \emph{Bellman operator} $\Bellman_\policy: [\States, \Statistics] \to [\States, \Statistics]$ for a function $\statistic: \States \to \Statistics$ is defined by
\begin{equation}
\Bellman_\policy \statistic: \States \to \Statistics
\defeq
\anon{s}{\scases{\init}{
\reward_\policy(s) \update \statistic(\transition_\policy(s))
}}.
\end{equation}
\end{restatable}

According to the Bellman equation in \cref{thm:bellman_equation_state}, we have $\Bellman_\policy \statistic_\policy = \statistic_\policy$, which means that the state statistic function $\statistic_\policy$ is a fixed point of the Bellman operator.
Then, we can generalize the classical fixed point theorem under the following condition:
\begin{definition}[Contractive update function]
\label{def:contraction}
An update function $\update: \Rewards \times \Statistics \to \Statistics$ is \emph{contractive} with respect to a premetric $d_\Statistics$ on statistics $\Statistics$ if
$
\forall r \in \Rewards.\;
\forall \tau_1, \tau_2 \in \Statistics.\;
d_\Statistics(r \update \tau_1, r \update \tau_2) 
\leq
k \cdot d_\Statistics(\tau_1, \tau_2)
$,
where $k \in [0, 1)$ is a constant.
In other words, $r \update (-): \Statistics \to \Statistics$ is a contraction for all $r \in \Rewards$.
\end{definition}

\begin{restatable}[Uniqueness of fixed points of Bellman operator]{theorem}{BellmanFixpointState}
\label{thm:bellman_fixpoint_state}
Let $\statistic_1, \statistic_2: \States \to \Statistics$ be fixed points of the Bellman operator $\Bellman_\policy$ (\cref{def:bellman_operator_state}).
If the update function $\update$ is contractive with respect to a premetric $d_\Statistics$ on statistics $\Statistics$ (\cref{def:contraction}), then $d_\Statistics(\statistic_1(s), \statistic_2(s)) = 0$ for all states $s \in \States$.
If $d_\Statistics$ is a strict premetric, then $\statistic_1 = \statistic_2 = \statistic_\policy$.
\end{restatable}

This result applies to a broad class of recursive aggregation functions beyond the discounted sum.
See \cref{app:metric} for further discussion on the premetric $d_\Statistics$ and the Bellman operator $\Bellman_\policy$.


\subsection{Policy optimization: Optimal policies and optimal value functions}
\label{ssec:policy_optimization}

Finally, we consider how to find an \emph{optimal policy} and compute its statistic/value functions recursively based on the Bellman equation in \cref{thm:bellman_equation_state}:
\begin{definition}[Optimal policy]
\label{def:optimality}
Given a preorder $\leq_\Statistics$ on statistics $\Statistics$, a policy $\policy_*$ is an \emph{optimal policy} if
$
\forall \policy.\;
\forall s \in \States.\;
\statistic_\policy(s) \leq_\Statistics \statistic_{\policy_*}(s)
$,
which has the \emph{optimal state statistic function} $\statistic_*: \States \to \Statistics \defeq \statistic_{\policy_*}$ and the \emph{optimal state value function} $\svalue_*: \States \to \Rewards \defeq {\post} \compL \statistic_*$.
\end{definition}

\begin{restatable}[Bellman optimality equation for the state statistic function]{theorem}{OptimalityEquationState}
\label{thm:optimality_equation_state}
Given a preorder $\leq_\Statistics$ on statistics $\Statistics$, the optimal state statistic function $\statistic_*$ (\cref{def:optimality}) satisfies
\begin{equation}
\statistic_*: \States \to \Statistics
\defeq
\anon{s}{\scases{\init}{
\displaystyle
\sup_{a \in \Actions} \parens*{\reward(s, a) \update \statistic_*(\transition(s, a))}
}}.
\end{equation}
\end{restatable}

\cref{def:optimality} and \cref{thm:optimality_equation_state} are analogous to their classical counterparts \citep[Section~3.6]{sutton1998introduction}, but they extend to arbitrary recursive aggregation functions and allow comparisons using a preorder $\leq_\Statistics$ on statistics.
A \emph{Bellman optimality operator} $\Bellman_*$ can be defined similarly to the Bellman operator in \cref{def:bellman_operator_state}, and we can prove the uniqueness of its fixed points under certain conditions.
This result enables the \emph{value iteration} algorithm \citep[Section~4.4]{sutton1998introduction}, \emph{temporal difference} methods such as \emph{Q-learning} \citep{watkins1989learning}, and deep Q-network (DQN) based methods \citep{mnih2013playing, bellemare2017distributional} to find the optimal policy $\policy_*$.
See \cref{app:order} for further discussion on the preorder $\leq_\Statistics$ and the Bellman optimality operator $\Bellman_*$.

\section{From deterministic to stochastic Markov decision processes}
\label{sec:stochastic}
In this section, we briefly discuss the extension of our framework to the stochastic setting.
We show that the deterministic and stochastic settings share a fundamental similarity: all \emph{recursive structures} remain unchanged, except that (deterministic) functions are replaced by \emph{stochastic functions}, and function composition is replaced by marginalization over the intermediate variable, as described by the \emph{Chapman--Kolmogorov equation} \citep{giry1982categorical, puterman1994markov}.
The main difference is that the stochastic setting allows for a richer class of aggregation functions \citep{bellemare2023distributional}, where the non-commutativity and non-distributivity of certain operations can lead to more complex behaviors.


\paragraph{Notation}

Slightly abusing notation, we use the same symbols to denote the \emph{measurable spaces} of states $\States$, actions $\Actions$, rewards $\Rewards$, and statistics $\Statistics$.
For a measurable space $C$, we write $\Probability{C}$ for the measurable space of all \emph{probability measures} on $C$, and we denote by $\delta_c \in \Probability{C}$ the \emph{Dirac measure} concentrated at $c \in C$.
An identity stochastic function $\id_C: C \to \Probability{C}: \anon{c}{\delta_c}$ maps an element $c \in C$ to the Dirac measure $\delta_c \in \Probability{C}$.
We consider stochastic transition $\transition: \StateActions \to \Probability\States$ and policy $\policy: \States \to \Probability\Actions$, while other functions can be deterministic.
We also use the usual conditional distribution notation such as  $\transition(s' | s, a)$ and $\policy(a | s)$.
 

\paragraph{Stochastic composite functions}

In the stochastic setting, we can compose two stochastic functions by marginalizing over the intermediate variable.
Additionally, we can compose a stochastic function with a deterministic one using the \emph{pushforward} operation, which is equivalent to treating deterministic functions as stochastic functions to Dirac measures.
Then, we can define stochastic versions of
\begin{itemize}
\item \emph{state transition}
$
\transition_\policy: \States \to \Probability\States
\defeq
\transition \compL {\pair}
=
\anon{s}{s' \sim \int_\Actions \transition(s' | s, a) \policy(a | s) \D{a}}
$
and
\item \emph{state reward function}
$
\reward_\policy: \States \to \Probability\Rewards
\defeq
\reward \compL {\pair}
=
\anon{s}{r \sim \int_\Actions \delta_{\reward(s, a)}(r) \policy(a | s) \D{a}}
$.
\end{itemize}


\paragraph{Stochastic recursive functions}

Analogous to \cref{thm:bellman_equation_state}, we can derive the recursive calculation of the \emph{stochastic state statistic function} $\statistic_\policy: \States \to \Probability\Statistics$, known as the \emph{distributional Bellman equation} \citep{ morimura2010nonparametric, morimura2010parametric, bellemare2017distributional}, for any recursive aggregation function $\aggiu$:
\begin{equation}
\label{eq:stochastic_state_statistic}
\statistic_\policy: \States \to \Probability\Statistics
=
\anon{s}{\tau \sim \scases{\delta_{\init}}{
r \update \tau'
\Bigm|
r \sim \reward_\policy(r | s),
\tau' \sim \int_S \statistic_\policy(\tau' | s') \transition_\policy(s' | s) \D{s'}
}}.
\end{equation}


\begin{figure}
\centering
\vspace{-2em}
\begin{adjustbox}{scale=.8}
\begin{tikzpicture}
\node (Ti) at (0, 0) {$\Probability\Rewards$};
\node (R) at (2, 1) {$\Probability\Rewards$};
\node at (2, 2) {$\vdots$};
\node (Tj) at (4.2, 0) {$\Probability\Rewards$};
\node at (5, 0) {$\dots$};
\node at (6, 0) {$\delta_0$};
\node at (0, 2.2) {};

\node (add) [morphism, minimum width=.6cm, minimum height=.6cm] at (2, 0) {$+$};
\node (discount) [morphism, minimum width=.6cm, minimum height=.6cm] at (3, 0) {$\gamma$};
\draw (R) [->] to (add);
\draw (Tj) [->] to (discount);
\draw (discount) [->] to (add);
\draw (add) [->] to (Ti);
\node (copyT) [diagonal] at (1, 0) {};
\node (exp) [morphism, minimum width=.6cm, minimum height=.6cm] at (1, -1) {$\E$};
\node (Ri) at (0, -1) {$\Rewards$};
\draw (copyT) [->] to (exp);
\draw (exp) [->] to (Ri);

\node [draw, rectangle, dashed, minimum width=2cm, minimum height=1cm, color=\red, line width=1.5pt] at (2.5, 0) {};
\node [draw, rectangle, dashed, minimum width=1cm, minimum height=1cm, line width=1.5pt] at (6, 0) {};

\node [above=0 of Ti.north west, black!42] {$r \dadd s$};
\node [above=0 of Tj.north east, black!42] {$s$};
\node [right=0 of R, black!42] {$r$};
\node [above=0 of Ri.north west, black!42] {$\E[r \dadd s]$};
\end{tikzpicture}
\hspace{1cm}
\begin{tikzpicture}
\node (Ti) at (0, 0) {$\Rewards$};
\node (R) at (2, 2) {$\Probability\Rewards$};
\node (Tj) at (4.2, 0) {$\Rewards$};
\node at (5, 0) {$\dots$};
\node at (6, 0) {$0$};
\node at (0, 2.2) {};

\node (add) [morphism, minimum width=.6cm, minimum height=.6cm] at (2, 0) {$+$};
\node (discount) [morphism, minimum width=.6cm, minimum height=.6cm] at (3, 0) {$\gamma$};
\node (exp) [morphism, minimum width=.6cm, minimum height=.6cm] at (2, 1) {$\E$};
\draw (R) [->] to (exp);
\draw (exp) [->] to (add);
\draw (Tj) [->] to (discount);
\draw (discount) [->] to (add);
\draw (add) [->] to (Ti);
\node (copyT) [diagonal] at (1, 0) {};
\node (Ri) at (0, -1) {$\Rewards$};
\draw (copyT) [->] to (1, -1) to (Ri);

\node [draw, rectangle, dashed, minimum width=2cm, minimum height=2cm, color=\red, line width=1.5pt] at (2.5, .5) {};
\node [draw, rectangle, dashed, minimum width=1cm, minimum height=1cm, line width=1.5pt] at (6, 0) {};

\node [above=0 of Ti.north west, black!42] {$\E[r] \dadd \E[s]$};
\node [above=0 of Tj.north east, black!42] {$\E[s]$};
\node [right=0 of R, black!42] {$r$};
\end{tikzpicture}
\end{adjustbox}
\caption[Expected discounted sum of rewards vs.~discounted sum of expected rewards]{%
The recursive structures of
(left) the \emph{expected discounted sum of rewards} $\E[r \dadd s]$ and 
(right) the \emph{discounted sum of expected rewards} $\E[r] \dadd \E[s]$.
}
\label{fig:expected_sum}
\end{figure}
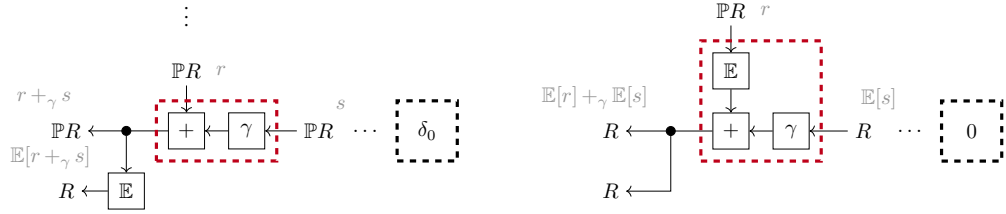


\paragraph{Stochastic aggregation functions}

Note that this framework also accommodates the traditional \emph{expected discounted sum of rewards} $\E\bracks*{\sum_{t=1}^\horizon \gamma^{t-1} r_t}$ learning objective, by selecting $\delta_0$ as $\init$, the (pushforward through) discounted addition function $\dadd: \Rewards \times \Rewards \to \Rewards$ as the update function $\update$, and the expectation operator $\E: \Probability{\Rewards} \to \Rewards$ as $\post$.
The stochastic statistic function $\statistic_\policy: S \to \Probability{\Rewards}$ in \cref{eq:stochastic_state_statistic}, refered to as the \emph{value distribution} in \citet{bellemare2017distributional}, outputs the distribution of the discounted sum of rewards, while the value function outputs its expectation.
Since the expectation distributes over the discounted addition, by changing the update function and initial value, we can recursively calculate the \emph{discounted sum of expected rewards} $\sum_{t=1}^\horizon \gamma^{t-1} \E[r_t]$ instead (see \cref{fig:expected_sum}), which is the traditional approach in RL \citep{sutton1998introduction}.
In this case, the statistic function and the value function coincide, as no post-processing is required.
However, \citet{bellemare2017distributional} have shown that even in the discounted sum setting, the Bellman operator may be a contraction in some metrics but not in others, while the Bellman optimality operator is a contraction only in expectation and not in any distributional metric, leading to different convergence behaviors.
These challenges persist and may become unavoidable when using alternative aggregation functions due to the inconsistency between expected aggregated rewards and aggregated expected rewards.
We discuss this further in \cref{app:stochastic} and leave a full investigation for future work.

\section{Experiments}
\label{sec:experiments}
In this section, we empirically evaluate the proposed \emph{recursive reward aggregation} technique across a variety of environments and optimization objectives to support the following claims:
\begin{itemize}
\item
Different aggregation functions significantly influence policy preferences.
Selecting an appropriate aggregation function is an alternative approach to optimizing policies for specific objectives and aligning agent behaviors with task-specific goals without modifying rewards (\cref{ssec:grid,ssec:wind,ssec:physics}).
\item
In challenging real-world applications such as portfolio optimization, our method can directly optimize desired evaluation criteria, demonstrating superior performance compared to existing approaches and showcasing its practical effectiveness (\cref{ssec:portfolio}).
\end{itemize}


\subsection{Grid-world: Value-based methods for discrete planning}
\label{ssec:grid}

First, we present illustrative experiments in a simple grid-world environment to demonstrate the fundamental impact of different recursive reward aggregation functions on learned policies.


\paragraph{Environment}

\begin{figure}
\centering
\begin{subfigure}{0.2\linewidth}
\includegraphics[width=\textwidth]{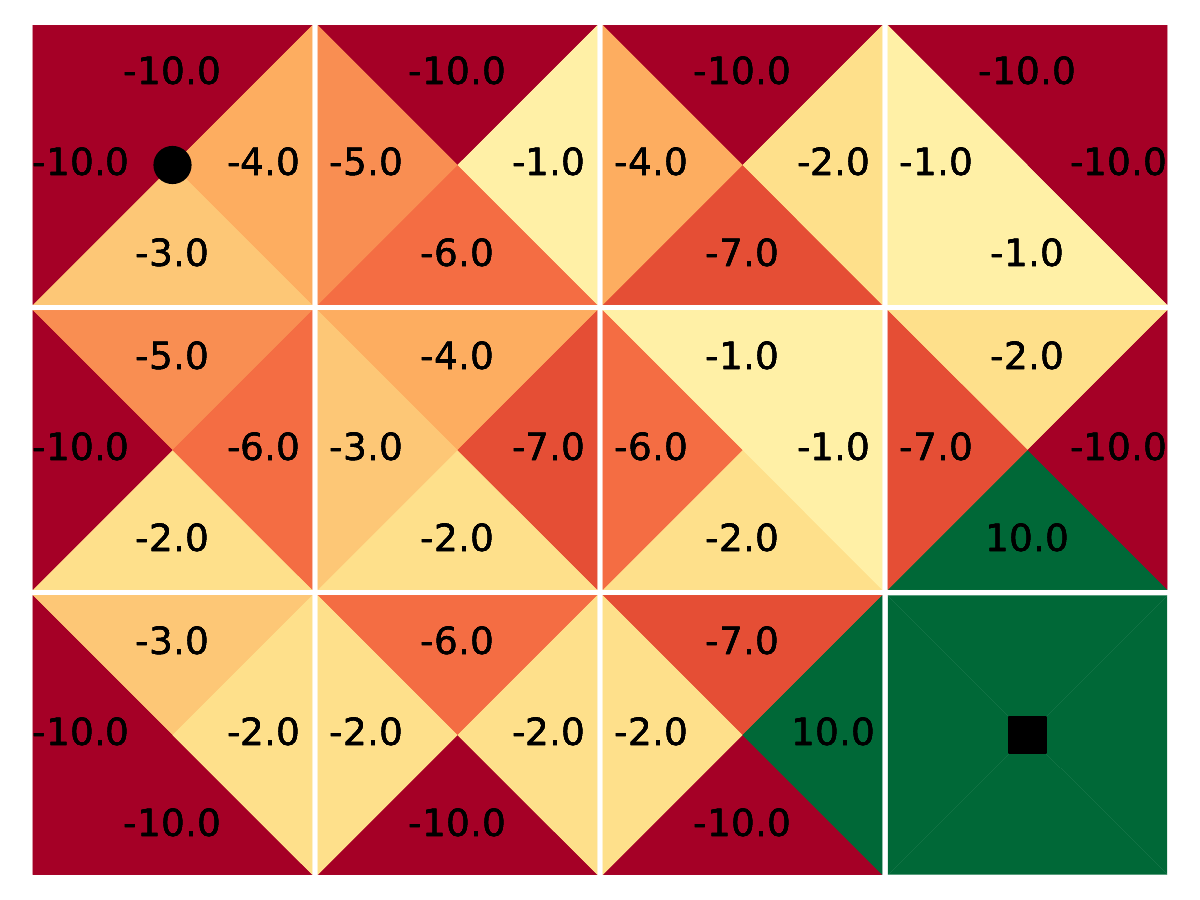}
\caption{Environment}
\label{sfig:grid_env}
\end{subfigure}%
\begin{subfigure}{0.2\linewidth}
\includegraphics[width=\textwidth]{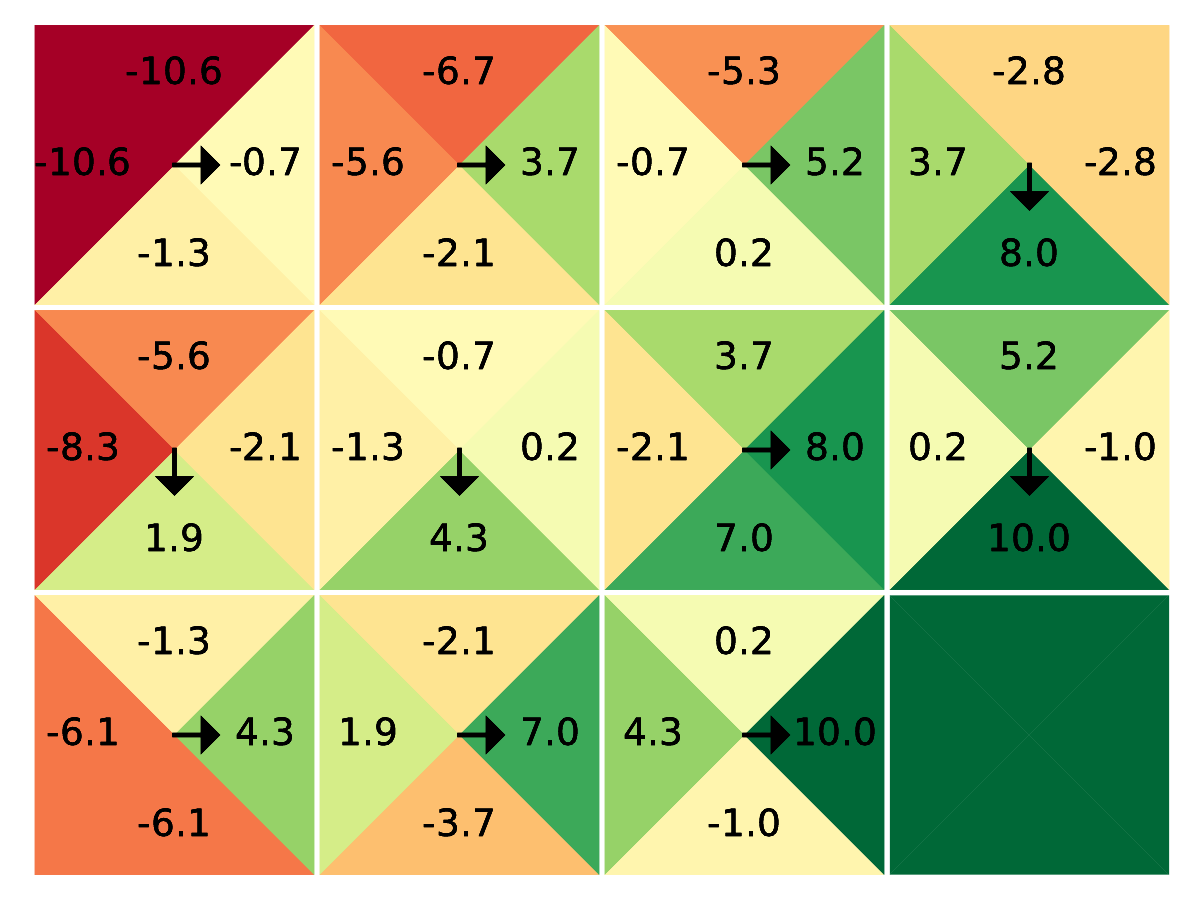}
\caption{$\fsum_{0.9}$}
\label{sfig:grid_dsum}
\end{subfigure}%
\begin{subfigure}{0.2\linewidth}
\includegraphics[width=\textwidth]{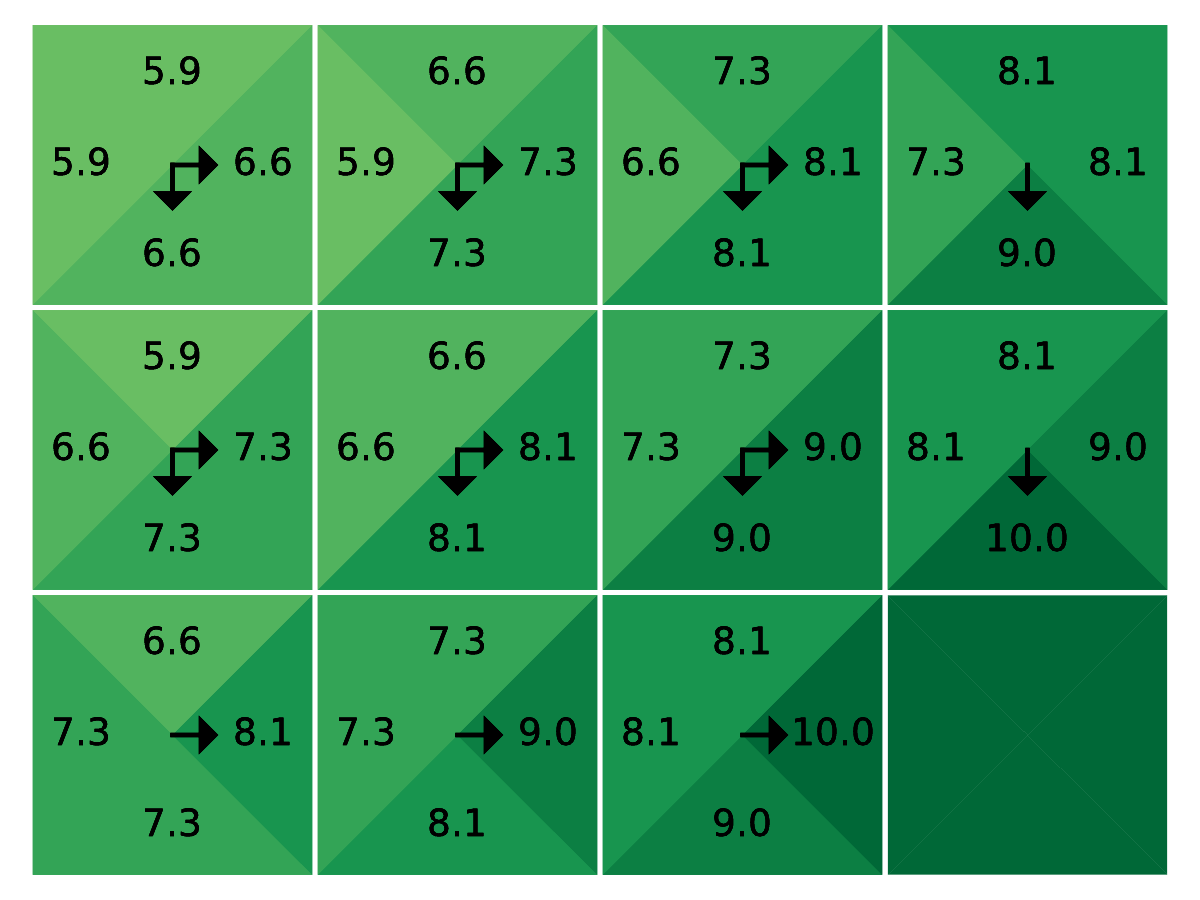}
\caption{$\max_{0.9}$}
\label{sfig:grid_dmax}
\end{subfigure}%
\begin{subfigure}{0.2\linewidth}
\includegraphics[width=\textwidth]{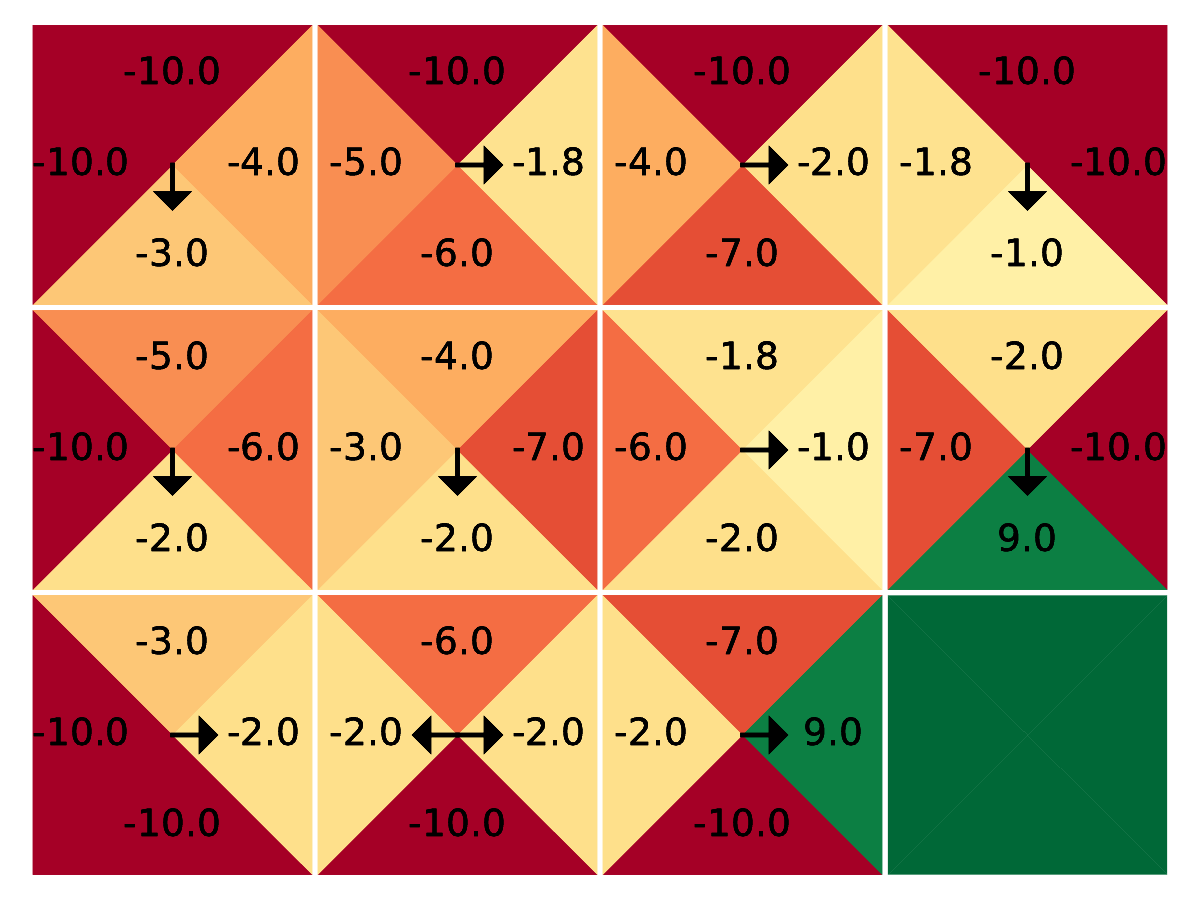}
\caption{$\min_{0.9}$}
\label{sfig:grid_dmin}
\end{subfigure}%
\begin{subfigure}{0.2\linewidth}
\includegraphics[width=\textwidth]{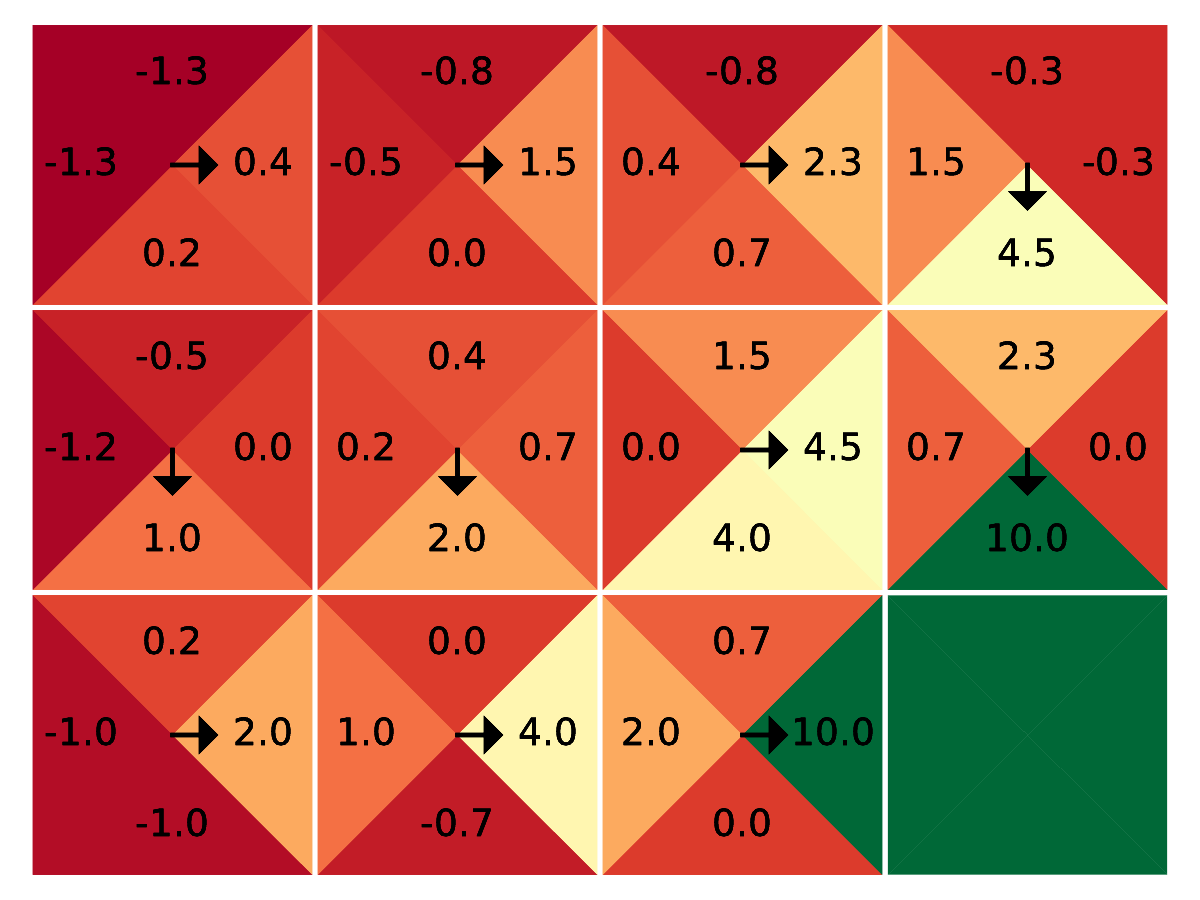}
\caption{$\mean$}
\label{sfig:grid_mean}
\end{subfigure}%
\vspace{-1ex}
\caption[Grid-world]{%
\textbf{Grid-world}:
\cref{sfig:grid_env} shows the discrete environment and the reward function $\reward(s, a)$, where the agent starts from the top-left corner $\bullet$ and needs to reach the goal at the bottom-right corner {\tiny$\blacksquare$}.
\cref{sfig:grid_dsum,sfig:grid_dmax,sfig:grid_dmin,sfig:grid_mean} show the optimal state-action value functions $\qvalue_*(s, a)$ under different aggregations.
}
\label{fig:grid}
\end{figure}

\cref{sfig:grid_env} shows the results for a $3 \times 4$ grid environment, where an agent navigates from the top-left corner to a fixed goal at the bottom-right corner.
As shown in \cref{sfig:grid_env}, the agent receives a small negative reward at each step, which varies across states, and a positive reward upon reaching the terminal state.


\paragraph{Method}

For this discrete environment, we modified the Q-learning algorithm \citep{watkins1989learning, watkins1992q} using the Bellman optimality operator introduced in \cref{ssec:policy_optimization} (more specifically, the one for the state-action statistic function in \cref{def:optimality_operator_state-action}).
We used four recursive aggregation functions: discounted sum, discounted max, min, and mean, as detailed in \cref{tab:recursive_aggregation_functions}.
The detailed algorithm is provided in \cref{alg:q-learning} in \cref{app:algorithms}.


\paragraph{Results}

Compared to the standard discounted sum aggregation (\cref{sfig:grid_dsum}), optimizing for the discounted max reward (\cref{sfig:grid_dmax}) makes the agent indifferent to intermediate costs, favoring shorter paths to the goal.
In contrast, the discounted min (\cref{sfig:grid_dmin}) encourages risk-averse behavior, while the mean aggregation (\cref{sfig:grid_mean}) promotes efficiency by maximizing average reward per step.
Overall, these results demonstrate how each aggregation function uniquely impacts reward evaluation and policy preferences.


\begin{figure}
\centering
\begin{subfigure}{0.25\linewidth}
\centering
\includegraphics[height=3.3cm]{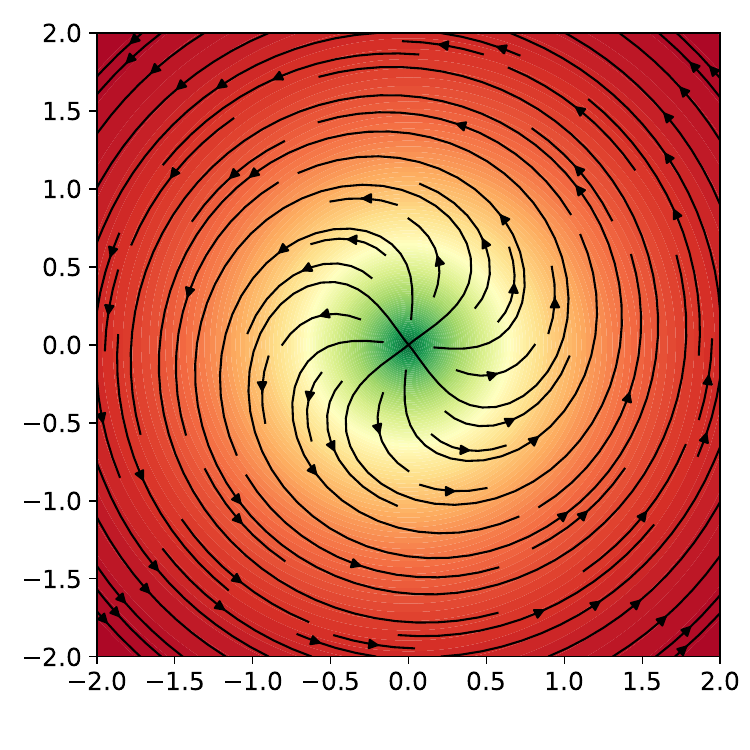}
\vspace{-1ex}
\caption{Environment}
\label{sfig:wind_env}
\end{subfigure}%
\hfill
\begin{subfigure}{0.25\linewidth}
\centering
\includegraphics[height=3.3cm]{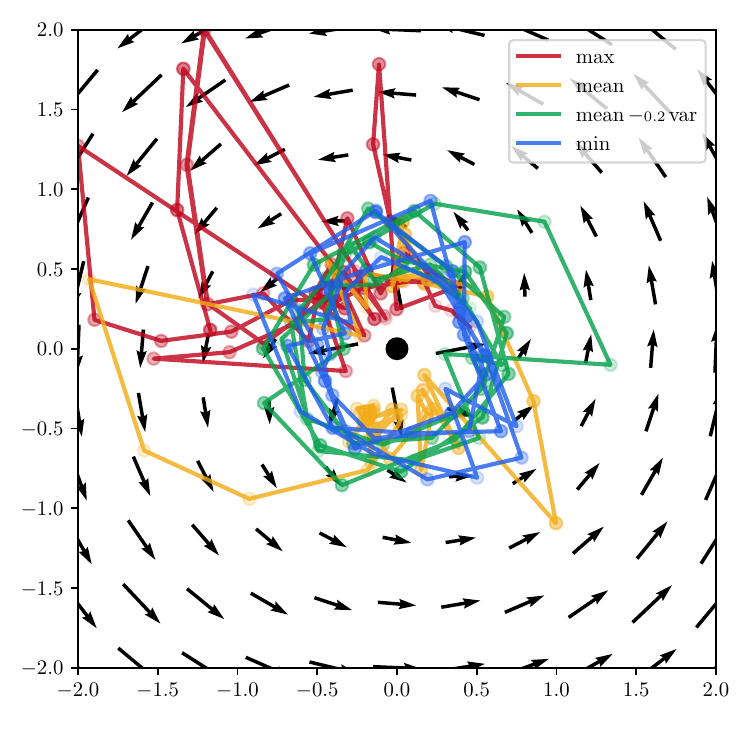}
\vspace{-1ex}
\caption{Trajectory}
\label{sfig:wind_trajectory}
\end{subfigure}%
\hfill
\begin{subfigure}{0.25\linewidth}
\centering
\includegraphics[height=3.3cm]{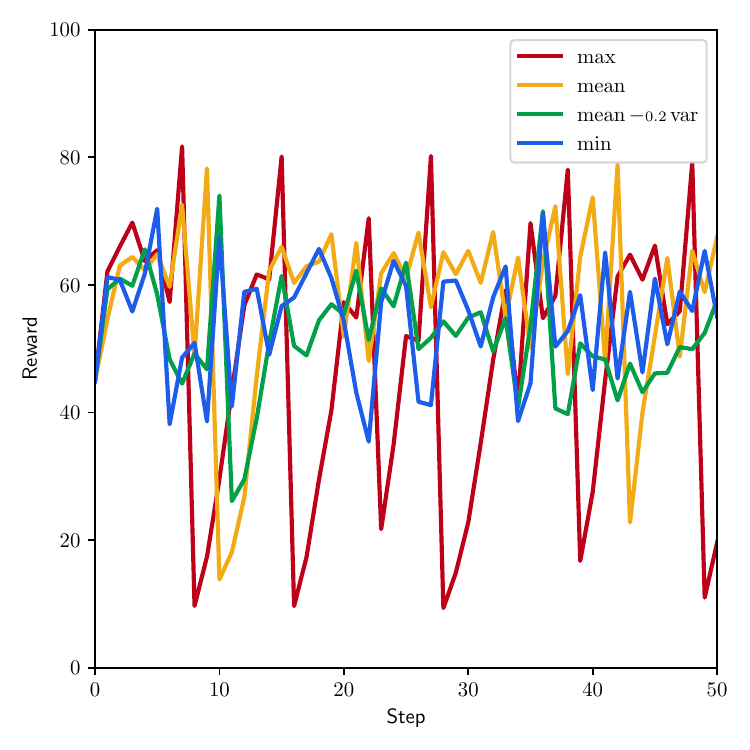}
\vspace{-1ex}
\caption{Rewards}
\label{sfig:wind_rewards}
\end{subfigure}%
\hfill
\begin{subfigure}{0.25\linewidth}
\centering
\includegraphics[height=3.3cm]{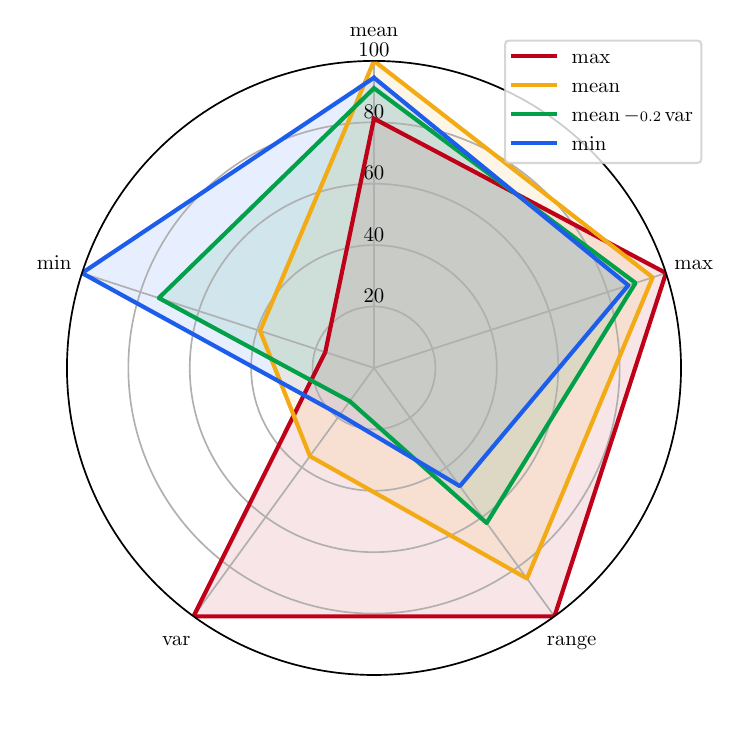}
\vspace{-1ex}
\caption{Metrics}
\label{sfig:wind_radar}
\end{subfigure}%
\vspace{-1ex}
\caption[Wind-world]{%
\textbf{Wind-world}:
\cref{sfig:wind_env} shows the continuous environment, where the agent encounters wind disturbances (visualized with streamlines) and receives higher rewards near the center (depicted with colored contours).
\cref{sfig:wind_trajectory} illustrates the trajectories of agents trained using different aggregation functions, while \cref{sfig:wind_rewards} compares the rewards obtained by each agent.
\cref{sfig:wind_radar} presents the evaluation metrics, highlighting the impact of aggregation functions on performance.
}
\label{fig:wind}
\end{figure}


\subsection{Wind-world: Policy improvement methods for trajectory optimization}
\label{ssec:wind}

Next, we show that the recursive reward aggregation technique can also be seamlessly integrated into methods for continuous state and action spaces to optimize trajectories in complex environments.


\paragraph{Environment}

Inspired by \citet{dorfman2021offline, ackermann2024offline}, we designed a two-dimensional continuous environment where an agent navigates to a fixed goal amidst varying wind disturbances, as shown in \cref{sfig:wind_env}.
This setup allows us to evaluate the impact of different aggregation functions on trajectory optimization.


\paragraph{Method}

For this continuous environment, we utilized the Proximal Policy Optimization (PPO) algorithm \citep{schulman2017proximal}, which is a widely used policy improvement method.
We estimated the value function using the Bellman operator for the state statistic function in \cref{def:bellman_operator_state}.
The detailed algorithm is provided in \cref{alg:ppo} in \cref{app:algorithms}.


\paragraph{Results}

The results in \cref{sfig:wind_trajectory,sfig:wind_rewards,sfig:wind_radar} show that different aggregation functions lead to distinct trade-offs in trajectory optimization.
Specifically, the max aggregation function prioritizes high-reward paths, while the min function ensures more conservative and consistent behavior.
The variance-regularized mean aggregation provide balanced strategies, demonstrating the flexibility of the recursive reward aggregation technique in optimizing diverse objectives.


\begin{figure}
\centering
\includegraphics[width=\linewidth]{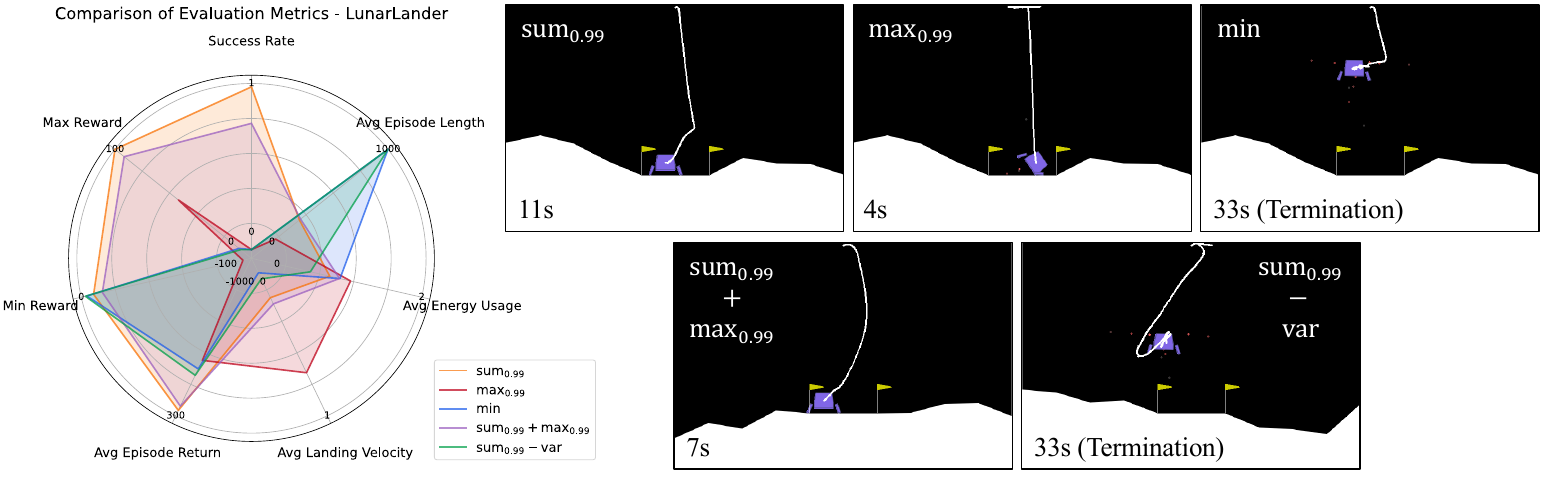}
\caption[Lunar Lander]{%
\textbf{Lunar Lander Continuous}: Comparison of five reward aggregation methods.
(Left) Radar plot showing performance across seven evaluation metrics, averaged over four random seeds.
(Right) Sample trajectories illustrating the qualitative behaviors induced by each aggregation method.
}
\label{fig:lunar_lander}
\end{figure}


\subsection{Physics simulation: Actor-critic methods for continuous control}
\label{ssec:physics}

Then, we extend our evaluation to more complex physics simulation environments.


\paragraph{Environment}

We conducted experiments on three continuous control environments:
(i) Lunar Lander Continuous \citep{gym} from the Box2D environment,
(ii) Hopper \citep{erez2012infinite}, and
(iii) Ant \citep{schulman2016high} simulated using MuJoCo \citep{mujoco}.
A detailed description of these environments can be found in \cref{ssec:physics_details}.


\paragraph{Method}

In these experiments, we employed the Twin Delayed Deep Deterministic Policy Gradient (TD3) algorithm \citep{fujimoto2018addressing}, with a modified recursive version detailed in \cref{alg:td3} in \cref{app:algorithms}.
We considered five different reward aggregation functions:
(i) discounted sum ($\fsum_{0.99}$),
(ii) discounted max ($\max_{0.99}$),
(iii) min ($\min$),
(iv) discounted sum plus discounted max ($\fsum_{0.99} + \max_{0.99}$), and
(v) discounted sum minus variance ($\fsum_{0.99} - \var$).


\paragraph{Results}

The results for Lunar Lander Continuous are provided in \cref{fig:lunar_lander}, with results for other environments in \cref{ssec:physics_details}.
As a goal-reaching task, Lunar Lander Continuous reveals how different aggregation strategies influence landing behavior and overall performance.

With the $\fsum_{0.99}$ aggregation, which serves as the baseline, the agent learns a balanced landing strategy, effectively managing thrust control to achieve a smooth descent while minimizing fuel consumption.
In contrast, the $\max_{0.99}$ aggregation encourages the agent to seek high instantaneous rewards, leading to aggressive thrusting behaviors.
As a consequence, the lander may exhibit erratic flight patterns, either applying excessive thrust to maximize immediate reward or failing to decelerate properly, which increases the likelihood of hard landings, instability, or even complete mission failure.
This outcome underscores the risk of optimizing for short-term reward spikes at the expense of long-term stability and control.
The $\min$ aggregation demonstrates its effectiveness in risk-averse tasks, as it prioritizes maximizing the worst-case outcomes rather than accumulate reward.
The agent adopts a cautious descent strategy, reducing the likelihood of crashes by avoiding sudden thrust changes.
Furthermore, since goal-reaching tasks inherently align cumulative and peak rewards, the $\fsum_{0.99} + \max_{0.99}$ performs similarly to $\fsum_{0.99}$.
However, compared to $\fsum_{0.99}$, it encourages slightly more aggressive strategies, potentially enabling faster landings but at the cost of a higher risk of failure when pursuing large single-step rewards.
Finally, in the $\fsum_{0.99} - \var$ aggregation, the lander remains airborne, ultimately leading to mission termination.
This occurs because both successful and failed landings yield large positive or negative rewards, the agent attempts to avoid these extremes, increasing variance and leading to hesitant and inefficient control.
This highlights the conflict between variance minimization and goal-reaching tasks, where effective performance relies on high-reward actions often discouraged by variance penalties.
These findings emphasize the need to choose aggregation strategies that align with the specific demands of the task.


\begin{table}[t]
\centering
\caption[Sharpe ratio]{%
Performance comparison of different methods for portfolio optimization using the Sharpe ratio.
The table reports the mean and standard deviation of the Sharpe ratio across five random seeds during the test period, where a higher value indicates better risk-adjusted returns.
}
\begin{tabular}{cccc}
\toprule
& DiffSharpe & NCMDP & Ours
\\
\midrule
Sharpe Ratio (Test) & $0.29 \pm 1.22$ & $0.48 \pm 0.79$ & $1.12 \pm 0.92$ \\
\bottomrule
\end{tabular}
\label{tab:portfolio}
\end{table}


\subsection{Real-world application: Sharpe ratio in portfolio optimization}
\label{ssec:portfolio}

Lastly, we evaluate the practical applicability of our method in a real-world application.

\emph{Portfolio optimization} \citep{moody1998performance, sood2023deep, liu2024dynamic} is a real-world financial application where an agent (or investor) determines the optimal allocation of assets across different investment options.
It can be framed as a sequential decision-making problem as the agent continuously adjusts the portfolio in response to evolving market conditions, fluctuating asset prices, and shifting risk preferences, rather than setting a static allocation.
Each decision not only influences immediate returns but also conditions future decisions.

The \emph{Sharpe ratio} \citep{sharpe1966mutual} is a standard metric for evaluating the performance of investment strategies by quantifying the trade-off between return and risk.
It is defined as the ratio of the average return (arithmetic mean) to the volatility of return (standard deviation) \citep[Eq.~(5.18)]{bodie2011investments}:
\begin{equation}
\operatorname{SharpeRatio}(r_{1:t})
\defeq
\frac{\mean(r_{1:t})}{\std(r_{1:t})},
\end{equation}
where $r_t \defeq (P_{t+1} - P_t) / P_t$ represents the simple return, and $P_t$ is the portfolio value at time $t$.
Since the Sharpe ratio is non-cumulative, previous RL approaches have relied on the approximate differential Sharpe ratio \citep{moody1998performance, moody2001learning} as a reward signal to facilitate learning.
However, this approach introduces an inconsistency between the learning objective and the actual Sharpe ratio, potentially leading to suboptimal policy learning.


\paragraph{Environment}

This experiment was conducted in a financial market simulation, where an agent learned to optimize portfolio allocations across 11 different {S\&P 500} sector indices from 2006 to 2021.
The environment is the same as that described by \citet{sood2023deep, nagele2024tackling}, with further details provided in \cref{ssec:portfolio_details}.


\paragraph{Baselines}

We considered two baseline methods:
(i) DiffSharpe \citep{moody1998performance, moody2001learning}, which optimizes an approximate differential Sharpe ratio, and
(ii) a non-cumulative Markov decision process (NCMDP) method proposed by \citet{nagele2024tackling}, which maps NCMDPs to standard MDPs and defines per-step rewards based on consecutive differences.


\paragraph{Method}

As demonstrated in \cref{tab:recursive_aggregation_functions}, since both mean and variance admit recursive computation, the Sharpe ratio can also be expressed and updated in a recursive manner.
This property allows our method to address the aforementioned inconsistency, aligning the learning objective with the true Sharpe ratio.
Our method is built upon the PPO \citep{schulman2017proximal} algorithm, with specific modifications on Bellman equation detailed in \cref{alg:ppo} in \cref{app:algorithms}.


\paragraph{Results}

We conducted experiments across five random seeds, reporting the mean and standard deviation of test performance.
Since a higher Sharpe ratio reflects superior risk-adjusted returns, the results in \cref{tab:portfolio} and \cref{fig:portfolio} in \cref{ssec:portfolio_details} indicate that our method often attains improved risk-reward balance relative to the baselines.
These results illustrate that modifying either the local reward signal or the global performance measure can create misalignment, leading to inconsistencies in policy training and suboptimal outcomes.
Unlike baseline methods, our method maintains the original per-step reward structure while estimating and optimizing the exact Sharpe ratio over the entire trajectory.
This design may help maintain alignment between training and evaluation, enabling the agent to focus more on long-term performance and become less sensitive to short-term fluctuations.

\section{Conclusion}
\paragraph{Summary}

In this paper, we revealed that the recursive structures in the standard MDP can be generalized to a broader class of recursive reward aggregation functions, resulting in generalized Bellman equations and operators.
Our theoretical analysis on the existence and uniqueness of fixed points of the generalized Bellman operators provides a solid foundation for designing RL algorithms based on recursive reward aggregation and understanding their convergence properties.
Empirical evaluations across discrete and continuous environments confirmed that different aggregation functions significantly influence policy preferences, and we can align the agent behavior with the task requirements by selecting appropriate aggregation functions.
These findings highlight the flexibility of recursive reward aggregation, paving the way for more versatile RL algorithms that can be tailored to complex task requirements.


\paragraph{Scope and limitations}

Our framework is designed for recursive aggregations.
As such, it does not directly support non-recursive objectives such as the median or semivariance, which cannot be computed using a bounded-size accumulator in an online fashion with a single pass.
Although approximate solutions may be feasible, e.g., sketching algorithms such as online quantile estimation \citep{greenwald2001space}, they fall outside the exact scope of our algebraic formulation.

Additionally, while our method frees the designer from modifying the reward function itself, it introduces a different axis of design: selecting or constructing a suitable aggregation function.
The space of meaningful aggregations is vast and may require domain-specific insight or empirical tuning.

Finally, our work is agnostic to the validity of the \emph{reward hypothesis} \citep{bowling2023settling}: the idea that all goals can be expressed as the maximization of expected cumulative scalar rewards.
We neither rely on this assumption nor seek to refute it.
Instead, we explore an orthogonal dimension of goal specification: how reward signals are aggregated over time.
This perspective complements traditional reward design and provides a flexible mechanism for aligning behavior with complex objectives, without requiring any claims about the ultimate expressiveness or limitations of scalar rewards.


\paragraph{Future work}

Future research could explore several extensions and applications of the proposed recursive reward aggregation framework.

First, since the abstract framework does not require the outputs of the generation function and the inputs of the aggregation function to be real values, one promising direction is to investigate the use of \emph{multi-dimensional objectives} or \emph{non-numerical feedback signals}, enhancing the flexibility and expressiveness of policy preferences, particularly in complex environments with intricate reward structures \citep{pitis2023consistent, wiltzer2024foundations} or constraints \citep{gattami2021reinforcement, wachi2024survey}.

Second, an important direction is to study generalized Bellman operators in the \emph{stochastic setting}, particularly their convergence behavior under distributional metrics for non-deterministic, non-Markovian, and non-stationary policies \citep{bellemare2023distributional}.
Another is to analyze quantile-based risk measures such as value-at-risk (VaR), conditional value-at-risk (CVaR), and entropic value-at-risk (EVaR) \citep{rockafellar2000optimization, sugiyama2010least, tamar2015optimizing, hau2023dynamic, hau2023entropic, hau2025q}, which are widely used in risk-sensitive decision-making, such as in fields like finance \citep{manganelli2001value}.

Third, extending the framework to \emph{approximate non-recursive aggregations} \citep{greenwald2001space} or to \emph{learn aggregation functions} from data \citep{zaheer2017deep, ong2022learnable} could broaden its applicability and automate goal specification.

Finally, applying recursive reward aggregation to real-world settings such as
(i) risk-sensitive decision-making,
(ii) risk-adjusted return optimization and portfolio diversification in finance, and
(iii) safe, robust, and multi-objective control in robotics and autonomous driving,
presents promising directions \citep{kober2013reinforcement, kiran2021deep, liu2024dynamic}.


\clearpage
\subsubsection*{Acknowledgments}
We are grateful to Tongtong Fang for carefully reviewing the abstract and introduction and offering insightful suggestions.
We also thank Qi Chen, Silviu Pitis, and Harley Wiltzer for their insightful discussions, as well as the anonymous reviewers and conference attendees for their constructive feedback and thought-provoking questions.

YT was supported by Institute for AI and Beyond, UTokyo.
SN was supported by JSPS KAKENHI Grant Number JP24KJ0818.
MS was supported by Institute for AI and Beyond, UTokyo.

\addcontentsline{toc}{section}{Bibliography}
\bibliography{references,references_category,references_reinforcement,references_asset}
\bibliographystyle{rlj}


\beginSupplementaryMaterials

{
\hypersetup{linkcolor=black}
\tableofcontents
\listoffigures
\listoftables
\listofalgorithms
}

\appendix

\clearpage
\section{State-action recursion}
\label{app:state-action}
In \cref{sec:fusion}, we introduced the recursive generation of rewards by iterating over \emph{states} $\States$.
In this section, we extend this framework to iterate over \emph{state-action pairs} $\StateActions$, which is crucial for defining the \emph{state-action value function} $\qvalue_\policy: \StateActions \to \Rewards$.


\subsection{State-action transition}

First, note that both \emph{pre-composing} and \emph{post-composing} the pairing function $\pair: \States \to \StateActions$ with the transition function $\transition: \StateActions \to \States$ yield transition functions:
\begin{itemize}
\item 
\emph{state transition}
$
\transition^\States_\policy: \States \to \States
\defeq
\transition \compL {\pair}
=
\anon{s}{\transition(s, \policy(s))}
$ and
\item
\emph{state-action transition}
$
\transition^\StateActions_\policy: \StateActions \to \StateActions
\defeq
\pair \compL \transition
=
\anon{s, a}{\transition(s, a), \policy(\transition(s, a))}
$.
\end{itemize}
We use the superscripts $\States$ and $\StateActions$ to indicate the domains/codomains of these transition functions.


\subsection{State-action step function and generation function}

Then, following the definitions of the state step function $\step^\States_\ptro: \States \to \singleton + \Rewards \times \States$ in \cref{eq:state_step} and generation function $\gen^\States_\ptro: \States \to \List{\Rewards}$ in \cref{eq:state_generation}, we can define the \emph{state-action step/generation functions} using the state-action transition $\transition^\StateActions_\policy$ and the reward function $\reward$:
\begin{align}
\step^\StateActions_\ptro &: \StateActions \to \singleton + \Rewards \times (\StateActions)
\defeq
\anon{s, a}{\scases{*}{
(\reward(s, a), \transition^\StateActions_\policy(s, a))
}},
\\
\gen^\StateActions_\ptro &: \StateActions \to \List{\Rewards}
\defeq
\anon{s, a}{\scases{\emptylist}{
\cons(\reward(s, a), \gen^\StateActions_\ptro(\transition^\StateActions_\policy(s, a)))
}}.
\end{align}


\subsection{State-action statistic function and value function}

Applying the same algebraic fusion technique \citep{hinze2010theory} used for the state statistic function $\statistic^\States_\policy: \States \to \Statistics$ in \cref{thm:bellman_equation_state}, we can define the \emph{state-action statistic function} $\statistic^\StateActions_\policy: \StateActions \to \Statistics$ and derive its corresponding Bellman equation as follows:
\begin{restatable}[Bellman equation for the state-action statistic function]{theorem}{BellmanEquationStateAction}
\label{thm:bellman_equation_state-action}
Given a recursive reward generation function $\gen^\StateActions_\ptro$ and a recursive statistic aggregation function $\aggiu$ (\cref{def:recursive_aggregation}), their composition, called the state-action statistic function $\statistic^\StateActions_\policy: \States \to \Statistics$, satisfies
\begin{align}
\label{eq:state-action_statistic}
\nonumber
\statistic^\StateActions_\policy: \StateActions \to \Statistics
\defeq{}&
{\aggiu} \compL {\gen^\StateActions_\ptro}
\\
={}&
\anon{s, a}{\scases{\init}{
\reward(s, a) \update \statistic^\StateActions_\policy(\transition^\StateActions_\policy(s, a))
}}.
\end{align}
\end{restatable}

Similarly, the \emph{state-action value function} $\qvalue_\policy: \StateActions \to \Rewards \defeq {\post} \compL \statistic^\StateActions_\policy$ is the composition of the state-action statistic function $\statistic^\StateActions_\policy: \StateActions \to \Statistics$ with the post-processing function $\post: \Statistics \to \Rewards$.


\subsection{Relationship between state and state-action statistic functions}

We can now state the theorem that relates the state and state-action statistic functions:
\begin{restatable}[Relationship between state and state-action statistic functions]{theorem}{StateStateActionRelationship}
\label{thm:state/state-action_relationship}
Given a recursive reward generation function $\gen_\ptro$ (\cref{def:recursive_generation}) and a recursive statistic aggregation function $\aggiu$ (\cref{def:recursive_aggregation}), the state statistic function $\statistic^\States_\policy: \States \to \Statistics$ in \cref{eq:state_statistic} and the state-action statistic function $\statistic^\StateActions_\policy: \StateActions \to \Statistics$ in \cref{eq:state-action_statistic} satisfy
\begin{align}
\label{eq:state_state-action}
\statistic^\States_\policy
&=
\statistic^\StateActions_\policy \compL {\pair}
: \States \to \Statistics
& \text{(for all states),}
\\
\label{eq:state-action_state}
\statistic^\StateActions_\policy
&=
\reward \update (\statistic^\States_\policy \compL \transition)
: \StateActions \to \Statistics
& \text{(for all non-terminal states).}
\end{align}
\end{restatable}

\begin{corollary}[Relationship between state and state-action value functions]
The state value function $\svalue_\policy: \States \to \Rewards$ and the state-action value function $\qvalue_\policy: \StateActions \to \Rewards$ satisfy
\begin{equation}
  \svalue_\policy
=
  \qvalue_\policy \compL {\pair}
: \States \to \Rewards.
\end{equation}
\end{corollary}

In summary, the relationships between the state/state-action step, generation, statistic, and value functions are shown in the following diagram:
\begin{equation}
\begin{tikzcd}[column sep=0em, row sep=0em]
&[1em]
\singleton + \Rewards \times \States
\arrow[rr, unique morphism]
\arrow[rd]
\arrow[rrrd]
&&[1em]
\singleton + \Rewards \times \List{\Rewards}
\arrow[rd, unique morphism]
\arrow[dd, "\bracks{\nil, \cons}" description, pos=.2]
\\[6em]
&&
\singleton + \Rewards \times (\StateActions)
\arrow[ru, unique morphism]
\arrow[rr]
&&[1em]
\singleton + \Rewards \times \Statistics
\arrow[dd, "\bracks{\init, \update}" description, pos=.2]
\\[3em]
\singleton
\arrow[r, "s_0"]
&
\States
\arrow[uu, "\step^\States_\ptro" description, pos=.8]
\arrow[rr, "\gen^\States_\ptro" description, pos=.75, unique morphism]
\arrow[rd, "\pair" description]
\arrow[rrrd, "\statistic^\States_\policy" description, pos=.75]
\arrow[rrrrd, "\svalue_\policy", bend right=60, bezier bounding box]
&&
\List{\Rewards}
\arrow[rd, "\aggiu" description, unique morphism]
\\[6em]
&&
\StateActions
\arrow[uu, "\step^\StateActions_\ptro" description, pos=.8]
\arrow[ru, "\gen^\StateActions_\ptro" description, pos=.75, unique morphism]
\arrow[rr, "\statistic^\StateActions_\policy" description]
\arrow[rrr, "\qvalue_\policy", bend right=30, looseness=.5, xshift=-.4em]
&&
|[inner xsep=1em]| \Statistics
\arrow[r, "\post"]
&[1em]
\Rewards
\end{tikzcd}
\end{equation}


\subsection{Advantage function}

The \emph{advantage function} \citep{baird1994reinforcement},
\begin{equation}
\advantage_\policy: \StateActions \to \Rewards
\defeq
\qvalue_\policy - \svalue_\policy \compL p_1
=
\anon{s, a}{\qvalue_\policy(s, a) - \svalue_\policy(s)},
\end{equation}
is defined as the difference between the state-action value function $\qvalue_\policy: \StateActions \to \Rewards$ and the state value function $\svalue_\policy: \States \to \Rewards$, where $p_1: \StateActions \to \States$ is the projection function that extracts the state from a state-action pair.
The advantage function measures the advantage of taking an action $a$ in a state $s$ over the average value of all actions in that state following the policy $\policy$, which is widely used in RL algorithms.

For a general recursive reward aggregation function ${\post} \compL {\aggiu}$, the advantage function can be expressed using the state-action statistic function $\statistic^\StateActions_\policy: \StateActions \to \Statistics$ and the state statistic function $\statistic^\States_\policy: \States \to \Statistics$ as follows:
\begin{align}
\advantage_\policy: \StateActions \to \Rewards
={}&
\anon{s, a}{\post(\statistic^\StateActions_\policy(s, a)) - \post(\statistic^\States_\policy(s))}
\\
={}&
\anon{s, a}{\scases{0}{
\post(\reward(s, a) \update \statistic^\States_\policy(\transition(s, a))) - \post(\statistic^\States_\policy(s))
}}.
\end{align}

Because the statistic function can be computed recursively, given a sequence of states, rewards, and statistics, we can obtain a sequence of advantage estimators:
\begin{align}
\hat{\advantage}^{(1)}_t
&=
\post(r_t \update \statistic_{t+1}) - \post(\statistic_t),
\\
\hat{\advantage}^{(2)}_t
&=
\post(r_t \update r_{t+1} \update \statistic_{t+2}) - \post(\statistic_t),
\\
\hat{\advantage}^{(3)}_t
&=
\post(r_t \update r_{t+1} \update r_{t+2} \update \statistic_{t+3}) - \post(\statistic_t),
\\
\nonumber
\vdots
\end{align}

\begin{align}
\hat{\advantage}^{(1)}_{t+1}
&=
\post(r_{t+1} \update \statistic_{t+2}) - \post(\statistic_{t+1}),
\\
\hat{\advantage}^{(2)}_{t+1}
&=
\post(r_{t+1} \update r_{t+2} \update \statistic_{t+3}) - \post(\statistic_{t+1}),
\\
\nonumber
\vdots
\end{align}

The \emph{generalized advantage estimator} (GAE) proposed by \citet{schulman2016high} combines these advantage estimators with a discount factor $\lambda \in [0, 1]$:
\begin{align}
\label{eq:gae}
\hat{\advantage}_t
\defeq{}&
  \hat{\advantage}^{(1)}_t
+ \lambda \hat{\advantage}^{(2)}_t
+ \lambda^2 \hat{\advantage}^{(3)}_t
+ \cdots
\\
={}&
\setlength\arraycolsep{2pt}
\begin{array}[t]{c l l cl cl cl}
  & 1 & (\post(r_t &
&&
\update & \statistic_{t+1}) &
- & \post(\statistic_t))
\\
+ & \lambda & (\post(r_t &
\update & r_{t+1} & 
\update & \statistic_{t+2}) &
- & \post(\statistic_t))
\\
+ & \lambda^2 & (\post(r_t &
\update & r_{t+1} \update r_{t+2} &
\update & \statistic_{t+3}) &
- & \post(\statistic_t))
\\
+ & \cdots
\end{array}
\end{align}

The original GAE formulation \citep{schulman2016high} considered only the discounted sum and an infinite horizon.
For a finite horizon $\horizon$, the advantage estimator can be expressed as follows:
\begin{align}
\advantage_t
=&
\setlength\arraycolsep{2pt}
\begin{array}[t]{c l l cl cl cl cl cl cl}
  & 1 & (r_t &
&&&&&&&&
+ & \gamma v_{t+1} &
- & v_t)
\\
+ & \lambda & (r_t &
+ & \gamma r_{t+1} &
&&&&&&
+ & \gamma^2 v_{t+2} &
- & v_t)
\\
+ & \lambda^2 & (r_t &
+ & \gamma r_{t+1} &
+ & \gamma^2 r_{t+2} &
&&&&
+ & \gamma^3 v_{t+3} &
- & v_t)
\\
+ & \cdots
\\
+ & \lambda^{\horizon-t-1} & (r_t &
+ & \gamma r_{t+1} &
+ & \gamma^2 r_{t+2} &
+ & \cdots &
+ & \gamma^{\horizon-t-1} r_{\horizon-1} &
+ & \gamma^{\horizon-t} v_{\horizon} &
- & v_t)
\end{array}
\\
={}&
\sum_{i=0}^{\horizon-t-1}
\lambda^i \gamma^i \parens*{
\frac{1 - \lambda^{\horizon-t-i}}{1 - \lambda} r_{t+i}
+ \gamma v_{t+i+1}
}
- \frac{1 - \lambda^{\horizon-t}}{1 - \lambda} v_t,
\end{align}
which has a recursive form:
\begin{equation}
\advantage_t
=
\frac{1 - \lambda^{\horizon-t}}{1 - \lambda} \parens*{r_t + \gamma v_{t+1} - v_t} + \lambda \gamma \advantage_{t+1}.
\end{equation}

However, when considering a general recursive reward aggregation function ${\post} \compL {\aggiu}$, a recursive expression for the advantage estimator is not always available.
Therefore, the advantage estimator may need to be computed directly using its original definition in \cref{eq:gae}.

\clearpage

\begin{figure}
\centering
\begin{adjustbox}{scale=.95}
\begin{tikzpicture}
\coordinate (i) at (0, 0);
\coordinate (r) at (3, 0);
\coordinate (j) at (6, 0);
\coordinate (d) at (8, 0);
\coordinate (k) at (10, 0);

\coordinate (s) at (0, 4);
\coordinate (t) at (0, 0);

\node (Si) at (i |- s) {$\States$};
\node (Ti) at (i |- t) {$\Statistics$};
\node (R) at (r |-, 1) {$\Rewards$};
\node (Sj) at (j |- s) {$\States$};
\node (Tj) at (j |- t) {$\Statistics$};
\node (Sd) at (d |- s) {$\dots$};
\node (Td) at (d |- t) {$\dots$};
\node (Sk) at (k |- s) {$\singleton$};
\node (Tk) at (k |- t) {$\Statistics$};

\node (copyS) [diagonal] at (2, |- s) {};
\node (copySa) [diagonal] at (2, 3) {};
\node (copyA) [diagonal] at (4, 3) {};
\draw (Si) [->] to (copyS);
\draw (copyS) [->] to (copySa);

\node (policy) [morphism, minimum width=.6cm, minimum height=.6cm] at (3, 3) {$\policy$};
\draw (copySa) [->] to (policy);
\draw (policy) [->] to (copyA);

\node (transition) [morphism, minimum width=.6cm, minimum height=.6cm] at (4, |- s) {$\transition$};
\draw (copyS) [->] to (transition);
\draw (copyA) [->] to (transition);
\draw (transition) [->] to (Sj);

\node (reward) [morphism, minimum width=2.6cm, minimum height=.6cm] at (r |-, 2) {$\reward$};
\draw (copySa) [->] to (copySa |- reward.north);
\draw (copyA) [->] to (copyA |- reward.north);
\draw (reward) [->] to (R);

\node (update) [morphism, minimum width=.6cm, minimum height=.6cm] at (r |- t) {$\update$};
\draw (R) [->] to (update);
\draw (Tj) [->] to (update);
\draw (update) [->] to (Ti);

\draw (Sj) [->] to (Sd);
\draw (Sd) [->] to (Sk);
\draw (Tk) [->] to (Td);
\draw (Td) [->] to (Tj);

\node (init) [morphism, minimum width=1cm, minimum height=.6cm] at ($(Sk)!0.5!(Tk)$) {$\init$};
\draw (Sk) [->] to (init);
\draw (init) [->] to (Tk);

\node [draw, rectangle, dashed, minimum width=3cm, minimum height=3cm] at (3, 3) {};
\node [draw, rectangle, dashed, minimum width=3cm, minimum height=1cm] at (3, 0) {};
\node [draw, rectangle, dashed, minimum width=1.5cm, minimum height=5cm] at (init) {};

\node [right=0 of Si.north east] {$s_t$};
\node [right=0 of Sj.north east] {$s_{t+1}$};
\node [right=0 of Ti.north east] {$\tau_t$};
\node [right=0 of Tj.north east] {$\tau_{t+1}$};
\node [right=0 of R.north] {$r_{t+1}$};

\end{tikzpicture}
\end{adjustbox}
\caption{State statistic bidirectional process $\statistic^\States_\policy: \States \to \Statistics$}
\label{fig:optic_state_statistic}
\end{figure}

\begin{figure}
\centering
\begin{adjustbox}{scale=.95}
\begin{tikzpicture}
\coordinate (i) at (0, 0);
\coordinate (r) at (2.5, 0);
\coordinate (j) at (5, 0);
\coordinate (d) at (8, 0);
\coordinate (k) at (10, 0);

\coordinate (s) at (0, 5);
\coordinate (t) at (0, 0);

\node (Si) at (i |- s) {$\States$};
\node (Ti) at (i |- t) {$\Statistics$};
\node (R) at (r |-, 1) {$\Rewards$};
\node (Sj) at (j |- s) {$\States$};
\node (Tj) at (j |- t) {$\Statistics$};
\node (Sd) at (d |- s) {$\dots$};
\node (Td) at (d |- t) {$\dots$};
\node (Sk) at (k |- s) {$\singleton$};
\node (Tk) at (k |- t) {$\Statistics$};

\node (copyS) [diagonal] at (2, |- s) {};
\node (copySt) [diagonal] at (2, 4) {};
\node (copySr) [diagonal] at (2, 3) {};
\draw (Si) [->] to (copyS);
\draw (copyS) [->] to (copySt);
\draw (copyS) [->] to (copySr);

\node (transition_policy) [morphism, minimum width=.6cm, minimum height=.6cm] at (3, 4) {$\policy$};
\draw (copySt) [->] to (transition_policy);

\node (transition) [morphism, minimum width=.6cm, minimum height=.6cm] at (3, |- s) {$\transition$};
\draw (copyS) [->] to (transition);
\draw (transition_policy) [->] to (transition);
\draw (transition) [->] to (Sj);

\node (reward_policy) [morphism, minimum width=.6cm, minimum height=.6cm] at (3, 3) {$\policy_\theta$};
\draw (copySr) [->] to (reward_policy);

\node (reward) [morphism, minimum width=1.6cm, minimum height=.6cm] at (r |-, 2) {$\reward$};
\draw (copySr) [->] to (copySr |- reward.north);
\draw (reward_policy) [->] to (reward_policy |- reward.north);
\draw (reward) [->] to (R);

\node (update) [morphism, minimum width=.6cm, minimum height=.6cm] at (r |- t) {$\update$};
\draw (R) [->] to (update);
\draw (Tj) [->] to (update);
\draw (update) [->] to (Ti);

\draw (Sj) [->] to (Sd);
\draw (Sd) [->] to (Sk);
\draw (Tk) [->] to (Td);
\draw (Td) [->] to (Tj);

\node (init) [morphism, minimum width=1cm, minimum height=.6cm] at ($(Sk)!0.5!(Tk)$) {$\init$};
\draw (Sk) [->] to (init);
\draw (init) [->] to (Tk);

\node [draw, rectangle, dashed, minimum width=2cm, minimum height=4cm] at (2.5, 3.5) {};
\node [draw, rectangle, dashed, minimum width=2cm, minimum height=1cm] at (2.5, 0) {};
\node [draw, rectangle, dashed, minimum width=1.5cm, minimum height=6cm] at (init) {};

\node [right=0 of Si.north east] {$s_t$};
\node [right=0 of Sj.north east] {$s_{t+1}$};
\node [right=0 of Ti.north east] {$\tau_t$};
\node [right=0 of Tj.north east] {$\tau_{t+1}$};
\node [right=0 of R.north] {$r_{t+1}$};

\end{tikzpicture}
\end{adjustbox}
\caption{State statistic bidirectional process (with different behavior and target policies)}
\label{fig:optic_state_statistic_policy}
\end{figure}

\begin{figure}
\centering
\begin{adjustbox}{scale=.95}
\begin{tikzpicture}
\coordinate (i) at (0, 0);
\coordinate (r) at (2.5, 0);
\coordinate (j) at (5, 0);
\coordinate (d) at (8, 0);
\coordinate (k) at (10, 0);

\coordinate (s) at (0, 5);
\coordinate (t) at (0, 0);

\node (Si) at (i |- s) {$\States$};
\node (Ti) at (i |- t) {$\Statistics$};
\node (Sr) at (2, 3) {$\States$};
\node (Sj) at (j |- s) {$\States$};
\node (Tj) at (j |- t) {$\Statistics$};
\node (Sd) at (d |- s) {$\dots$};
\node (Td) at (d |- t) {$\dots$};
\node (Sk) at (k |- s) {$\singleton$};
\node (Tk) at (k |- t) {$\Statistics$};

\node (copyS) [diagonal] at (2, |- s) {};
\node (copySt) [diagonal] at (2, 4) {};
\node (copySr) [diagonal] at (2, 2) {};
\draw (Si) [->] to (copyS);
\draw (copyS) [->] to (copySt);
\draw (copyS) [->] to (Sr);
\draw (Sr) [->] to (copySr);

\node (transition_policy) [morphism, minimum width=.6cm, minimum height=.6cm] at (3, 4) {$\policy$};
\draw (copySt) [->] to (transition_policy);

\node (transition) [morphism, minimum width=.6cm, minimum height=.6cm] at (3, |- s) {$\transition$};
\draw (copyS) [->] to (transition);
\draw (transition_policy) [->] to (transition);
\draw (transition) [->] to (Sj);

\node (reward_policy) [morphism, minimum width=.6cm, minimum height=.6cm] at (3, 2) {$\policy_\theta$};
\draw (copySr) [->] to (reward_policy);

\node (reward) [morphism, minimum width=1.6cm, minimum height=.6cm] at (r |-, 1) {$\reward$};
\draw (copySr) [->] to (copySr |- reward.north);
\draw (reward_policy) [->] to (reward_policy |- reward.north);

\node (update) [morphism, minimum width=.6cm, minimum height=.6cm] at (r |- t) {$\update$};
\draw (reward) [->] to (update);
\draw (Tj) [->] to (update);
\draw (update) [->] to (Ti);

\draw (Sj) [->] to (Sd);
\draw (Sd) [->] to (Sk);
\draw (Tk) [->] to (Td);
\draw (Td) [->] to (Tj);

\node (init) [morphism, minimum width=1cm, minimum height=.6cm] at ($(Sk)!0.5!(Tk)$) {$\init$};
\draw (Sk) [->] to (init);
\draw (init) [->] to (Tk);

\node [draw, rectangle, dashed, minimum width=2cm, minimum height=2cm] at (2.5, 4.5) {};
\node [draw, rectangle, dashed, minimum width=2cm, minimum height=3cm] at (2.5, 1) {};
\node [draw, rectangle, dashed, minimum width=1.5cm, minimum height=6cm] at (init) {};

\node [right=0 of Si.north east] {$s_t$};
\node [right=0 of Sj.north east] {$s_{t+1}$};
\node [right=0 of Ti.north east] {$\tau_t$};
\node [right=0 of Tj.north east] {$\tau_{t+1}$};
\node [right=0 of Sr.north] {$s_t$};

\end{tikzpicture}
\end{adjustbox}
\caption{State statistic bidirectional process (with state as the residual)}
\label{fig:optic_state_statistic_trajectory}
\end{figure}


\clearpage

\begin{figure}
\centering
\begin{adjustbox}{scale=.95}
\begin{tikzpicture}
\coordinate (i) at (0, 0);
\coordinate (r) at (2.5, 0);
\coordinate (j) at (6, 0);
\coordinate (d) at (8, 0);
\coordinate (k) at (10, 0);

\coordinate (s) at (0, 4);
\coordinate (a) at (0, 3);
\coordinate (t) at (0, 0);

\node (Si) at (i |- s) {$\States$};
\node (Ai) at (i |- a) {$\Actions$};
\node (Ti) at (i |- t) {$\Statistics$};
\node (R) at (r |-, 1) {$\Rewards$};
\node (Sj) at (j |- s) {$\States$};
\node (Aj) at (j |- a) {$\Actions$};
\node (Tj) at (j |- t) {$\Statistics$};
\node (Sd) at (d |- s) {$\dots$};
\node (Ad) at (d |- a) {$\dots$};
\node (Td) at (d |- t) {$\dots$};
\node (Sk) at (k |- s) {$\singleton$};
\node (Tk) at (k |- t) {$\Statistics$};

\node (copyS)  [diagonal] at (2, |- s) {};
\node (copyA)  [diagonal] at (3, |- a) {};
\node (copySt) [diagonal] at (4, |- s) {};
\draw (Si) [->] to (copyS);
\draw (Ai) [->] to (copyA);
\draw (copySt) [->] to (Sj);

\node (transition) [morphism, minimum width=.6cm, minimum height=.6cm] at (3, |- s) {$\transition$};
\node (policy) [morphism, minimum width=.6cm, minimum height=.6cm] at (4, |- a) {$\policy$};
\draw (copyS) [->] to (transition);
\draw (copyA) [->] to (transition);
\draw (transition) [->] to (copySt);
\draw (copySt) [->] to (policy);
\draw (policy) [->] to (Aj);

\node (reward) [morphism, minimum width=1.6cm, minimum height=.6cm] at (r |-, 2) {$\reward$};
\draw (copyS) [->] to (copyS |- reward.north);
\draw (copyA) [->] to (copyA |- reward.north);
\draw (reward) [->] to (R);

\node (update) [morphism, minimum width=.6cm, minimum height=.6cm] at (r |- t) {$\update$};
\draw (R) [->] to (update);
\draw (Tj) [->] to (update);
\draw (update) [->] to (Ti);

\draw (Sj) [->] to (Sd);
\draw (Aj) [->] to (Ad);
\draw (Sd) [->] to (Sk);
\draw (Tk) [->] to (Td);
\draw (Td) [->] to (Tj);

\node (init) [morphism, minimum width=1cm, minimum height=.6cm] at ($(Sk)!0.5!(Tk)$) {$\init$};
\draw (Sk) [->] to (init);
\draw (init) [->] to (Tk);

\node [draw, rectangle, dashed, minimum width=3cm, minimum height=3cm] at (3, 3) {};
\node [draw, rectangle, dashed, minimum width=3cm, minimum height=1cm] at (3, 0) {};
\node [draw, rectangle, dashed, minimum width=1.5cm, minimum height=5cm] at (init) {};

\node [right=0 of Si.north east] {$s_t$};
\node [right=0 of Sj.north east] {$s_{t+1}$};
\node [right=0 of Ai.north east] {$a_t$};
\node [right=0 of Aj.north east] {$a_{t+1}$};
\node [right=0 of Ti.north east] {$\tau_t$};
\node [right=0 of Tj.north east] {$\tau_{t+1}$};
\node [right=0 of R.north] {$r_{t+1}$};

\end{tikzpicture}
\end{adjustbox}
\caption{State-action statistic bidirectional process $\statistic^\StateActions_\policy: \StateActions \to \Statistics$}
\label{fig:optic_state-action_statistic}
\end{figure}

\begin{figure}
\centering
\begin{adjustbox}{scale=.95}
\begin{tikzpicture}
\coordinate (i) at (0, 0);
\coordinate (r) at (2.5, 0);
\coordinate (j) at (6, 0);
\coordinate (d) at (8, 0);
\coordinate (k) at (10, 0);

\coordinate (s) at (0, 5);
\coordinate (a) at (0, 4);
\coordinate (t) at (0, 0);

\node (Si) at (i |- s) {$\States$};
\node (Ai) at (i |- a) {$\Actions$};
\node (Ti) at (i |- t) {$\Statistics$};
\node (R) at (r |-, 1) {$\Rewards$};
\node (Sj) at (j |- s) {$\States$};
\node (Aj) at (j |- a) {$\Actions$};
\node (Tj) at (j |- t) {$\Statistics$};
\node (Sd) at (d |- s) {$\dots$};
\node (Ad) at (d |- a) {$\dots$};
\node (Td) at (d |- t) {$\dots$};
\node (Sk) at (k |- s) {$\singleton$};
\node (Tk) at (k |- t) {$\Statistics$};

\node (copyS)  [diagonal] at (2, |- s) {};
\node (copySt) [diagonal] at (4, |- s) {};
\node (copySr)  [diagonal] at (2, 3) {};
\draw (Si) [->] to (copyS);
\draw (copyS) [->] to (copySr);
\draw (copySt) [->] to (Sj);

\node (transition) [morphism, minimum width=.6cm, minimum height=.6cm] at (3, |- s) {$\transition$};
\draw (copyS) [->] to (transition);
\draw (Ai) [->] to (3, |- a) to (transition);
\draw (transition) [->] to (copySt);

\node (transition_policy) [morphism, minimum width=.6cm, minimum height=.6cm] at (4, |- a) {$\policy$};
\draw (copySt) [->] to (transition_policy);
\draw (transition_policy) [->] to (Aj);

\node (reward_policy) [morphism, minimum width=.6cm, minimum height=.6cm] at (3, 3) {$\policy_\theta$};
\draw (copySr) [->] to (reward_policy);

\node (reward) [morphism, minimum width=1.6cm, minimum height=.6cm] at (r |-, 2) {$\reward$};
\draw (copySr) [->] to (copySr |- reward.north);
\draw (reward_policy) [->] to (reward_policy |- reward.north);
\draw (reward) [->] to (R);

\node (update) [morphism, minimum width=.6cm, minimum height=.6cm] at (r |- t) {$\update$};
\draw (R) [->] to (update);
\draw (Tj) [->] to (update);
\draw (update) [->] to (Ti);

\draw (Sj) [->] to (Sd);
\draw (Aj) [->] to (Ad);
\draw (Sd) [->] to (Sk);
\draw (Tk) [->] to (Td);
\draw (Td) [->] to (Tj);

\node (init) [morphism, minimum width=1cm, minimum height=.6cm] at ($(Sk)!0.5!(Tk)$) {$\init$};
\draw (Sk) [->] to (init);
\draw (init) [->] to (Tk);

\node [draw, rectangle, dashed, minimum width=3cm, minimum height=4cm] at (3, 3.5) {};
\node [draw, rectangle, dashed, minimum width=3cm, minimum height=1cm] at (3, 0) {};
\node [draw, rectangle, dashed, minimum width=1.5cm, minimum height=6cm] at (init) {};

\node [right=0 of Si.north east] {$s_t$};
\node [right=0 of Sj.north east] {$s_{t+1}$};
\node [right=0 of Ai.north east] {$a_t$};
\node [right=0 of Aj.north east] {$a_{t+1}$};
\node [right=0 of Ti.north east] {$\tau_t$};
\node [right=0 of Tj.north east] {$\tau_{t+1}$};
\node [right=0 of R.north] {$r_{t+1}$};

\end{tikzpicture}
\end{adjustbox}
\caption{State-action statistic bidirectional process (with different behavior and target policies)}
\label{fig:optic_state-action_statistic_policy}
\end{figure}

\begin{figure}
\centering
\begin{adjustbox}{scale=.95}
\begin{tikzpicture}
\coordinate (i) at (0, 0);
\coordinate (r) at (2.5, 0);
\coordinate (j) at (6, 0);
\coordinate (d) at (8, 0);
\coordinate (k) at (10, 0);

\coordinate (s) at (0, 6);
\coordinate (a) at (0, 5);
\coordinate (t) at (0, 0);

\node (Si) at (i |- s) {$\States$};
\node (Ai) at (i |- a) {$\Actions$};
\node (Ti) at (i |- t) {$\Statistics$};
\node (Sr) at (2, 4) {$\States$};
\node (Ar) at (3, 4) {$\Actions$};
\node (Sj) at (j |- s) {$\States$};
\node (Aj) at (j |- a) {$\Actions$};
\node (Tj) at (j |- t) {$\Statistics$};
\node (Sd) at (d |- s) {$\dots$};
\node (Ad) at (d |- a) {$\dots$};
\node (Td) at (d |- t) {$\dots$};
\node (Sk) at (k |- s) {$\singleton$};
\node (Tk) at (k |- t) {$\Statistics$};

\node (copyS)  [diagonal] at (2, |- s) {};
\node (copyA)  [diagonal] at (3, |- a) {};
\node (copySt) [diagonal] at (4, |- s) {};
\node (copySr)  [diagonal] at (2, 2) {};
\node (delAr)  [diagonal] at (3, 3) {};
\draw (Si) [->] to (copyS);
\draw (copyS) [->] to (Sr);
\draw (copyA) [->] to (Ar);
\draw (copySt) [->] to (Sj);
\draw (Sr) [->] to (copySr);
\draw (Ai) [->] to (copyA);
\draw (Ar) [->] to (delAr);

\node (transition) [morphism, minimum width=.6cm, minimum height=.6cm] at (3, |- s) {$\transition$};
\draw (copyS) [->] to (transition);
\draw (copyA) [->] to (transition);
\draw (transition) [->] to (copySt);

\node (transition_policy) [morphism, minimum width=.6cm, minimum height=.6cm] at (4, |- a) {$\policy$};
\draw (copySt) [->] to (transition_policy);
\draw (transition_policy) [->] to (Aj);

\node (reward_policy) [morphism, minimum width=.6cm, minimum height=.6cm] at (3, 2) {$\policy_\theta$};
\draw (copySr) [->] to (reward_policy);

\node (reward) [morphism, minimum width=1.6cm, minimum height=.6cm] at (r |-, 1) {$\reward$};
\draw (copySr) [->] to (copySr |- reward.north);
\draw (reward_policy) [->] to (reward_policy |- reward.north);

\node (update) [morphism, minimum width=.6cm, minimum height=.6cm] at (r |- t) {$\update$};
\draw (reward) [->] to (update);
\draw (Tj) [->] to (update);
\draw (update) [->] to (Ti);

\draw (Sj) [->] to (Sd);
\draw (Aj) [->] to (Ad);
\draw (Sd) [->] to (Sk);
\draw (Tk) [->] to (Td);
\draw (Td) [->] to (Tj);

\node (init) [morphism, minimum width=1cm, minimum height=.6cm] at ($(Sk)!0.5!(Tk)$) {$\init$};
\draw (Sk) [->] to (init);
\draw (init) [->] to (Tk);

\node [draw, rectangle, dashed, minimum width=3cm, minimum height=2cm] at (3, 5.5) {};
\node [draw, rectangle, dashed, minimum width=3cm, minimum height=4cm] at (3, 1.5) {};
\node [draw, rectangle, dashed, minimum width=1.5cm, minimum height=7cm] at (init) {};

\node [right=0 of Si.north east] {$s_t$};
\node [right=0 of Sj.north east] {$s_{t+1}$};
\node [right=0 of Ai.north east] {$a_t$};
\node [right=0 of Aj.north east] {$a_{t+1}$};
\node [right=0 of Ti.north east] {$\tau_t$};
\node [right=0 of Tj.north east] {$\tau_{t+1}$};
\node [right=0 of Sr.north] {$s_t$};
\node [right=0 of Ar.north] {$a_t$};

\end{tikzpicture}
\end{adjustbox}
\caption{State-action statistic bidirectional process (with state-action as the residual)}
\label{fig:optic_state-action_statistic_trajectory}
\end{figure}
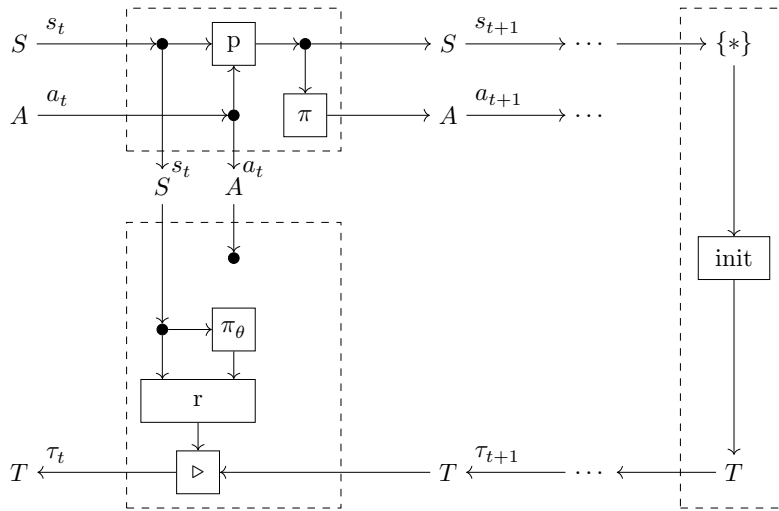

\clearpage
\section{Algebraic structures in Markov decision process}
\label{app:algebra}
In this section, we briefly discuss the algebraic structures used in this work.
For a tutorial on algebraic programming, we refer the reader to \citet{hutton1999tutorial}.
For a theoretical treatment of algebra fusion, see \citet{hinze2010theory}.
For an accessible and illustrative introduction to bidirectional processes, we recommend \citet{gavranovic2022space}.


\subsection{Algebra fusion}

In this work, we mainly considered algebras and coalgebras of signature $\singleton + \Rewards \times (-)$, i.e., lists of rewards.
An \emph{algebra} is a pair $(A, f)$ consisting of a carrier set $A$ and a function $f: \singleton + \Rewards \times A \to A$.
A \emph{coalgebra} is a pair $(C, g)$ consisting of a carrier set $C$ and a function $g: C \to \singleton + \Rewards \times C$.
For example, the list construction $\bracks{\nil, \cons}: \singleton + \Rewards \times \List{\Rewards} \to \List{\Rewards}$ is an algebra on the set $\List{\Rewards}$ of lists of rewards, while the step function $\step^\States_\ptro: \States \to \singleton + \Rewards \times \States$ is a coalgebra on the set $\States$ of states.

Note that the list construction $\bracks{\nil, \cons}$ is the \emph{initial algebra}, the discounted sum function $\dsum$ is defined as the \emph{catamorphism} from the initial algebra to the algebra $[0, \dadd]$, while the recursive reward generation function $\gen_\ptro$ is defined as the \emph{hylomorphism} from the coalgebra $\step_\ptro$ to the initial algebra.
In the field of functional programming, such operations are also known as \texttt{fold} and \texttt{unfold} \citep{meijer1991functional, bird1997algebra, hutton1999tutorial,yang2022fantastic}.

Due to the recursive nature of the generation and aggregation functions, we can derive the recursive structure of their composition using the algebra fusion technique \citep{hinze2010theory}, which leads to the Bellman equations for the state statistic function $\statistic^\States_\policy: \States \to \Statistics$ in \cref{thm:bellman_equation_state} and the state-action statistic function $\statistic^\StateActions_\policy: \StateActions \to \Statistics$ in \cref{thm:bellman_equation_state-action}.


\subsection{Bidirectional process}

In \cref{fig:optic_state_value}, we illustrate the bidirectional processes for the state statistic function and state value function.
In algebra, such bidirectional processes are called \emph{lenses} and \emph{optics} \citep{riley2018categories}.
Such bidirectional processes \citep{riley2018categories} have been applied to study supervised learning \citep{fong2019lenses}, Bayesian inference \citep{smithe2020bayesian}, gradient-based learning \citep{cruttwell2022categorical}, and reinforcement learning \citep{hedges2022value}.

Note that there is a slight difference between the definitions of step/generation/statistic functions in \cref{eq:state_step,eq:state_generation,eq:state_statistic} and the bidirectional process in \cref{fig:optic_state_value} (reproduced in \cref{fig:optic_state_statistic}).
In \cref{eq:state_step}, a state $s$ is duplicated and passed separately to the transition function $\transition_\policy$ and the reward function $\reward_\policy$, requiring the policy $\policy$ to compute the action $a$ twice.
In contrast, in \cref{fig:optic_state_statistic}, the state $s$ is passed to the policy $\policy$ only once, and the action $a$ is computed only once and then copied to the transition function $\transition$ and the reward function $\reward$.
These two approaches are equivalent only when the following equation holds:
\begin{equation}
\begin{tikzpicture}
\node (A) [matrix] {
  \node (S) at (1, 0) {$\States$};
  \node (A1) at (0, 3) {$\Actions$};
  \node (A2) at (2, 3) {$\Actions$};
  \node (policy1) [morphism] at (0, 2) {$\policy$};
  \node (policy2) [morphism] at (2, 2) {$\policy$};
  \node (copy) [diagonal] at (1, 1) {};
  \draw (S) -- (copy);
  \draw (copy) [out=180, in=-90] to (policy1) -- (A1);
  \draw (copy) [out=0, in=-90] to (policy2) -- (A2);
  \\
};
\node (B) [matrix, right=3cm of A, anchor=center] {
  \node (S) at (1, 0) {$\States$};
  \node (A1) at (0, 3) {$\Actions$};
  \node (A2) at (2, 3) {$\Actions$};
  \node (policy) [morphism] at (1, 1) {$\policy$};
  \node (copy) [diagonal] at (1, 2) {};
  \draw (S) -- (policy) -- (copy);
  \draw (copy) [out=180, in=-90] to (A1);
  \draw (copy) [out=0, in=-90] to (A2);
  \\
};
\node at ($(A)!0.5!(B)$) {$=$};
\end{tikzpicture}
\end{equation}

For functions, copying an input and then passing the copies to two identical functions is equivalent to passing the input to the function once and then copying the output.
However, for stochastic functions, these two approaches are not equivalent, which requires additional care when defining bidirectional processes for stochastic functions \citep[see also][Definition~10.1]{fritz2020synthetic}.

Strictly speaking, the definitions in \cref{eq:state_step,eq:state_generation,eq:state_statistic} correspond to a bidirectional process illustrated in \cref{fig:optic_state_statistic_policy}, where different behavior and target policies can be considered.
In this setting, the target policy $\policy_\theta$, parameterized by $\theta$, is used to compute the reward and is optimized, while the potentially unknown behavior policy $\policy$ is passed to the transition function.
Further, the \emph{internal state} between the forward and backward processes --- also known as the \emph{residual} \citep{gavranovic2022space} --- can be the state itself rather than the reward, as shown in \cref{fig:optic_state_statistic_trajectory}.
Similar considerations extend to the state-action statistic function, as illustrated in \cref{fig:optic_state-action_statistic,fig:optic_state-action_statistic_policy,fig:optic_state-action_statistic_trajectory}.

We believe that such bidirectional processes offer a clearer framework for reinforcement learning, including offline reinforcement learning, inverse reinforcement learning, and imitation learning \citep{hussein2017imitation, arora2021survey, hedges2022value, murphy2024reinforcement}.
Further research is needed to explore the full potential of bidirectional processes in reinforcement learning.


\subsection{Non-uniqueness of update function and post-processing function}

It is important to note that for a given aggregation function, the corresponding update function $\update: \Rewards \times \Statistics \to \Statistics$ and post-processing function $\post: \Statistics \to \Rewards$ are not necessarily unique.


\paragraph{Mean}

For example, the mean function can be computed recursively in different ways: one approach updates the sum and the length, while another updates the mean and the length.
Each approach has its own advantages and disadvantages.
Updating the sum allows for a straightforward implementation, but when both the sum and the length are large, numerical instability may arise.
In contrast, updating the mean may require additional computation, but if the rewards are bounded, the mean remains bounded as well, which can improve numerical stability.


\paragraph{Variance}

Similarly, the variance can also be computed recursively through multiple formulations.
A common method maintains the sum of squares, the mean, and the length, while a more numerically stable alternative, Welford's algorithm \citep{welford1962note}, updates the variance directly using incremental differences.
Specifically, the update rule is given by:
\begin{equation}
\sigma_{t+1}^2
=
\sigma_{t}^2 + \frac{ t (r_{t+1} - \mu_{t})^2 - (t + 1)\sigma_{t}^2}{(t + 1)^2},
\end{equation}
where $r_t$, $\mu_t$, and $\sigma_t^2$ denote the reward observed at time step $t$, the mean of the rewards up to time $t$, and the variance of the rewards up to time $t$, respectively.
To compute the variance iteratively using this formulation, it is sufficient to maintain and update the length, the mean, and the variance at each step.
This formulation improves numerical stability by preventing catastrophic cancellation \citep{goldberg1991every, muller2018handbook}, which occurs when subtracting two large and nearly identical values, leading to significant precision loss in floating-point arithmetic.

\clearpage
\section{Metrics and Bellman operators}
\label{app:metric}
In this section, we discuss the \emph{metrics} on the statistics $\Statistics$ and rewards $\Rewards$ and the \emph{Bellman operators} for the state/state-action statistic functions.


\begin{table}
\centering
\caption{Properties of metrics}
\label{tab:metric}
\begin{tabular}{cccc}
\toprule
& Premetric & Strict premetric & Metric
\\
\midrule
\begin{tabular}{@{}c@{}}
Indiscernibility of identities\\
$(a_1 = a_2) \limp (d_A(a_1, a_2) = 0)$
\end{tabular}
& \checkmark & \checkmark & \checkmark
\\
\midrule
\begin{tabular}{@{}c@{}}
Identity of indiscernibles\\
$(d_A(a_1, a_2) = 0) \limp (a_1 = a_2)$
\end{tabular}
& & \checkmark & \checkmark
\\
\midrule
\begin{tabular}{@{}c@{}}
Symmetry\\
$d_A(a_1, a_2) = d_A(a_2, a_1)$
\end{tabular}
& & & \checkmark
\\
\midrule
\begin{tabular}{@{}c@{}}
Triangle inequality\\
$d_A(a_1, a_3) \leq d_A(a_1, a_2) + d_A(a_2, a_3)$
\end{tabular}
& & & \checkmark
\\
\bottomrule
\end{tabular}
\end{table}


\subsection{Preliminaries}

Recall the definitions of metrics, as summarized in \cref{tab:metric}:
\begin{definition}[Premetric]
A \emph{premetric} on a set $A$ is a function $d_A: A \times A \to \quant$ such that $\forall a \in A.\; d_A(a, a) = 0$.
\end{definition}

\begin{definition}[Strict premetric]
A \emph{strict premetric} on a set $A$ is a function $d_A: A \times A \to \quant$ such that $\forall a_1, a_2 \in A.\; (d_A(a_1, a_2) = 0) \leqv (a_1 = a_2)$.
\end{definition}

Given a function to a premetric space, we can define a premetric on the domain by pullback:
\begin{restatable}[Pullback premetric]{lemma}{PullbackPremetric}
\label{lem:pullback_premetric}
Let $d_B: B \times B \to \quant$ be a premetric on a set $B$, and let $f: A \to B$ be a function.
The pullback premetric $d_A: A \times A \to \quant$ is defined by
\begin{equation}
\forall a_1, a_2 \in A.\;
d_A(a_1, a_2)
\defeq
d_B(f(a_1), f(a_2)).
\end{equation}
If $d_B$ is a strict premetric, then $d_A$ is also a strict premetric if and only if the function $f$ is injective.
\end{restatable}


\subsection{Metrics on statistics and rewards}

By \cref{lem:pullback_premetric}, we can define a premetric $d_\Statistics$ on statistics $\Statistics$ by pulling back a premetric $d_\Rewards$ on rewards $\Rewards$ through a post-processing function $\post: \Statistics \to \Rewards$:
\begin{equation}
\forall \tau_1, \tau_2 \in \Statistics.\;
d_\Statistics(\tau_1, \tau_2)
\defeq
d_\Rewards(\post(\tau_1), \post(\tau_2)).
\end{equation}

However, when rewards $\Rewards$ are real-valued while statistics $\Statistics$ are multi-dimensional, the pullback premetric $d_\Statistics$ may not be a strict premetric, as different statistics may map to the same reward value.

For example, consider the range of rewards, where the statistics $\Statistics = \R^2$ are the maximum and minimum of rewards.
We can directly define a metric on statistics by
\begin{equation}
d_\Statistics\parens*{
\begin{bmatrix}m_1\\n_1\end{bmatrix},
\begin{bmatrix}m_2\\n_2\end{bmatrix}
}
\defeq
\sqrt{(m_1 - m_2)^2 + (n_1 - n_2)^2}.
\end{equation}
If we use the pullback premetric, we have
\begin{align}
d_\Statistics\parens*{
\begin{bmatrix}m_1\\n_1\end{bmatrix},
\begin{bmatrix}m_2\\n_2\end{bmatrix}
}
\defeq{}&
d_\Rewards\parens*{
\post\parens*{\begin{bmatrix}m_1\\n_1\end{bmatrix}},
\post\parens*{\begin{bmatrix}m_2\\n_2\end{bmatrix}}
}
\\
={}&
d_\Rewards\parens*{m_1 - n_1, m_2 - n_2}
\\
={}&
\abs{(m_1 - n_1) - (m_2 - n_2)}.
\end{align}


\subsection{Bellman operators}

Recall the definition of the Bellman operator for a state statistic function $\statistic^\States: \States \to \Statistics$:
\BellmanOperatorState*

We can define a Bellman operator for a state-action statistic function $\statistic^\StateActions: \StateActions \to \Statistics$ similarly:
\begin{definition}[Bellman operator]
\label{def:bellman_operator_state-action}
Given a policy $\policy$, a transition function $\transition$, a reward function $\reward$, a terminal condition $\terminal$, and a recursive statistic aggregation function $\aggiu$ (\cref{def:recursive_aggregation}), the \emph{Bellman operator} $\Bellman^\StateActions_\policy: [\StateActions, \Statistics] \to [\StateActions, \Statistics]$ for a function $\statistic^\StateActions: \StateActions \to \Statistics$ is defined by
\begin{equation}
\Bellman^\StateActions_\policy \statistic^\StateActions: \StateActions \to \Statistics
\defeq
\anon{s, a}{\scases{\init}{
\reward(s, a) \update \statistic^\StateActions(\transition^\StateActions_\policy(s, a))
}}.
\end{equation}
\end{definition}


\subsection{Existence of fixed points of Bellman operators}

The existence of fixed points of the Bellman operators $\Bellman^\States_\policy$ and $\Bellman^\StateActions_\policy$ is established by the Bellman equations for the state statistic function $\statistic^\States_\policy: \States \to \Statistics$ in \cref{thm:bellman_equation_state} and the state-action statistic function $\statistic^\StateActions_\policy: \StateActions \to \Statistics$ in \cref{thm:bellman_equation_state-action}.

\begin{remark}[Banach fixed point theorem]
Note that the classical fixed point theorem for Bellman operators typically relies on the \emph{Banach fixed point theorem}, which requires the underlying space to be a \emph{complete metric space}.
This is not an issue in the standard discounted sum setting, as the space $\R$ of real numbers has a complete metric structure.
However, in our setting, the space $\Statistics$ of statistics may lack such a complete metric structure, posing potential challenges for establishing fixed point guarantees.
That said, the triangle inequality of the metric and the completeness of the space are only necessary for ensuring the \emph{existence} of fixed points: the triangle inequality guarantees that the iterative sequence is a Cauchy sequence, while completeness ensures that the sequence has a limit within the space.
Since the existence of fixed points follows directly from the Bellman equations, our focus shifts to the \emph{uniqueness} of fixed points, which only requires the space to be a premetric space.
\end{remark}


\subsection{Uniqueness of fixed points of Bellman operators}

Recall that \cref{thm:bellman_fixpoint_state} establishes the uniqueness of fixed points of the Bellman operator $\Bellman^\States_\policy$ for state statistic functions $\statistic^\States: \States \to \Statistics$:
\BellmanFixpointState*

Similarly, we can extend this result to the Bellman operator $\Bellman^\StateActions_\policy$ for state-action statistic functions $\statistic^\StateActions: \StateActions \to \Statistics$:
\begin{restatable}[Uniqueness of fixed points of the Bellman operator]{theorem}{BellmanFixpointStateAction}
\label{thm:bellman_fixpoint_state-action}
Let $\statistic^\StateActions_1, \statistic^\StateActions_2: \StateActions \to \Statistics$ be fixed points of the Bellman operator $\Bellman^\StateActions_\policy$ (\cref{def:bellman_operator_state-action}).
If the update function $\update$ is contractive with respect to a premetric $d_\Statistics$ on statistics $\Statistics$ (\cref{def:contraction}), then $d_\Statistics(\statistic^\StateActions_1(s, a), \statistic^\StateActions_2(s, a)) = 0$ for all states $s \in \States$ and actions $a \in \Actions$.
If $d_\Statistics$ is a strict premetric, then $\statistic^\StateActions_1 = \statistic^\StateActions_2 = \statistic^\StateActions_\policy$.
\end{restatable}

\clearpage
\section{Orders and Bellman optimality operators}
\label{app:order}
In this section, we discuss the \emph{orders} on the statistics $\Statistics$ and rewards $\Rewards$ and the \emph{Bellman optimality operators} for the state/state-action statistic functions.


\begin{table}
\centering
\caption{Properties of orders}
\label{tab:order}
\begin{adjustbox}{width=.9\linewidth}
\begin{tabular}{ccccc}
\toprule
& Preorder & Partial order & Total preorder & Total order
\\
\midrule
\begin{tabular}{@{}c@{}}
Reflexivity\\
$a \leq_A a$
\end{tabular}
& \checkmark & \checkmark & \checkmark & \checkmark
\\
\midrule
\begin{tabular}{@{}c@{}}
Transitivity\\
$(a_1 \leq_A a_2) \lcon (a_2 \leq_A a_3) \limp (a_1 \leq_A a_3)$
\end{tabular}
& \checkmark & \checkmark & \checkmark & \checkmark
\\
\midrule
\begin{tabular}{@{}c@{}}
Antisymmetry\\
$(a_1 \leq_A a_2) \lcon (a_2 \leq_A a_1) \limp (a_1 = a_2)$
\end{tabular}
& & \checkmark & & \checkmark
\\
\midrule
\begin{tabular}{@{}c@{}}
Totality\\
$(a_1 \leq_A a_2) \ldis (a_2 \leq_A a_1)$
\end{tabular}
& & & \checkmark & \checkmark
\\
\bottomrule
\end{tabular}
\end{adjustbox}
\end{table}


\subsection{Preliminaries}

Recall the definitions of orders, as summarized in \cref{tab:order}:
\begin{definition}[Preorder]
A \emph{preorder} on a set $A$ is a relation $\leq_A$ that is reflexive $\forall a \in A.\; a \leq_A a$ and transitive $\forall a_1, a_2, a_3 \in A.\; (a_1 \leq_A a_2) \lcon (a_2 \leq_A a_3) \limp (a_1 \leq_A a_3)$.
\end{definition}

\begin{definition}[Partial order]
A \emph{partial order} on a set $A$ is a relation $\leq_A$ that is reflexive, transitive, and antisymmetric $\forall a_1, a_2 \in A.\; (a_1 \leq_A a_2) \lcon (a_2 \leq_A a_1) \limp (a_1 = a_2)$.
\end{definition}

\begin{definition}[Total preorder]
A \emph{total preorder} on a set $A$ is a relation $\leq_A$ that is reflexive, transitive, and total $\forall a_1, a_2 \in A.\; (a_1 \leq_A a_2) \ldis (a_2 \leq_A a_1)$.
\end{definition}

\begin{definition}[Total order]
A \emph{total order} on a set $A$ is a relation $\leq_A$ that is reflexive, transitive, antisymmetric, and total.
\end{definition}

Given a function to a preorder space, we can define a preorder on the domain by pullback:
\begin{restatable}[Pullback preorder]{lemma}{PullbackPreorder}
\label{lem:pullback_preorder}
Let $\leq_B$ be a preorder on a set $B$, and let $f: A \to B$ be a function.
The \emph{pullback preorder} $\leq_A$ on a set $A$ is defined by
\begin{equation}
\forall a_1, a_2 \in A.\;
(a_1 \leq_A a_2)
\defeq
(f(a_1) \leq_B f(a_2)).
\end{equation}
If $\leq_B$ is total, then $\leq_A$ is also total.
If $\leq_B$ is antisymmetric, then $\leq_A$ is also antisymmetric if and only if $f$ is injective.
\end{restatable}

Given a preorder and a premetric, wen can consider how the premetric preserves the preorder:
\begin{definition}[Preorder-preserving premetric]
\label{def:preorder-preserving_premetric}
A \emph{premetric} $d_B: B \times B \to \quant$ on a set $B$ \emph{preserves a preorder} $\leq_B$ on the set $B$ if
\begin{equation}
\forall b_1, b_2, b_3 \in B.\;
(b_1 \leq_B b_2 \leq_B b_3)
\limp
(d_B(b_1, b_2) \leq d_B(b_1, b_3))
\lcon
(d_B(b_3, b_2) \leq d_B(b_3, b_1)).
\end{equation}
\end{definition}

Note that since a premetric is not required to be symmetric, there are in total eight possible inequalities that we can consider for the preorder preservation of a premetric, which are omitted here for brevity.

Given a preorder-preserving premetric, we can consider an inequality for the supremum of functions:
\begin{restatable}[Preorder-preserving premetric's supremum inequality]{lemma}{SupremumInequality}
\label{lem:supremum_inequality}
Let $d_B: B \times B \to \quant$ be a premetric that preserves a premetric $\leq_B$ on a set $B$.
Then, for functions $f_1, f_2: A \to B$ whose suprema are attained in $B$, we have
\begin{equation}
d_B(\sup_{a \in A} f_1(a), \sup_{a \in A} f_2(a))
\leq
\sup_{a \in A} d_B(f_1(a), f_2(a)).
\end{equation}
\end{restatable}

This lemma is useful for proving the contraction property of the Bellman optimality operator, as we will see later.


\subsection{Orders on statistics and rewards}
\label{app:orders_on_statistics_and_rewards}

By \cref{lem:pullback_preorder}, we can define a preorder $\leq_\Statistics$ on statistics $\Statistics$ by pulling back a preorder $\leq_\Rewards$ on rewards $\Rewards$ through a post-processing function $\post: \Statistics \to \Rewards$:
\begin{equation}
\forall \tau_1, \tau_2 \in \Statistics.\;
(\tau_1 \leq_\Statistics \tau_2)
\defeq
(\post(\tau_1) \leq_R \post(\tau_2)).
\end{equation}

Since the (pre)order $\leq_\Rewards$ on rewards $\Rewards$ is usually the total order of real numbers, we can guarantee that the preorder $\leq_\Statistics$ on statistics $\Statistics$ is also total.

For example, consider the arithmetic mean of rewards, where the statistics $\Statistics = \N \times \R$ are the length and the sum of rewards.
We can compare two statistics $(n_1, s_1)$ and $(n_2, s_2)$ by comparing the means $\frac{s_1}{n_1}$ and $\frac{s_2}{n_2}$.
This is a total preorder on the statistics $\Statistics$.


\subsection{Bellman optimality operators}

We can define the Bellman optimality operators as follows:
\begin{definition}[Bellman optimality operator]
\label{def:optimality_operator_state}
Given a policy $\policy$, a transition function $\transition$, a reward function $\reward$, a terminal condition $\terminal$, a recursive statistic aggregation function $\aggiu$ (\cref{def:recursive_aggregation}), and a preorder $\leq_\Statistics$ on statistics $\Statistics$, the \emph{Bellman optimality operator} $\Bellman^\States_*: [\States, \Statistics] \to [\States, \Statistics]$ for a function $\statistic^\States: \States \to \Statistics$ is defined by
\begin{equation}
\Bellman^\States_* \statistic^\States: \States \to \Statistics
\defeq
\anon{s}{\scases{\init}{
\displaystyle
\sup_{a \in \Actions} \parens*{\reward(s, a) \update \statistic^\States(\transition(s, a))}
}}.
\end{equation}
\end{definition}

\begin{definition}[Bellman optimality operator]
\label{def:optimality_operator_state-action}
Given a policy $\policy$, a transition function $\transition$, a reward function $\reward$, a terminal condition $\terminal$, a recursive statistic aggregation function $\aggiu$ (\cref{def:recursive_aggregation}), and a preorder $\leq_\Statistics$ on statistics $\Statistics$, the \emph{Bellman optimality operator} $\Bellman^\StateActions_*: [\StateActions, \Statistics] \to [\StateActions, \Statistics]$ for a function $\statistic^\StateActions: \StateActions \to \Statistics$ is defined by
\begin{equation}
\Bellman^\StateActions_* \statistic^\StateActions: \StateActions \to \Statistics
\defeq
\anon{s, a}{\scases{\init}{
\displaystyle
\sup_{a' \in \Actions} \parens*{\reward(s, a) \update \statistic^\StateActions(\transition(s, a), a')}
}}.
\end{equation}
\end{definition}


\subsection{Existence of fixed points of Bellman optimality operators}

Recall that \cref{thm:optimality_equation_state} establishes the existence of a fixed point of the Bellman optimality operator $\Bellman^\States_*$ for state statistic functions $\statistic^\States: \States \to \Statistics$:
\OptimalityEquationState*

We can similarly establish the existence of a fixed point of the Bellman optimality operator $\Bellman^\StateActions_*$ for state-action statistic functions $\statistic^\StateActions: \StateActions \to \Statistics$:
\begin{restatable}[Bellman optimality equation for the state-action statistic function]{theorem}{OptimalityEquationStateAction}
\label{thm:optimality_equation_state-action}
Given a preorder $\leq_\Statistics$ on statistics $\Statistics$, the optimal state-action statistic function $\statistic^\StateActions_*$ satisfies
\begin{equation}
\statistic^\StateActions_*: \StateActions \to \Statistics
\defeq
\anon{s, a}{\scases{\init}{
\displaystyle
\sup_{a' \in \Actions} \parens*{\reward(s, a) \update \statistic^\StateActions_*(\transition(s, a), a')}
}}.
\end{equation}
\end{restatable}


\subsection{Uniqueness of fixed points of Bellman optimality operators}

Similarly to \cref{thm:bellman_fixpoint_state}, we can guarantee the uniqueness of fixed points of the Bellman optimality operators $\Bellman^\States_*$ and $\Bellman^\StateActions_*$ under certain conditions:
\begin{restatable}[Uniqueness of fixed points of Bellman optimality operator]{theorem}{OptimalityFixpointState}
\label{thm:optimality_fixpoint_state}
Let $\statistic^\States_1, \statistic^\States_2: \States \to \Statistics$ be fixed points of the Bellman optimality operator $\Bellman^\States_*$ (\cref{def:optimality_operator_state}).
If the update function $\update$ is contractive with respect to a premetric $d_\Statistics$ on statistics $\Statistics$ (\cref{def:contraction}), and the premetric $d_\Statistics$ preserves the preorder $\leq_\Statistics$ on statistics $\Statistics$ (\cref{def:preorder-preserving_premetric}), then $d_\Statistics(\statistic^\States_1(s), \statistic^\States_2(s)) = 0$ for all states $s \in \States$.
If $d_\Statistics$ is a strict premetric, then $\statistic^\States_1 = \statistic^\States_2 = \statistic^\States_*$.
\end{restatable}

\begin{restatable}[Uniqueness of fixed points of Bellman optimality operator]{theorem}{OptimalityFixpointStateAction}
\label{thm:optimality_fixpoint_state-action}
Let $\statistic^\StateActions_1, \statistic^\StateActions_2: \StateActions \to \Statistics$ be fixed points of the Bellman optimality operator $\Bellman^\StateActions_*$ (\cref{def:optimality_operator_state-action}).
If the update function $\update$ is contractive with respect to a premetric $d_\Statistics$ on statistics $\Statistics$ (\cref{def:contraction}), and the premetric $d_\Statistics$ preserves the preorder $\leq_\Statistics$ on statistics $\Statistics$ (\cref{def:preorder-preserving_premetric}), then $d_\Statistics(\statistic^\StateActions_1(s, a), \statistic^\StateActions_2(s, a)) = 0$ for all states $s \in \States$ and actions $a \in \Actions$.
If $d_\Statistics$ is a strict premetric, then $\statistic^\StateActions_1 = \statistic^\StateActions_2 = \statistic^\StateActions_*$.
\end{restatable}


In summary, the definitions and results on the fixed points of the Bellman operators and the Bellman optimality operators are summarized in \cref{tab:fixed_points}.

\begin{table}
\centering
\caption{Fixed points of the Bellman operators and the Bellman optimality operators}
\label{tab:fixed_points}
\begin{tabular}{llccc}
\toprule
&& Definition & Existence & Uniqueness
\\
\midrule
\multirow{2}{*}{Bellman operator}
& $\Bellman^\States_\policy$
& \cref{def:bellman_operator_state}
& \cref{thm:bellman_equation_state}
& \cref{thm:bellman_fixpoint_state}
\\
& $\Bellman^\StateActions_\policy$
& \cref{def:bellman_operator_state-action}
& \cref{thm:bellman_equation_state-action}
& \cref{thm:bellman_fixpoint_state-action}
\\
\midrule
\multirow{2}{*}{Bellman optimality operator}
& $\Bellman^\States_*$
& \cref{def:optimality_operator_state}
& \cref{thm:optimality_equation_state}
& \cref{thm:optimality_fixpoint_state}
\\
& $\Bellman^\StateActions_*$
& \cref{def:optimality_operator_state-action}
& \cref{thm:optimality_equation_state-action}
& \cref{thm:optimality_fixpoint_state-action}
\\
\bottomrule
\end{tabular}
\end{table}

\clearpage
\section{Stochastic Markov decision process}
\label{app:stochastic}
In this section, we discuss the stochastic extension of the deterministic Markov decision processes introduced in \cref{sec:fusion,sec:aggregation}.


\subsection{Composition of stochastic functions}

The composition rules of stochastic functions and deterministic functions are defined as follows:
\begin{itemize}
\item 
Composition of two stochastic functions 
$f: A \to \Probability{B}$ and $g: B \to \Probability{C}$ by marginalizing over the intermediate variable, as described by the \emph{Chapman--Kolmogorov equation} \citep{giry1982categorical}:
\begin{equation}
(g \compL f)(c | a)
\defeq \int_B g(c | b) f(b | a) \D{b}.
\end{equation}
\begin{equation}
\begin{tikzcd}[column sep=6em, row sep=3em, labels={description}]
A
\arrow[r, "f", rightsquigarrow, \red]
\arrow[rd, "f", \red]
\arrow[rr, "g \compL f", bend left, rightsquigarrow, \yellow]
\arrow[rrd, "g \compL f", \yellow]
&
B
\arrow[r, "g", rightsquigarrow, \blue]
\arrow[rd, "g", \blue]
&
C
\\
&
\Probability{B}
\arrow[rd, "\Probability{g}" description]
&
\Probability{C}
\\
&&
\Probability{\Probability{C}}
\arrow[u, "\mu_C"]
\end{tikzcd}
\end{equation}

\item
Composition of a stochastic function $f: A \to \Probability{B}$ with a deterministic function $g: B \to C$:
\begin{equation}
(g \compL f)(c | a)
\defeq g_* f(b | a)
= \int_B \delta_g(b) f(b | a) \D{b}.
\end{equation}
\begin{equation}
\begin{tikzcd}[column sep=6em, row sep=3em, labels={description}]
A
\arrow[r, "f", rightsquigarrow, \red]
\arrow[rd, "f", \red]
\arrow[rr, "g \compL f", bend left, rightsquigarrow, \yellow]
\arrow[rrd, "g \compL f", \yellow]
&
B
\arrow[r, "g", \blue]
\arrow[rd, "\delta_g", \blue]
&
C
\\
&
\Probability{B}
\arrow[r, "g_*"]
\arrow[rd, "\Probability{\delta_g}"]
&
\Probability{C}
\\
&&
\Probability{\Probability{C}}
\arrow[u, "\mu_C"]
\end{tikzcd}
\end{equation}

\item
Composition of a deterministic function $f: A \to B$ with a stochastic function $g: B \to \Probability{C}$:
\begin{equation}
(g \compL f)(c | a)
\defeq g(c | f(a)).
\end{equation}
\begin{equation}
\begin{tikzcd}[column sep=6em, row sep=3em, labels={description}]
A
\arrow[r, "f", \red]
\arrow[rd, "\delta_f", \red]
\arrow[rr, "g \compL f", bend left, rightsquigarrow, \yellow]
\arrow[rrd, "g \compL f", \yellow]
&
B
\arrow[r, "g", rightsquigarrow, \blue]
\arrow[rd, "g", \blue]
&
C
\\
&
\Probability{B}
\arrow[rd, "\Probability{g}"]
&
\Probability{C}
\\
&&
\Probability{\Probability{C}}
\arrow[u, "\mu_C"]
\end{tikzcd}
\end{equation}
\end{itemize}

\clearpage


\subsection{Stochastic recursion}

In \cref{sec:stochastic}, we introduced the stochastic state transition and statistic functions.
Similarly, we can define the stochastic state-action transition $\transition^\StateActions_\policy$ as follows:
\begin{flalign}
\nonumber&
\transition^\StateActions_\policy: \StateActions \to \Probability{(\StateActions)}
\defeq
{\pair} \compL \transition
&
\\
={}&
\anon{s, a}{\parens*{
s' \sim \transition(s' | s, a),
a' \sim \int_S \policy(a' | s') \transition(s' | s, a) \D{s'}
}}.
\end{flalign}

The stochastic state-action statistic function $\statistic^\StateActions_\policy$ satisfies the following recursive equation:
\begin{flalign}
\nonumber&
\statistic^\StateActions_\policy: \StateActions \to \Probability\Statistics
&
\\
={}&
\anon{s, a}{\tau \sim \scases{\delta_{\init}}{
\displaystyle
\reward(s, a) \update \tau'
\Bigm|
\tau' \sim \int_\StateActions \statistic^\StateActions_\policy(\tau' | s', a') \transition^\StateActions_\policy(s', a' | s, a) \D{s'}\D{a'}
}}.
\end{flalign}

Further characterizations of stochastic state/state-action statistic functions, including the (pre)metrics and (pre)orders on statistics, as well as the contractivity of stochastic Bellman (optimality) operators, are left for future work.


\subsection{Relationship between stochastic state and state-action statistic functions}

In the stochastic setting, the state/state-action statistic functions are related by the following equations, which are analogous to \cref{thm:state/state-action_relationship}:
\begin{align}
\statistic^\States_\policy(\tau | s)
&=
\int_\Actions \statistic^\StateActions_\policy(\tau | s, a) \policy(a | s) \D{a}
& \text{(for all states),}
\\
\statistic^\StateActions_\policy(\tau | s, a)
&=
\reward(s, a) \update \int_\States \statistic^\States_\policy(\tau | s') \transition(s' | s, a) \D{s'}
& \text{(for all non-terminal states).}
\end{align}


\subsection{Expected aggregated rewards vs.~aggregated expected rewards}

As discussed in \cref{sec:stochastic}, the expected discounted sum of rewards equals the discounted sum of expected rewards.
However, the expected aggregated rewards and the aggregated expected rewards are not equal in general.
For example, the expected maximum reward is not equal to the maximum expected reward because the expectation operator does not distribute over the maximum operator, as shown in \cref{tab:expected_aggregated_vs_aggregated_expected}.
This issue was also raised by \citet{gottipati2020maximum,
cui2023reinforcement, veviurko2024max}.
However, we argue that even though the expected aggregated rewards and the aggregated expected rewards are not equal, they are both valid and useful learning objectives for different purposes, and the choice between them depends on the specific application.
If we want to optimize the expected aggregated rewards, a more straightforward approach is to estimate the distributions of the aggregated rewards, using \emph{distributional reinforcement learning} \citep{morimura2010nonparametric,morimura2010nonparametric, bellemare2017distributional,bellemare2023distributional}.
Further theoretical and empirical investigations are left for future work.

\begin{table}
\centering
\caption{Expected aggregated rewards vs.~aggregated expected rewards: maximum as an example}
\label{tab:expected_aggregated_vs_aggregated_expected}
\begin{tabular}{lll}
\toprule
& expected maximum rewards
& maximum expected rewards
\\
\midrule
definition
& $\E_\pi[\max(r_1, r_2, \dots, r_\horizon)]$
& $\max(\E_\pi[r_1], \E_\pi[r_2], \dots, \E_\pi[r_\horizon])$
\\
statistic $\Statistics$
& max reward distribution $\in \Probability\eR$
& max reward expectation $\in \eR$
\\
initial value
& Dirac delta measure $\delta_{-\infty} \in \Probability\eR$
& reward value $-\infty \in \eR$
\\
update function
& pushforward measure update
& expected value update
\\
& $\Probability\eR \times \Probability\eR \to P(\eR \times \eR) \xto{\max_*} \Probability\eR$
& $\Probability\eR \times \eR \xto{\mathcolor{\red}{\E_\eR} \times \id_\eR} \eR \times \eR \xto{\max} \eR$
\\
post-processing
& expectation $\mathcolor{\red}{\E_\eR}: \Probability\eR \to \eR$
& identity $\id_\eR: \eR \to \eR$
\\
\bottomrule
\end{tabular}
\end{table}

\clearpage
\section{Proofs}
\label{app:proofs}
In this section, we present the proofs of the theorems and lemmas introduced in the main text.

In some derivations, we use \uwave{underwave notation} to highlight the specific subterm being rewritten or replaced.
This syntactic marking corresponds to substituting one path in a commutative diagram with another path sharing the same source and target:
\begin{equation}
\uwave{\hspace{1ex}f\hspace{1ex}} = h \compL g
\text{\quad or \quad}
\uwave{\hspace{1ex}f\hspace{1ex}}(a) = h(g(a))
\text{\quad means \quad}
\begin{tikzcd}
&
B
\arrow[rd, "h"]
\\
A
\arrow[rr, "f"]
\arrow[ru, "g"]
&&
C
\end{tikzcd}
\end{equation}

\vspace{1em}


\BellmanEquationState*
\begin{pf}
Similarly to the diagram in \cref{diag:fusion_state_value_discounted_sum}, the state statistic function $\statistic_\policy: \States \to \Statistics$ can be represented using the following diagram:
\begin{equation*}
\begin{tikzcd}[column sep=0em]
\singleton + \Rewards \times \States
\arrow[r, "\id_\singleton + \id_\Rewards \times \mathcolor{\blue}{\gen_\ptro}", unique morphism]
\arrow[rr, "\id_\singleton + \id_\Rewards \times \mathcolor{\yellow}{\statistic_\policy}", bend left=60, looseness=.2]
&[10em]
\singleton + \Rewards \times \List{\Rewards}
\arrow[r, "\id_\singleton + \id_\Rewards \times \mathcolor{\red}{\aggiu}", unique morphism]
\arrow[d, "\bracks{\nil, \cons}"']
&[9em]
\singleton + \Rewards \times \Statistics
\arrow[d, "\bracks{\init, \update}"']
\\
\States
\arrow[u, "\step_\ptro"]
\arrow[r, "\mathcolor{\blue}{\gen_\ptro}", unique morphism]
\arrow[rr, "\mathcolor{\yellow}{\statistic_\policy}", bend right=60, looseness=.2]
&
\List{\Rewards}
\arrow[r, "\mathcolor{\red}{\aggiu}", unique morphism]
&
\Statistics
\end{tikzcd}
\end{equation*}
which can be non-rigorously interpreted as a \say{combination} of the following two diagrams:
\begin{equation*}
\begin{tikzcd}[column sep=0em]
\singleton
\arrow[r, "\id_\singleton"]
\arrow[rr, "\id_\singleton", bend left=60, looseness=.4]
&[4em]
\singleton
\arrow[r, "\id_\singleton"]
\arrow[d, "\nil"']
&[4em]
\singleton
\arrow[d, "\init"']
\\
\States
\arrow[u, "e_\States"]
\arrow[r, "\mathcolor{\blue}{\gen_\ptro}"]
\arrow[rr, "\mathcolor{\yellow}{\statistic_\policy}", bend right=60, looseness=.4]
&
\List{\Rewards}
\arrow[r, "\mathcolor{\red}{\aggiu}"]
&
\Statistics
\end{tikzcd}
\quad
\begin{tikzcd}[column sep=0em]
\Rewards \times \States
\arrow[r, "\id_\Rewards \times \mathcolor{\blue}{\gen_\ptro}"]
\arrow[rr, "\id_\Rewards \times \mathcolor{\yellow}{\statistic_\policy}", bend left=60, looseness=.3]
&[6em]
\Rewards \times \List{\Rewards}
\arrow[r, "\id_\Rewards \times \mathcolor{\red}{\aggiu}"]
\arrow[d, "\cons"']
&[6em]
\Rewards \times \Statistics
\arrow[d, "\update"']
\\
\States
\arrow[u, "\angles{\reward_\policy, \transition_\policy}"]
\arrow[r, "\mathcolor{\blue}{\gen_\ptro}"]
\arrow[rr, "\mathcolor{\yellow}{\statistic_\policy}", bend right=60, looseness=.3]
&
\List{\Rewards}
\arrow[r, "\mathcolor{\red}{\aggiu}"]
&
\Statistics
\end{tikzcd}
\end{equation*}
where $e_\States: \States \to \singleton$ is the unique function from states to the singleton set, and $\angles{\reward_\policy, \transition_\policy}: \States \to \Rewards \times \States$ is the pairing of the reward and transition functions, which constitute the step function $\step_\ptro$.

The left diagram shows that when a state $s \in \States_\terminal$ is terminal,
\begin{flalign}
  \statistic_\policy(s)
&= \aggiu(\uwave{\gen_\ptro}(s))
& \text{(by definition of $\statistic_\policy$)}
\\
&= \uwave{\aggiu}(\nil)
& \text{(by terminal condition of $\gen_\ptro$)}
\\
\label{eq:state_statistic_terminal}
&= \init.
& \text{(by initial condition of $\aggiu$)}
\end{flalign}
The right diagram shows that when a state $s \notin \States_\terminal$ is non-terminal,
\begin{flalign}
  \statistic_\policy(s)
&= \aggiu(\uwave{\gen_\ptro}(s))
& \text{(by definition of $\statistic_\policy$)}
\\
&= \uwave{\aggiu(\cons}(\reward_\policy(s), \gen_\ptro(\transition_\policy(s))))
& \text{(by recursive definition of $\gen_\ptro$)}
\\
&= \reward_\policy(s) \update \uwave{\aggiu(\gen_\ptro}(\transition_\policy(s)))
& \text{(by recursive definition of $\aggiu$)}
\\
\label{eq:state_statistic_nonterminal}
&= \reward_\policy(s) \update \statistic_\policy(\transition_\policy(s)).
& \text{(by definition of $\statistic_\policy$)}
\end{flalign}
By combining \cref{eq:state_statistic_terminal} and \cref{eq:state_statistic_nonterminal}, we obtain the desired result in \cref{eq:state_statistic}.
\end{pf}

We omit the proof for \cref{thm:bellman_equation_state-action} as the derivation is similar to that of \cref{thm:bellman_equation_state}.


\clearpage

\PullbackPremetric*
\begin{pf}
The pullback premetric $d_A$ is a premetric because
\begin{equation}
\forall a \in A.\;
d_A(a, a) \defeq d_B(f(a), f(a)) = 0.
\end{equation}
If $d_B$ is a strict premetric, we have
\begin{equation}
\forall a_1, a_2 \in A.\;
(d_A(a_1, a_2) \defeq d_B(f(a_1), f(a_2)) = 0) \limp (f(a_1) = f(a_2)).
\end{equation}
For the pullback premetric $d_A$ to be a strict premetric, we require that
\begin{equation}
\forall a_1, a_2 \in A.\;
(f(a_1) = f(a_2)) \limp (a_1 = a_2),
\end{equation}
which is equivalent to the injectivity of the function $f$.
\end{pf}


\PullbackPreorder*
\begin{pf}
The pullback preorder $\leq_A$ is reflexive because
\begin{equation}
\forall a \in A.\;
(a \leq_A a) \defeq (f(a) \leq_B f(a)).
\end{equation}
The pullback preorder $\leq_A$ is transitive because
\begin{align}
\forall a_1, a_2, a_3 \in A.\;
(a_1 \leq_A a_2) \lcon (a_2 \leq_A a_3)
\defeq{}&
(f(a_1) \leq_B f(a_2)) \lcon (f(a_2) \leq_B f(a_3))
\\
\limp{}&
(f(a_1) \leq_B f(a_3))
\eqdef
(a_1 \leq_A a_3).
\end{align}
If $\leq_B$ is total, then $\leq_A$ is also total because
\begin{equation}
\forall a_1, a_2 \in A.\;
(a_1 \leq_A a_2) \ldis (a_2 \leq_A a_1)
\defeq
(f(a_1) \leq_B f(a_2)) \ldis (f(a_2) \leq_B f(a_1)).
\end{equation}
If $\leq_B$ is antisymmetric, we have
\begin{align}
\forall a_1, a_2 \in A.\;
(a_1 \leq_A a_2) \lcon (a_2 \leq_A a_1)
\defeq{}&
(f(a_1) \leq_B f(a_2)) \lcon (f(a_2) \leq_B f(a_1))
\\
\limp{}&
(f(a_1) = f(a_2)).
\end{align}
For the pullback preorder $\leq_A$ to be antisymmetric, we require that
\begin{equation}
\forall a_1, a_2 \in A.\;
(f(a_1) = f(a_2)) \limp (a_1 = a_2),
\end{equation}
which is equivalent to the injectivity of the function $f$.
\end{pf}


\SupremumInequality*
\begin{pf}
By assumption, the functions $f_1$ and $f_2$ have suprema in $B$.
We denote $a_1 = \argsup_{a \in A} f_1(a)$ and $a_2 = \argsup_{a \in A} f_2(a)$.
Then, $f_1(a_1) = \sup_{a \in A} f_1(a)$ and $f_2(a_2) = \sup_{a \in A} f_2(a)$.

If $f_1(a_1) \leq_B f_2(a_2)$, we have $f_1(a_2) \leq_B f_1(a_1) \leq_B f_2(a_2)$.
By the preorder preservation of the premetric $d_B$, we have
\begin{equation}
d_B(f_1(a_1), f_2(a_2))
\leq
d_B(f_1(a_2), f_2(a_2))
\leq
\sup_{a \in A} d_B(f_1(a), f_2(a)).
\end{equation}
Similarly, if $f_2(a_2) \leq_B f_1(a_1)$, we have $f_2(a_1) \leq_B f_2(a_2) \leq_B f_1(a_1)$.
By the preorder preservation of the premetric $d_B$, we have
\begin{align}
d_B(f_1(a_1), f_2(a_2))
\leq
d_B(f_1(a_1), f_2(a_1))
\leq
\sup_{a \in A} d_B(f_1(a), f_2(a)).
\end{align}
Therefore, we have
$
d_B(\sup_{a \in A} f_1(a), \sup_{a \in A} f_2(a))
\leq
\sup_{a \in A} d_B(f_1(a), f_2(a))
$.
\end{pf}


\clearpage

We use the following lemmas to prove \cref{thm:bellman_fixpoint_state}.


\begin{lemma}[Induced premetric on a set of functions]
\label{lem:induced_premetric}
Let $d_B: B \times B \to \quant$ be a premetric on a set $B$.
For functions $f, f': A \to B$, define $d_{[A, B]}: [A, B] \times [A, B] \to \quant$ as follows:
\begin{equation}
d_{[A, B]}(f, f')
\defeq
\sup_{a \in A} d_B(f(a), f'(a)).
\end{equation}
Then, $d_{[A, B]}$ is also a premetric.
Moreover, if $d_B$ is a strict premetric, $d_{[A, B]}$ is also a strict premetric.
\end{lemma}
\begin{pf}
$d_{[A, B]}$ is a premetric because $d_{[A, B]}(f, f) = \sup_{a \in A} d_B(f(a), f(a)) = 0$.
For two functions $f, f': A \to B$, $d_{[A, B]}(f, f') = \sup_{a \in A} d_B(f(a), f'(a)) = 0$ implies that $d_B(f(a), f'(a)) = 0$ for all $a \in A$.
If $d_B$ is a strict premetric, then $d_B(f(a), f'(a)) = 0$ implies $f(a) = f'(a)$ for all $a \in A$, which means that $f = f'$, hence if $d_B$ is a strict premetric, $d_{[A, B]}$ is also a strict premetric.
\end{pf}


\begin{lemma}[Data processing inequality]
\label{lem:data_processing_inequality}
Let $d_{[A, B]}$ be the induced premetric defined in \cref{lem:induced_premetric}.
For functions $f, f': A \to B$ and $g: A \to A$, we have
\begin{equation}
d_{[A, B]}(f \compL g, f' \compL g)
\leq
d_{[A, B]}(f, f').
\end{equation}
\end{lemma}
\begin{pf}
$
d_{[A, B]}(f \compL g, f' \compL g)
\defeq
\sup_{a \in A} d_B(f(g(a)), f'(g(a)))
=
\sup_{a' \in g(A)} d_B(f(a'), f'(a'))
\\
\leq
\sup_{a' \in A} d_B(f(a'), f'(a'))
\eqdef
d_{[A, B]}(f, f').
$
\end{pf}


\begin{lemma}[Uniqueness of fixed points of a premetric contraction]
\label{lem:unique_fixed_point_premetric_contraction}
Let $a_1$ and $a_2$ be fixed points of a function $f: A \to A$.
If the function $f$ is contractive with respect to a premetric $d_A$ on the set $A$, then $d_A(a_1, a_2) = 0$.
Moreover, if $d_A$ is a strict premetric, then $a_1 = a_2$.
\end{lemma}
\begin{pf}
Because $a_1$ and $a_2$ are fixed points of $f$, and $f$ is contractive with respect to $d_A$, there exists a constant $k \in [0, 1)$ such that
\begin{equation}
d_A(a_1, a_2)
=
d_A(f(a_1), f(a_2))
\leq
k \cdot d_A(a_1, a_2).
\end{equation}
Given that $d_A(a_1, a_2) \geq 0$, the only possible solution is $d_A(a_1, a_2) = 0$.
If $d_A$ is a strict premetric, then $d_A(a_1, a_2) = 0$ implies $a_1 = a_2$.
In other words, a premetric contraction has unique fixed points \emph{up to premetric indiscernibility}, while a strict premetric contraction has a unique fixed point.
\end{pf}


\begin{lemma}[Contraction of Bellman operator]
\label{lem:contraction_of_bellman_operator}
If the update function $\update$ is contractive with respect to a premetric $d_\Statistics$ on statistics $\Statistics$ (\cref{def:contraction}), then the Bellman operator $\Bellman^\States_\policy$ (\cref{def:bellman_operator_state}) is contractive with respect to the induced premetric $d_{[\States, \Statistics]}$ defined in \cref{lem:induced_premetric}.
\end{lemma}
\begin{pf}
For any functions $\statistic^\States_1, \statistic^\States_2: \States \to \Statistics$, we have
\begin{equation}
d_{[\States, \Statistics]}\parens{
\Bellman^\States_\policy \statistic^\States_1,
\Bellman^\States_\policy \statistic^\States_2
}
=
\sup_{s \in \States} d_\Statistics\parens{
(\Bellman^\States_\policy \statistic^\States_1)(s),
(\Bellman^\States_\policy \statistic^\States_2)(s)
}.
\end{equation}

When a state $s \in \States_\terminal$ is terminal, for any $k \in [0, 1)$, we have
\begin{flalign}
&
d_\Statistics\parens{
(\Bellman^\States_\policy \statistic^\States_1)(s),
(\Bellman^\States_\policy \statistic^\States_2)(s)
}
\\
={}&
d_\Statistics(\init, \init)
& \text{(by definition of $\Bellman_\policy$)}
\\
={}&
0
\leq
k \cdot d_\Statistics\parens{
\statistic^\States_1(\transition^\States_\policy(s)),
\statistic^\States_2(\transition^\States_\policy(s))
}
& \text{($d_\Statistics$ is a premetric)}
\end{flalign}
When a state $s \notin \States_\terminal$ is non-terminal, there exists a constant $k \in [0, 1)$ such that
\begin{flalign}
&
d_\Statistics\parens{
(\Bellman^\States_\policy \statistic^\States_1)(s),
(\Bellman^\States_\policy \statistic^\States_2)(s)
}
\\
={}&
d_\Statistics\parens{
\reward_\policy(s) \update \statistic^\States_1(\transition^\States_\policy(s)),
\reward_\policy(s) \update \statistic^\States_2(\transition^\States_\policy(s))
}
& \text{(by definition of $\Bellman^\States_\policy$)}
\\
\leq{}&
k \cdot d_\Statistics\parens{
\statistic^\States_1(\transition^\States_\policy(s)),
\statistic^\States_2(\transition^\States_\policy(s))
}
& \text{(by contractivity of $\update$)}
\end{flalign}
Then, we have
\begin{flalign}
&
d_{[\States, \Statistics]}\parens{
\Bellman^\States_\policy \statistic^\States_1,
\Bellman^\States_\policy \statistic^\States_2
}
\\
\leq{}&
k \cdot \sup_{s \in \States} d_\Statistics\parens{
\statistic^\States_1(\transition^\States_\policy(s)),
\statistic^\States_2(\transition^\States_\policy(s))
}
& \text{(by monotonicity and homogeneity of $\sup$)}
\\
={}&
k \cdot d_{[\States, \Statistics]}\parens{
\statistic^\States_1 \compL \transition^\States_\policy,
\statistic^\States_2 \compL \transition^\States_\policy
}
& \text{(by definition of $d_{[\States, \Statistics]}$)}
\\
\leq{}&
k \cdot d_{[\States, \Statistics]}\parens{
\statistic^\States_1,
\statistic^\States_2
}
& \text{(\cref{lem:data_processing_inequality})}
\end{flalign}
Therefore, the Bellman operator $\Bellman^\States_\policy$ is contractive with respect to the premetric $d_{[\States, \Statistics]}$.
\end{pf}


\clearpage

\BellmanFixpointState*
\begin{pf}
Let $d_{[\States, \Statistics]}$ be the induced premetric defined in \cref{lem:induced_premetric}.
By \cref{lem:contraction_of_bellman_operator,lem:unique_fixed_point_premetric_contraction}, we have
\begin{equation}
  d_{[\States, \Statistics]}(\statistic_1, \statistic_2)
= \sup_{s \in \States} d_\Statistics(\statistic_1(s), \statistic_2(s))
= 0, 
\end{equation}
which means that $d_\Statistics(\statistic_1(s), \statistic_2(s)) = 0$ for all states $s \in \States$.
When $d_\Statistics$ is a strict premetric, we have $\statistic_1 = \statistic_2$, which means that $\statistic_\policy$ is the unique fixed point of the Bellman operator $\Bellman_\policy$.
\end{pf}

We omit the proof for \cref{thm:bellman_fixpoint_state-action} as the derivation is similar to that of \cref{thm:bellman_fixpoint_state}.


\OptimalityEquationState*
\begin{pf}
When a state $s \in \States_\terminal$ is terminal, we have
$\statistic_*(s) = \init$.
When a state $s \notin \States_\terminal$ is non-terminal, we have
\begin{flalign}
\statistic_*(s)
\defeq{}&
\statistic_{\policy_*}(s)
& \text{(by definition of $\statistic_*$)}
\\
={}&
\reward_{\policy_*}(s) \update \statistic_*(\transition_{\policy_*}(s))
& \text{(by recursive definition of $\statistic_{\policy_*}$)}
\\
={}&
\reward(s, \policy_*(s)) \update \statistic_*(\transition(s, \policy_*(s)))
& \text{(by definitions of $\reward_{\policy_*}$ and $\transition_{\policy_*}$)}
\\
={}&
\sup_{a \in \Actions} \parens*{
\reward(s, a) \update \statistic_*(\transition(s, a))
}.
& \text{(pointwise maximization)}
\end{flalign}
\end{pf}

\OptimalityEquationStateAction*
\begin{pf}
When a state $s \in \States_\terminal$ is terminal, we have
$\statistic^\StateActions_*(s, a) = \init$ for all actions $a \in \Actions$.
When a state $s \notin \States_\terminal$ is non-terminal, we have
\begin{flalign}
\statistic^\StateActions_*(s, a)
\defeq{}&
\statistic^\StateActions_{\policy_*}(s, a)
& \text{(by definition of $\statistic^\StateActions_*$)}
\\
={}&
\reward(s, a) \update \statistic^\StateActions_*(\transition^\StateActions_{\policy_*}(s, a))
& \text{(by recursive definition of $\statistic^\StateActions_{\policy_*}$)}
\\
={}&
\reward(s, a) \update \statistic^\StateActions_*(\transition(s, a), \policy_*(\transition(s, a)))
& \text{(by definition of $\transition^\StateActions_{\policy_*}$)}
\\
={}&
\sup_{a' \in \Actions} \parens*{
\reward(s, a) \update \statistic^\StateActions_*(\transition(s, a), a')
}.
& \text{(pointwise maximization)}
\end{flalign}
\end{pf}


\clearpage

Similarly to \cref{lem:contraction_of_bellman_operator,thm:bellman_fixpoint_state}, we use the following lemma to prove \cref{thm:optimality_fixpoint_state}.

\begin{lemma}[Contraction of Bellman optimality operator]
\label{lem:contraction_of_bellman_optimality_operator}
If the update function $\update$ is contractive with respect to a premetric $d_\Statistics$ on statistics $\Statistics$ (\cref{def:contraction}), and the premetric $d_\Statistics$ preserves the preorder $\leq_\Statistics$ on statistics $\Statistics$ (\cref{def:preorder-preserving_premetric}), then the Bellman optimality operator $\Bellman^\States_*$ (\cref{def:optimality_operator_state}) is contractive with respect to the induced premetric $d_{[\States, \Statistics]}$ defined in \cref{lem:induced_premetric}.
\end{lemma}
\begin{pf}
For any functions $\statistic^\States_1, \statistic^\States_2: \States \to \Statistics$, we have
\begin{equation}
d_{[\States, \Statistics]}\parens{
\Bellman^\States_* \statistic^\States_1,
\Bellman^\States_* \statistic^\States_2
}
=
\sup_{s \in \States} d_\Statistics\parens{
(\Bellman^\States_* \statistic^\States_1)(s),
(\Bellman^\States_* \statistic^\States_2)(s)
}.
\end{equation}

When a state $s \in \States_\terminal$ is terminal, for any $k \in [0, 1)$, we have
\begin{flalign}
&
d_\Statistics\parens{
(\Bellman^\States_* \statistic^\States_1)(s),
(\Bellman^\States_* \statistic^\States_2)(s)
}
\\
={}&
d_\Statistics(\init, \init)
& \text{(by definition of $\Bellman^\States_*$)}
\\
={}&
0
\leq
k \cdot \sup_{a \in \Actions} d_\Statistics\parens{
\statistic^\States_1(\transition(s, a)),
\statistic^\States_2(\transition(s, a))
}
& \text{($d_\Statistics$ is a premetric)}
\end{flalign}
When a state $s \notin \States_\terminal$ is non-terminal, there exists a constant $k \in [0, 1)$ such that
\begin{flalign}
&
d_\Statistics\parens{
(\Bellman^\States_* \statistic^\States_1)(s),
(\Bellman^\States_* \statistic^\States_2)(s)
}
\\
={}&
d_\Statistics\parens{
\sup_{a \in \Actions} \parens{\reward(s, a) \update \statistic^\States_1(\transition(s, a))},
\sup_{a \in \Actions} \parens{\reward(s, a) \update \statistic^\States_2(\transition(s, a))}
}
& \text{(by definition of $\Bellman_*$)}
\\
\leq{}&
\sup_{a \in \Actions} d_\Statistics\parens{
\reward(s, a) \update \statistic^\States_1(\transition(s, a)),
\reward(s, a) \update \statistic^\States_2(\transition(s, a))
}
& \text{(by monotonicity of $d_\Statistics$)}
\\
\leq{}&
\sup_{a \in \Actions} k \cdot d_\Statistics\parens{
\statistic^\States_1(\transition(s, a)),
\statistic^\States_2(\transition(s, a))
}
& \text{(by contractivity of $\update$)}
\\
={}&
k \cdot \sup_{a \in \Actions} d_\Statistics\parens{
\statistic^\States_1(\transition(s, a)),
\statistic^\States_2(\transition(s, a))
}
& \text{(by homogeneity of $\sup$)}
\end{flalign}
Then, we have
\begin{flalign}
&
d_{[\States, \Statistics]}\parens{
\Bellman^\States_* \statistic_1,
\Bellman^\States_* \statistic_2
}
\\
\leq{}&
k \cdot \sup_{s \in \States} \sup_{a \in \Actions} d_\Statistics\parens{
\statistic^\States_1(\transition(s, a)),
\statistic^\States_2(\transition(s, a))
}
& \text{(by monotonicity and homogeneity of $\sup$)}
\\
={}&
k \cdot \sup_{a \in \Actions} \sup_{s \in \States} d_\Statistics\parens{
\statistic^\States_1(\transition(s, a)),
\statistic^\States_2(\transition(s, a))
}
& \text{(by commutativity of $\sup$)}
\\
={}&
k \cdot \sup_{a \in \Actions} d_{[\States, \Statistics]}\parens{
\statistic^\States_1 \compL \transition(-, a),
\statistic^\States_2 \compL \transition(-, a)
}
& \text{(by definition of $d_{[\States, \Statistics]}$)}
\\
\leq{}&
k \cdot d_{[\States, \Statistics]}\parens{
\statistic^\States_1,
\statistic^\States_2
}
& \text{(\cref{lem:data_processing_inequality})}
\end{flalign}
Therefore, the Bellman optimality operator $\Bellman^\States_*$ is contractive with respect to the premetric $d_{[\States, \Statistics]}$.
\end{pf}


\OptimalityFixpointState*
\begin{pf}
Let $d_{[\States, \Statistics]}$ be the induced premetric defined in \cref{lem:induced_premetric}.
By \cref{lem:contraction_of_bellman_optimality_operator,lem:unique_fixed_point_premetric_contraction}, we have
\begin{equation}
  d_{[\States, \Statistics]}(\statistic_1, \statistic_2)
= \sup_{s \in \States} d_\Statistics(\statistic^\States_1(s), \statistic^\States_2(s))
= 0, 
\end{equation}
which means that $d_\Statistics(\statistic^\States_1(s), \statistic^\States_2(s)) = 0$ for all states $s \in \States$.
When $d_\Statistics$ is a strict premetric, we have $\statistic^\States_1 = \statistic^\States_2$, which means that $\statistic^\States_*$ is the unique fixed point of the Bellman optimality operator $\Bellman^\States_*$.
\end{pf}

We omit the proof for \cref{thm:optimality_fixpoint_state-action} as the derivation is similar to that of \cref{thm:optimality_fixpoint_state}.


\clearpage

\StateStateActionRelationship*
\begin{pf}
Notice the following relation:
\begin{equation}
  \uwave{\transition^\StateActions_\policy \compL {\pair}}
= \pair \compL \transition \compL {\pair}
=  \uwave{\pair \compL \transition^\States_\policy}
: \States \to \StateActions.
\end{equation}

We can show that when a state $s \in \States_\terminal$ is terminal,
\begin{equation}
  \parens*{
  {\gen^\StateActions_\ptro} \compL {\pair}
  }(s)
= \gen^\States_\ptro(s)
= \emptylist,
\end{equation}
and when a state $s \notin \States_\terminal$ is non-terminal,
\begin{align}
  \parens*{
  {\gen^\StateActions_\ptro} \compL {\pair}
  }(s)
&=
  \parens*{
  {\cons} \compL \uwave{\angles{\reward, {\gen^\StateActions_\ptro} \compL \transition^\StateActions_\policy} \compL {\pair}}
  }(s)
\\
&=
  \parens*{
  {\cons} \compL \angles{\uwave{\reward \compL {\pair}}, {\gen^\StateActions_\ptro} \compL \uwave{\transition^\StateActions_\policy \compL {\pair}}}
  }(s)
\\
&=
  \parens*{
  {\cons} \compL \angles{\reward_\policy, \uwave{{\gen^\StateActions_\ptro} \compL {\pair}} \compL \transition^\States_\policy}
  }(s),
\end{align}
which shows that ${\gen^\StateActions_\ptro} \compL {\pair}$ satisfies the same recursive equation as ${\gen^\States_\ptro}$ in \cref{eq:state_generation}.
Due to the uniqueness of the recursive coalgebra \citep[Eq.~(5)]{hinze2010theory}, we can conclude that
\begin{equation}
\label{eq:state/state-action_generation}
  \gen^\States_\ptro
= {\gen^\StateActions_\ptro} \compL {\pair}
: \States \to \List{\Rewards}.
\end{equation}

Given \cref{eq:state/state-action_generation}, we have
\begin{equation}
  \statistic^\States_\policy
\defeq
  {\aggiu} \compL {\gen^\States_\ptro}
= {\aggiu} \compL {\gen^\StateActions_\ptro} \compL {\pair}
= \statistic^\StateActions_\policy \compL {\pair}
: \States \to \Statistics.
\end{equation}

Next, for a non-terminal state $s \notin \States_\terminal$ and an action $a \in \Actions$, we have
\begin{align}
  \statistic^\StateActions_\policy(s, a)
&=
  \parens*{\reward \update \parens*{
  {\statistic^\StateActions_\policy} \compL \uwave{\transition^\StateActions_\policy}
  }}(s, a)
\\
&=
  \parens*{\reward \update \parens*{
  \uwave{{\statistic^\StateActions_\policy} \compL {\pair}} \compL \transition
  }}(s, a)
\\
&=
  \parens*{\reward \update \parens*{
  {\statistic^\States_\policy} \compL \transition
  }}(s, a).
\end{align}

However, for a terminal state $s \in \States_\terminal$ and an action $a \in \Actions$, the equation
$
  \statistic^\StateActions_\policy
=
  \reward \update (\statistic^\States_\policy \compL \transition)
$
may not always hold and could require additional conditions on the transition function $\transition$, the reward function $\reward$, the initial value $\init$, and the update function $\update$.

Intuitively, \cref{eq:state_state-action,eq:state-action_state} arise from the decomposition of the bidirectional process, as illustrated in \cref{fig:optic_state_state-action}.
\end{pf}


\begin{figure}
\centering
\vspace{-1em}
\begin{adjustbox}{scale=.8}
\begin{tikzpicture}
\coordinate (i) at (0, 0);
\coordinate (k) at (4, 0);
\coordinate (j) at (9, 0);
\coordinate (d) at (11, 0);

\coordinate (s) at (0, 3);
\coordinate (a) at (0, 2);
\coordinate (t) at (0, 0);

\node (Si) at (i |- s) {$\States$};
\node (Ti) at (i |- t) {$\Statistics$};
\node (Sk) at (k |- s) {$\States$};
\node (Ak) at (k |- a) {$\Actions$};
\node (Tk) at (k |- t) {$\Statistics$};
\node (Sj) at (j |- s) {$\States$};
\node (Tj) at (j |- t) {$\Statistics$};
\node (Sd) at (d |- s) {$\dots$};
\node (Td) at (d |- t) {$\dots$};

\node (copyS) [diagonal] at (2, |- s) {};
\node (copySt) [diagonal] at (6, |- s) {};
\node (copyA) [diagonal] at (7, |- a) {};
\draw (Si) [->] to (copyS);
\draw (copyS) [->] to (Sk);
\draw (Sk) [->] to (copySt);
\draw (Ak) [->] to (copyA);

\node (policy) [morphism, minimum width=.6cm, minimum height=.6cm] at (2, |- a) {$\policy$};
\draw (copyS) [->] to (policy);
\draw (policy) [->] to (Ak);

\node (transition) [morphism, minimum width=.6cm, minimum height=.6cm] at (7, |- s) {$\transition$};
\draw (copySt) [->] to (transition);
\draw (copyA) [->] to (transition);
\draw (transition) [->] to (Sj);

\node (reward) [morphism, minimum width=1.6cm, minimum height=.6cm] at (6.5, 1) {$\reward$};
\draw (copySt) [->] to (copySt |- reward.north);
\draw (copyA) [->] to (copyA |- reward.north);

\node (update) [morphism, minimum width=.6cm, minimum height=.6cm] at (6.5, |- t) {$\update$};
\draw (reward) [->] to (update);
\draw (Tj) [->] to (update);
\draw (update) [->] to (Tk);
\draw (Tk) [->] to (Ti);

\draw (Sj) [->] to (Sd);
\draw (Tj) [->] to (Td);

\node [draw, rectangle, dashed, minimum width=1cm, minimum height=4cm, color=\blue, line width=1.5pt] at (2, 1.5) {};
\node [draw, rectangle, dashed, minimum width=2cm, minimum height=4cm, color=\red, line width=1.5pt] at (6.5, 1.5) {};

\node [right=0 of Si.north east] {$s_t$};
\node [right=0 of Sk.north east] {$s_t$};
\node [right=0 of Sj.north east] {$s_{t+1}$};
\node [right=0 of Ti.north east] {$\tau_t$};
\node [right=0 of Tk.north east] {$\tau_t$};
\node [right=0 of Tj.north east] {$\tau_{t+1}$};
\node [right=0 of Ak.north east] {$a_t$};

\end{tikzpicture}
\end{adjustbox}
\caption[Relationship between state and state-action statistic functions]{%
$\mathcolor{\blue}{
  \statistic^\States_\policy
=
\statistic^\StateActions_\policy \compL {\pair}
}$
and
$\mathcolor{\red}{
  \statistic^\StateActions_\policy
=
  \reward \update (\statistic^\States_\policy \compL \transition)
}$
}
\label{fig:optic_state_state-action}
\end{figure}


\begin{remark}
In fact, we can derive \cref{eq:state/state-action_generation} directly from the relation between the state step function $\step^\States_\ptro$ and the state-action step function $\step^\StateActions_\ptro$.

When a state $s \in \States_\terminal$ is terminal,
\begin{equation}
  \parens*{{\step^\StateActions_\ptro} \compL {\pair}}(s)
= \parens*{{\id_\singleton} \compL {\step^\States_\ptro}}(s)
= *,
\end{equation}
and when a state $s \notin \States_\terminal$ is non-terminal,
\begin{flalign}
  \parens*{
  {\step^\StateActions_\ptro} \compL {\pair}
  }(s)
&=
  \parens*{
  \uwave{\angles{\reward, \transition^\StateActions_\policy} \compL {\pair}}
  }(s)
\\
&=
  \parens*{
  \angles{\uwave{\reward \compL {\pair}}, \uwave{\transition^\StateActions_\policy \compL {\pair}}}
  }(s)
\\
&=
  \parens*{
  \uwave{\angles{\reward_\policy, \pair \compL \transition^\States_\policy}}
  }(s)
\\
&=
  \parens*{
  (\id_\Rewards \times \pair) \compL \uwave{\angles{\reward_\policy, \transition^\States_\policy}}
  }(s)
\\
&=
  \parens*{
  (\id_\Rewards \times \pair) \compL \step^\States_\ptro
  }(s).
\end{flalign}
We can conclude that
\begin{equation}
  {\step^\StateActions_\ptro} \compL {\pair}
= \parens*{\id_\singleton + \id_\Rewards \times \pair} \compL {\step^\States_\ptro}
: \States \to \singleton + \Rewards \times (\StateActions),
\end{equation}
which means that $\pair$ is a \emph{coalgebra homomorphism} from the state step function $\step^\States_\ptro$ to the state-action step function $\step^\StateActions_\ptro$.
Then, by the \emph{coalgebra fusion law} \citep[Eq.~(7)]{hinze2010theory}, we can get the result in \cref{eq:state/state-action_generation}.
\end{remark}

\clearpage
\section{Learning algorithms with recursive reward aggregation}
\label{app:algorithms}
In this section, we list the RL algorithms with recursive reward aggregation used in our experiments.
The colored lines indicate modifications compared to the standard discounted sum version.


\subsection{Q-learning}

\begin{algorithm}
\caption{Q-learning \citep{watkins1992q} with recursive reward aggregation}
\label{alg:q-learning}
\begin{algorithmic}
\State \textbf{Input:}
transition function $\transition: \StateActions \to \States$,
reward function $\reward: \StateActions \to \Rewards$,
terminal condition $\terminal$,
\modified{recursive reward aggregation function ${\post} \compL {\aggiu}: \List{\Rewards} \to \Rewards$}
\State \textbf{Parameters:}
learning rate $\alpha \in (0, 1]$,
exploration parameter $\epsilon \in (0, 1)$
\State \textbf{Initialize} \modified{ state-action statistic function $\statistic: \StateActions \to \Statistics$ with initial value $\init \in \Statistics$}
\For{each episode}
  \State \textbf{Initialize} state $s$
  \While{$s$ is not terminal}
    \State \textbf{Compute} \modified{state-action value function $\qvalue(s, a) = \post(\statistic(s, a))$} for state $s$ and all actions $a$
    \State Select action $a$ using $\epsilon$-greedy policy based on value function $\qvalue(s, a)$
    \State Execute action $a$, observe next state $s'$ and reward $r$
    \State \textbf{Update} \modified{state-action statistic function $\statistic$:
    \State
    $\displaystyle
    \statistic(s, a)
    \gets
    \statistic(s, a) + \alpha\parens*{\max_{a' \in \Actions}\parens*{r \update \statistic(s', a')} - \statistic(s, a)},
    $
    \State where
    $\displaystyle
    \max_{a' \in \Actions}\parens*{r \update \statistic(s', a')}
    =
    r \update \statistic(s', a^*)$
    and
    $\displaystyle
    a^*
    =
    \argmax_{a' \in \Actions} \post\parens{r \update \statistic(s', a')}
    $
    }
    \State \textbf{Update} state $s \gets s'$
  \EndWhile
\EndFor
\State \textbf{Output:}
estimated \modified{optimal statistic function $\statistic$} and optimal policy $\policy(s) = \argmax_{a \in \Actions} \qvalue(s, a)$, where $\qvalue(s, a) = \post(\statistic(s, a))$
\end{algorithmic}
\end{algorithm}


\clearpage
\subsection{PPO}

\begin{algorithm}
\caption{PPO \citep{schulman2017proximal} with recursive reward aggregation}
\label{alg:ppo}
\begin{algorithmic}
\State \textbf{Input:}
transition function $\transition: \StateActions \to \States$,
reward function $\reward: \StateActions \to \Rewards$,
terminal condition $\terminal$,
\modified{recursive reward aggregation function ${\post} \compL {\aggiu}: \List{\Rewards} \to \Rewards$}
\State \textbf{Parameters:}
bias-variance trade-off parameter $\lambda \in [0, 1]$,
critic loss coefficient $c_1$,
entropy regularization coefficient $c_2$
\State \textbf{Initialize} parameterized policy function (actor) $\policy_\actorP: \States \to \Actions$
\State \textbf{Initialize} \modified{parameterized  state statistic function (critic) $\statistic_\criticP: \States \to \Statistics$}
\For{each episode}
  \State \textbf{Initialize} state $s$
  \State Collect trajectories of states and rewards following policy $\policy_\actorP$ till the end of the horizon $\horizon$
  \State \textbf{Compute} \modified{statistics
  $\displaystyle
  \hat\tau^{(i)}_t
  =
  r_t \update r_{t+1} \update \dots \update r_{t+i-1} \update \statistic_\criticP(s_{t+i})
  $
  }
  for $i = 1, \dots, \horizon - t$
  \State \textbf{Compute} \modified{state value function $\svalue_\criticP(s_t) = \post(\statistic_\criticP(s_t))$}
  \State \textbf{Compute} advantage estimates
  $\displaystyle
  \hat\advantage^{(i)}_t
  =
  \modified{\post(\hat\tau^{(i)}_t)} - \svalue_\criticP(s_t)$
  for $i = 1, \dots, \horizon - t
  $
  \State Use one of the following as advantage $\hat\advantage_t$:
  \State \labelitemi\;
  $
  \hat\advantage^{(1)}_{t}
  =
  \modified{\post(r_t \update \statistic_\criticP(s_{t+1}))} - \svalue_\criticP(s_t)
  $
  \State \labelitemi\;
  $
  \hat\advantage^{(\horizon - t)}_{t}
  =
  \modified{\post(r_t \update r_{t+1} \update \dots \update \statistic_\criticP(s_\horizon))} - \svalue_\criticP(s_t)
  $
  \State \labelitemi\;
  generalized advantage estimates (GAE) \citep{schulman2016high}
  $\displaystyle
  (1 - \lambda) \sum_{i=1}^{\horizon - t} \lambda^{i-1} \hat\advantage^{(i)}_t
  $
  \State \textbf{Compute} critic loss:
  $\displaystyle
  L_c(\criticP)
  =
  \sum_{t=1}^\horizon
  \parens*{\svalue_\criticP(s_t) - \modified{\post({\hat\tau^{(\horizon - t)}_t})}}^2
  $
  \State \textbf{Compute} actor loss $L_a(\actorP)$ with clipping or penalty using advantage $\hat\advantage_t$ \citep{schulman2017proximal}
  \State \textbf{Compute} entropy regularization $\Eta(\actorP)$
  \State \textbf{Optimize} $L_a(\actorP) - c_1 L_c(\criticP) + c_2 \Eta(\actorP)$
\EndFor
\State \textbf{Output:}
estimated \modified{optimal statistic function $\statistic_\criticP$} and optimal policy $\policy_\actorP$
\end{algorithmic}
\end{algorithm}


\clearpage
\subsection{TD3}

\begin{algorithm}
\caption{TD3 \citep{fujimoto2018addressing} with recursive reward aggregation}
\label{alg:td3}
\begin{algorithmic}
\State \textbf{Input:}
transition function $\transition: \StateActions \to \States$,
reward function $\reward: \StateActions \to \Rewards$,
terminal condition $\terminal$,
\modified{recursive reward aggregation function ${\post} \compL {\aggiu}: \List{\Rewards} \to \Rewards$}
\State \textbf{Parameters:}
action variance $\sigma^2$,
soft target update rate $\lambda \in (0, 1)$
\State \textbf{Initialize} parameterized policy function (actor) $\policy_\actorP: \States \to \Actions$
\State \textbf{Initialize} \modified{two parameterized state-action statistic functions (critics) $\statistic_{\criticP_1}, \statistic_{\criticP_2}: \StateActions \to \Statistics$}
\State \textbf{Initialize} targets
$\actorP' \gets \actorP$,
\modified{
$\criticP_1' \gets \criticP_1$, 
$\criticP_2' \gets \criticP_2$
},
and replay buffer $\mathcal{D}$
\For{each episode}
  \State \textbf{Initialize} state $s$
  \While{$s$ is not terminal}
    \State Select action $a \sim \Normal(\policy_\actorP(s), \sigma^2)$
    (optionally with clipping)
    \State Execute action $a$, observe next state $s'$ and reward $r$
    \State Store transition tuple $(s, a, r, s')$ in buffer $\mathcal{D}$
  \EndWhile
  \State \textbf{Compute} \modified{state-action value functions $\qvalue_{\criticP_i}(s, a) = \post(\statistic_{\criticP_i}(s, a))$} for $i = 1, 2$
  \If{update critics}
    \State Sample a batch of transitions $B = \set{(s, a, r, s')}$ from buffer $\mathcal{D}$
    \State Select target action $\tilde{a} \sim \Normal(\policy_{\actorP'}(s'), \sigma^2)$
    (optionally with clipping)
    \State \textbf{Compute} \modified{target statistic $\statistic_{\text{target}}$:
    \State
    $\displaystyle
    \statistic_\text{target}
    =
    \scases{\init}{\displaystyle \min_{i = 1, 2} r \update \statistic_{\criticP_i'}(s', \tilde{a})}
    $
    \State where
    $\displaystyle
    \min_{i = 1, 2} r \update \statistic_{\criticP_i'}(s', \tilde{a})
    =
    \begin{cases}
    r \update \statistic_{\criticP_1'}(s', \tilde{a}) & \post\parens{r \update \statistic_{\criticP_1'}(s', \tilde{a})} \leq \post\parens{r \update \statistic_{\criticP_2'}(s', \tilde{a})}
    \\
    r \update \statistic_{\criticP_2'}(s', \tilde{a}) & \text{otherwise}
    \end{cases}
    $
    }
    \State \textbf{Update} critics $\statistic_{\criticP_i}$ by gradient descent:
    \State
    $\displaystyle
    \nabla_{\criticP_i}
    \frac1{\abs{B}} \sum_{(s, a, r, s') \in B}
    \parens*{\qvalue_{\criticP_i}(s, a) - \modified{\post(\statistic_{\text{target}})}}^2
    $
    for $i = 1, 2$
  \EndIf
  \If{update actor}
    \State \textbf{Update} actor by gradient ascent:
    \State
    $\displaystyle
    \nabla_\actorP
    \frac1{\abs{B}} \sum_{(s, a, r, s') \in B}
    \qvalue_{\criticP_1}(s, \policy_{\actorP}(s))
    $
    \State \textbf{Update} targets:
    \State $\criticP_i' \gets \lambda \criticP_i + (1 - \lambda) \criticP_i'$ for $i = 1, 2$
    \State $\actorP' \gets \lambda \actorP + (1 - \lambda) \actorP'$
  \EndIf
\EndFor
\State \textbf{Output:}
estimated \modified{optimal statistic functions $\statistic_{\criticP_1}$ and $\statistic_{\criticP_2}$} and optimal policy $\policy_\actorP$
\end{algorithmic}
\end{algorithm}

\clearpage
\section{Experiments}
\label{app:experiments}
In this section, we provide detailed descriptions of the environments used in our experiments and the specific configurations and hyperparameters employed for each task.
We also present additional results for the grid-world and continuous control environments.


\subsection{Grid-world environment}


\paragraph{Implementation}

We implemented the environment and the Q-learning \citep{watkins1992q} algorithm using NumPy \citep{numpy}.


\paragraph{Hyperparameters}

We used a fixed exploration parameter of $0.3$.
We trained agents for total training time steps of $\num{10000}$.
We repeated each experiment with three different random seeds and observed that all runs consistently converged to the same solution.
We therefore present the result from one representative run.


\paragraph{Additional results}

\begin{figure}
\centering
\begin{subfigure}{0.2\linewidth}
\includegraphics[width=\textwidth]{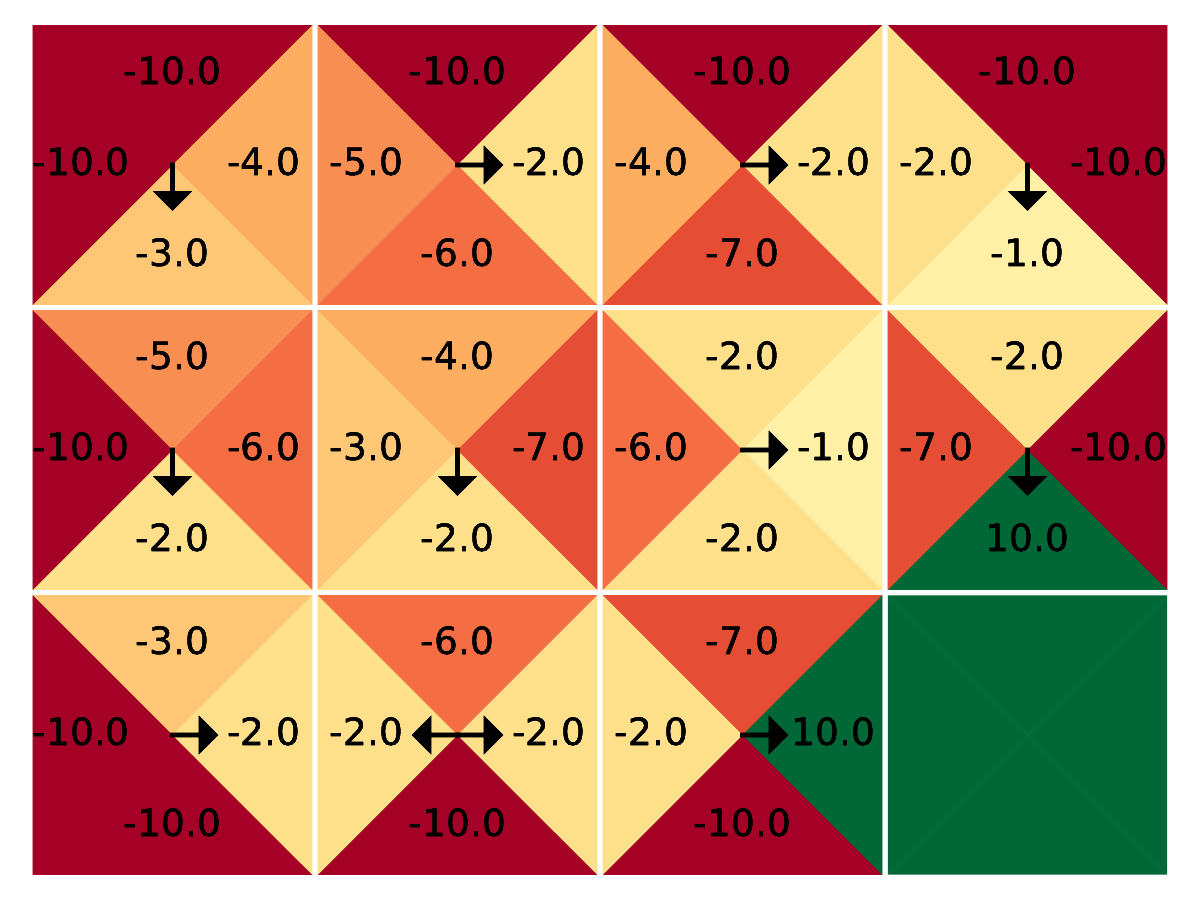}
\caption{$\min$}
\end{subfigure}%
\begin{subfigure}{0.2\linewidth}
\includegraphics[width=\textwidth]{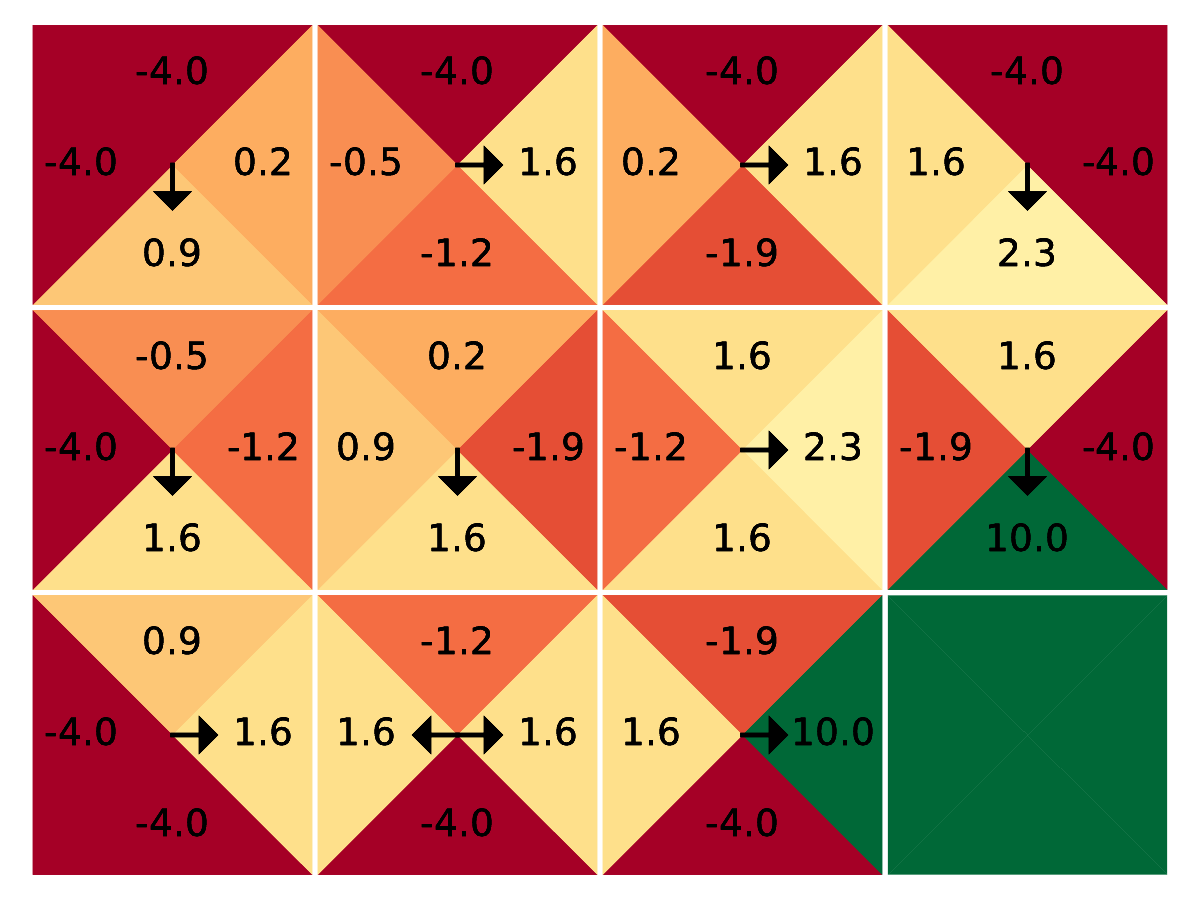}
\caption{${\scriptstyle 0.7} \min + {\scriptstyle 0.3} \max$}
\end{subfigure}%
\begin{subfigure}{0.2\linewidth}
\includegraphics[width=\textwidth]{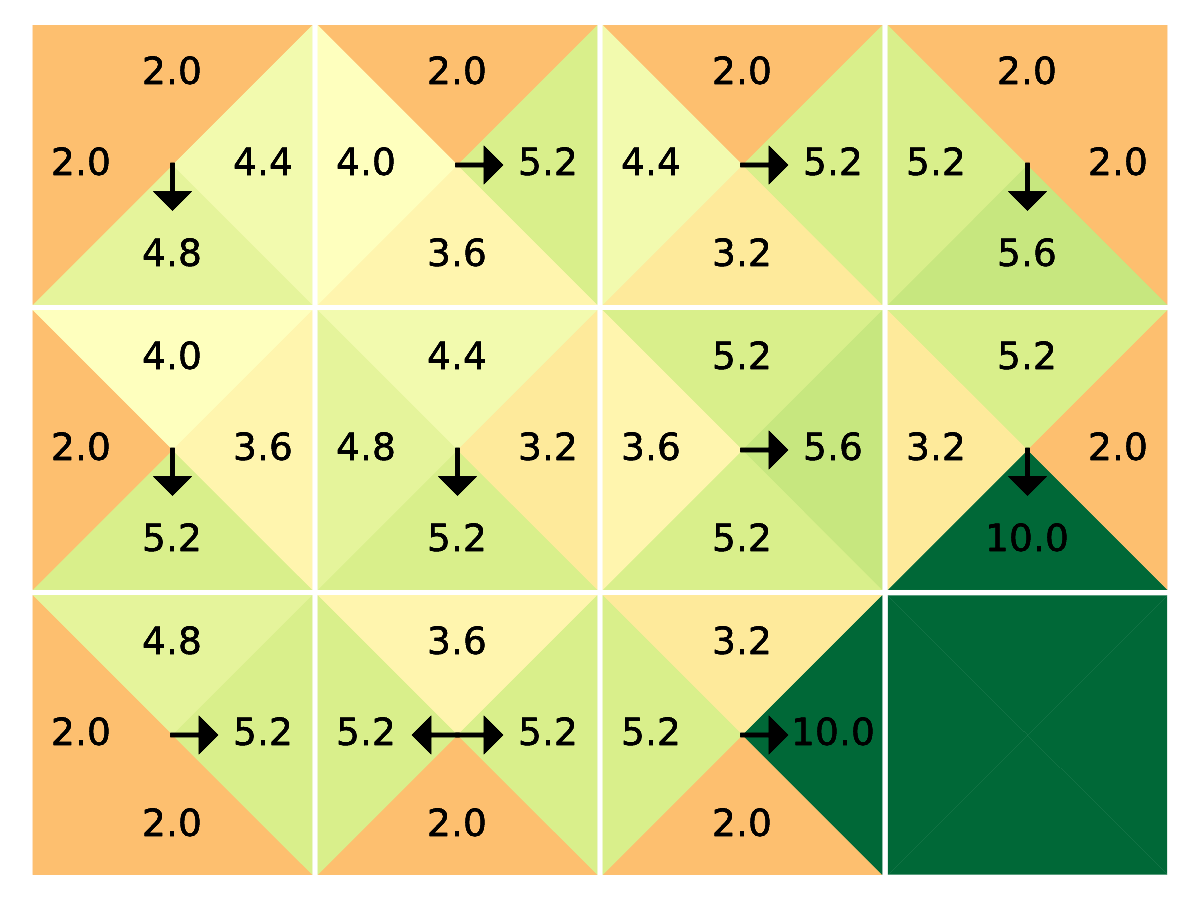}
\caption{${\scriptstyle 0.4} \min + {\scriptstyle 0.6} \max$}
\end{subfigure}%
\begin{subfigure}{0.2\linewidth}
\includegraphics[width=\textwidth]{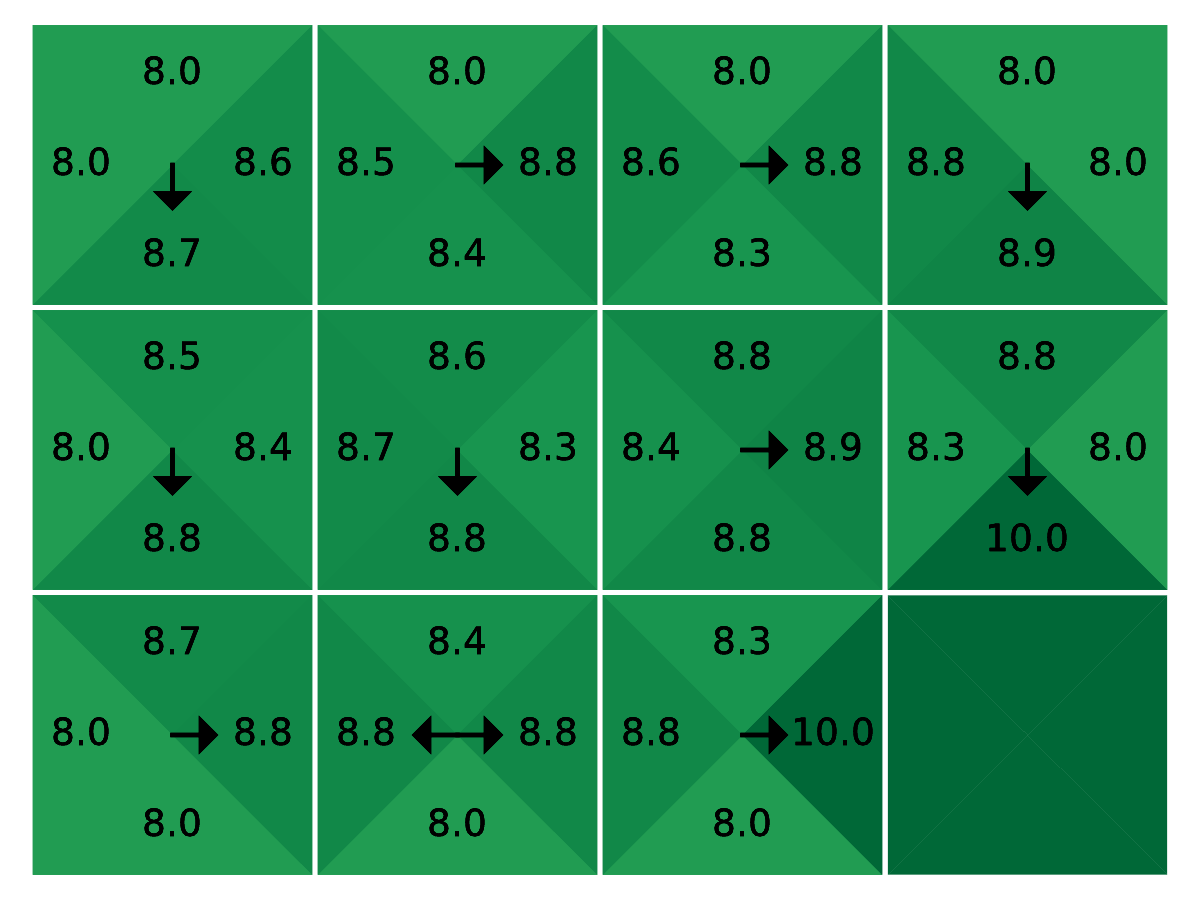}
\caption{${\scriptstyle 0.1} \min + {\scriptstyle 0.9} \max$}
\end{subfigure}%
\begin{subfigure}{0.2\linewidth}
\includegraphics[width=\textwidth]{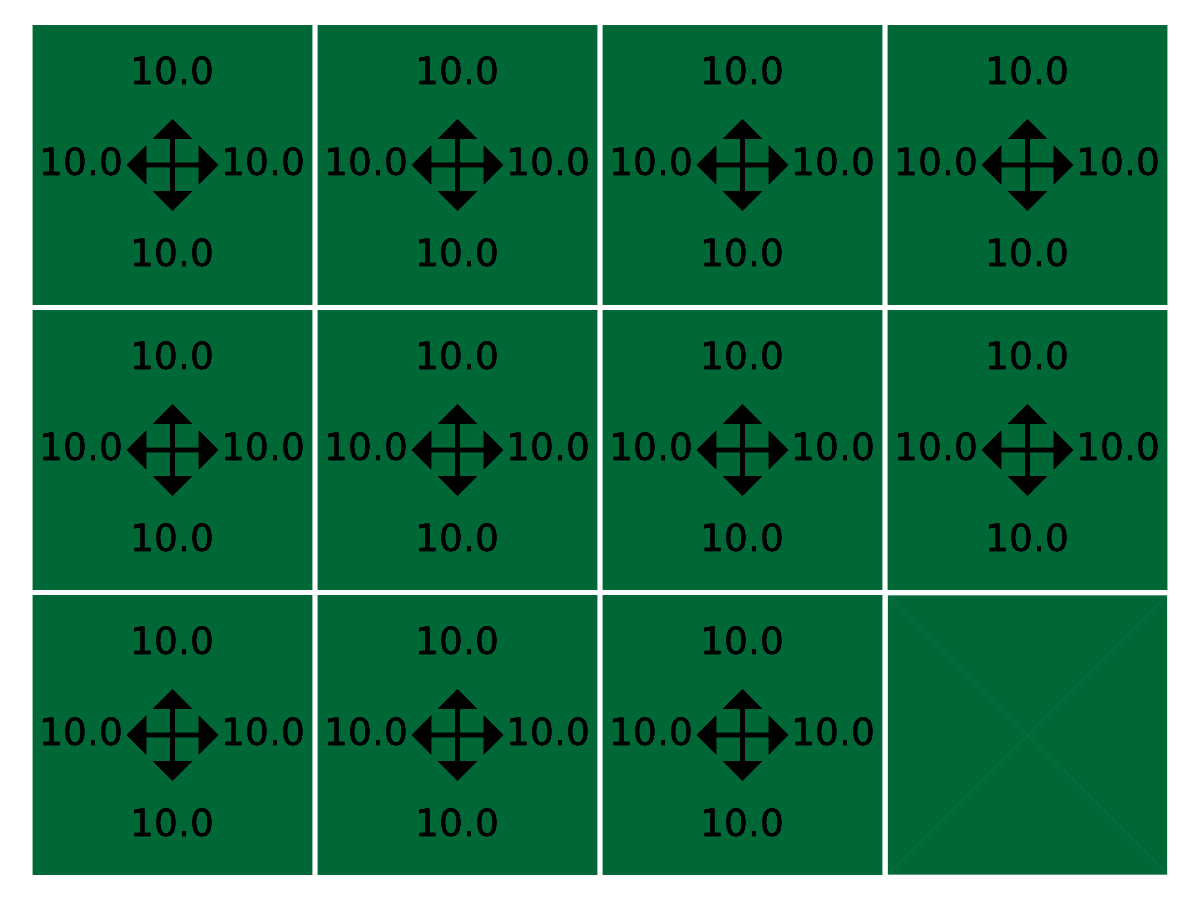}
\caption{$\max$}
\end{subfigure}%
\caption{
${\max} - {\lambda \range} = {\lambda \min} + {(1 - \lambda) \max}$
}
\label{fig:grid_min-max}
\end{figure}

Similarly to \cref{fig:grid}, \cref{fig:grid_min-max} shows the policy preferences for range-regularized max, which is an interpolation between min and max.


\subsection{Wind-world environment}


\paragraph{Implementation}

We implemented the environment and the PPO \citep{schulman2017proximal} algorithm using JAX \citep{jax} and gymnax \citep{gymnax}.

Note that because PPO uses a stochastic policy, our algorithm effectively optimizes the \emph{aggregated expected rewards}, which is different from the \emph{expected aggregated rewards}.
However, we argue that the aggregated expected rewards is still a meaningful objective.
The extension to expected aggregated rewards using distributed RL is left for future work.
See also \cref{sec:stochastic,app:stochastic} for details.


\paragraph{Hyperparameters}

The PPO clipping parameter was set to $0.2$.
We used a critic loss coefficient of $0.5$ and an entropy regularization coefficient of $0.01$.
Agents were trained for a total of $\num{500000}$ time steps using $64$ parallel environments, executed in batch via JAX to enable efficient data collection.


\subsection{Continuous control environments}
\label{ssec:physics_details}

The \textbf{Lunar Lander Continuous} environment, part of the Box2D physics simulation suite \citep{gym}, involves controlling a lunar lander to safely land on a designated landing pad.
The agent has continuous thrust control over the main engine and two side thrusters, which it must use efficiently to achieve a stable landing while minimizing fuel consumption.
The reward function is designed to encourage precise and efficient landings.
The agent receives positive rewards for
(i) moving closer to the landing pad,
(ii) achieving a soft landing, and
(iii) staying upright.
Conversely, penalties are applied for
(i) excessive fuel usage,
(ii) high-impact landings, and
(iii) drifting too far from the target.
The episode terminates if the lander successfully lands within the designated zone, crashes, or drifts out of bounds.
If none of these conditions occur, the episode continues until reaching the time limit.


We used the \textbf{Hopper} environment \citep{erez2012infinite} simulated using MuJoCo \citep{mujoco}, where a 2D one-legged robot must learn to balance and move forward efficiently.
The agent controls three joints (thigh, knee, and foot) to generate locomotion while maintaining stability.
The reward function in Hopper consists of three key components:
(i) \emph{healthy reward}, which incentivizes the agent to remain upright;
(ii) \emph{forward reward}, which encourages the agent to move forward; and
(iii) \emph{control cost}, which penalizes excessive energy use.
Then, the total reward function is given by
\begin{equation}
\text{reward}
=
\text{healthy reward} + \text{forward reward} - \text{control cost}.
\end{equation}
The Hopper environment terminates when the agent is deemed unhealthy or reaches the predefined episode length limit.
The agent is considered unhealthy if its state variables exceed the allowed range, its height falls below a certain threshold, or its torso angle deviates beyond a specified limit, indicating a loss of stability.
If none of these conditions occur, the episode continues until the maximum duration is reached.


We used the \textbf{Ant} environment \citep{schulman2016high} simulated using MuJoCo \citep{mujoco}, where a four-legged quadrupedal robot must learn to efficiently balance and move forward.
The agent controls eight joints (two per leg) to generate stable locomotion while adapting to dynamic interactions with the environment.
The reward function in the Ant environment is designed to encourage forward movement while maintaining stability and efficiency.
It consists of four key components:
(i) a \emph{healthy reward}, which provides a fixed bonus as long as the agent remains upright;
(ii) a \emph{forward reward}, which encourages movement in the positive x-direction;
(iii) a \emph{control cost}, which penalizes excessive actions to promote energy efficiency; and
(iv) a \emph{contact cost}, which discourages large external contact forces.
The total reward is calculated by summing the healthy and forward rewards while subtracting the penalties for control effort and contact forces:
\begin{equation}
\text{reward}
=
\text{healthy reward} + \text{forward reward} - \text{control cost} - \text{contact cost}.
\end{equation}
In some versions of the environment, the contact cost may be excluded from the reward calculation.
The Ant environment terminates when the agent is deemed unhealthy or when the episode reaches its maximum duration of 1000 time steps.
The agent is considered unhealthy if any of its state space values become non-finite or if its torso height falls outside a predefined range, indicating a loss of stability.
If neither of these conditions occur, the episode continues until it reaches the time limit.


\paragraph{Implementation}

We conducted experiments using a modified version of the TD3 \citep{fujimoto2018addressing} implementation from Stable-Baselines3 \citep{sb3}.


\paragraph{Hyperparameters}

Our agent performed $100$ gradient updates per training episode and used a learning rate of $\num{3e-4}$ to ensure stable learning.
Apart from these, our training setup adheres to the default hyperparameters and network architecture of Stable-Baselines3.


\paragraph{Computational resource}

Training a single agent takes approximately 1 hour on an NVIDIA RTX 2080 GPU, with a single CPU core used for environment simulation.


\paragraph{Additional results}

We provide additional results for Hopper and Ant environments.
To comprehensively assess the performance, we present the mean values of various evaluation metrics across four random seeds using radar charts.
Additionally, we visualize the trajectory of the agent in all environments, providing an intuitive representation of how different aggregation functions influence the learned policy.
Animations for all three environments (Lunar Lander Continuous, Ant, and Hopper) are also available at \codelink, offering an intuitive understanding of policy behavior.


\begin{figure}
\centering
\includegraphics[width=\linewidth]{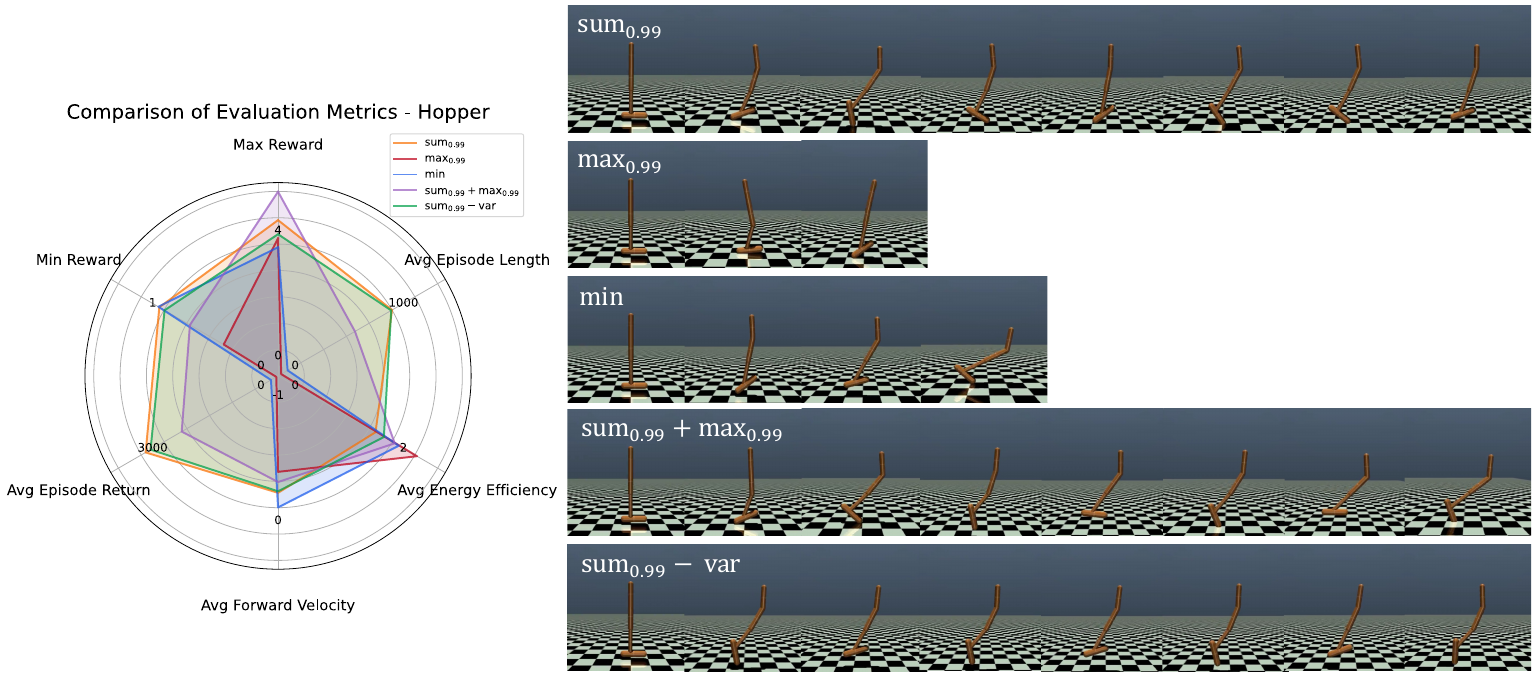}
\caption[Hopper]{
\textbf{Hopper}: Comparison of five reward aggregation methods.
(Left) Radar plot showing performance across six evaluation metrics, averaged over four random seeds.
(Right) Sample trajectories illustrating the qualitative behaviors induced by each aggregation method.
}
\label{fig:hopper}
\end{figure}


\begin{figure}
\centering
\includegraphics[width=\linewidth]{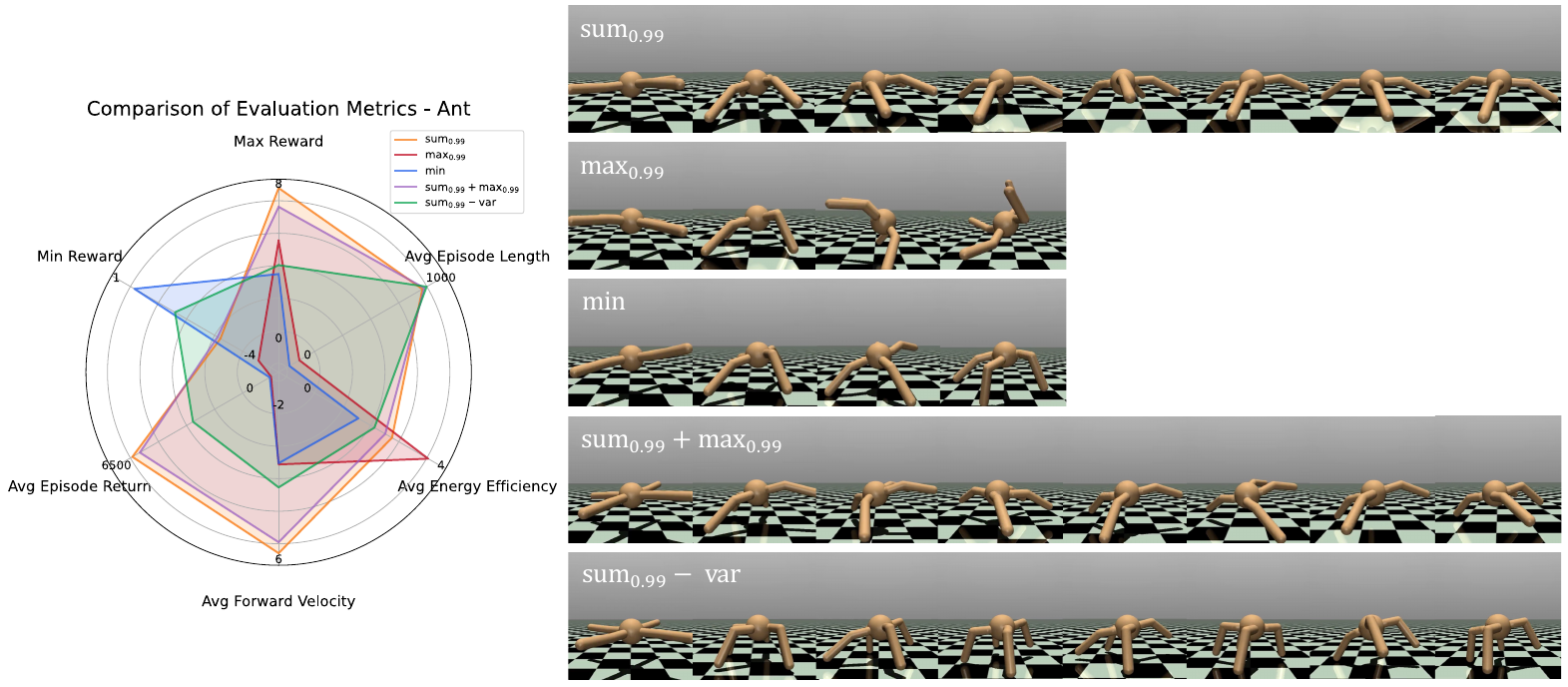}
\caption[Ant]{
\textbf{Ant}: Comparison of five reward aggregation methods.
(Left) Radar plot showing performance across six evaluation metrics, averaged over four random seeds.
(Right) Sample trajectories illustrating the qualitative behaviors induced by each aggregation method.
}
\label{fig:ant}
\end{figure}


For the \textbf{Hopper} environment, we observe distinct behavioral patterns and performance outcomes under different reward aggregation strategies.
The $\fsum_{0.99}$ aggregation, serving as the baseline method, demonstrates strong overall performance across multiple metrics, as reflected in both the radar chart and motion sequences.
In contrast, the $\max_{0.99}$ aggregation focuses solely on optimizing max reward, leading to strong performance in this specific metric but suboptimal outcomes in others.
The corresponding images show the agent taking overly aggressive actions to maximize max reward, which causes it to lose balance quickly as the torso angle exceeds the allowed range.
The $\min$ aggregation, on the other hand, drives the agent to maximize the minimum reward, which typically corresponds to the control cost. 
In an attempt to minimize this cost, the agent refrains from applying control inputs altogether. 
This lack of action causes the agent to fall quickly, as it fails to maintain balance or make corrective movements.
The $\fsum_{0.99} + \max_{0.99}$ aggregation encourages the agent to pursue both high cumulative reward and high per-step reward within an episode. 
This joint objective often leads to more aggressive behaviors, enabling the agent to achieve high peak rewards. 
However, the emphasis on maximizing single-step gains can also induce instability, occasionally causing the agent to fail due to unsafe actions.
While the $\fsum_{0.99} - \var$ aggregation prioritizes stability by minimizing the difference between the maximum and minimum rewards, resulting in more controlled and consistent behavior at the cost of slightly lower rewards.
These results highlight how different reward aggregation strategies shape the behavior of the agent and its learning outcomes.

For the \textbf{Ant} environment, different aggregation strategies lead to varied agent behaviors and trade-offs between stability, performance, and exploration.
The $\fsum_{0.99}$ aggregation, serving as the baseline, achieves balanced performance across multiple metrics, effectively promoting stable and efficient locomotion.
In contrast, the $\max_{0.99}$ aggregation prioritizes obtaining the highest possible reward at an individual time step, leading to highly aggressive movements.
As a result, the agent exhibits excessive speed, which ultimately causes instability and results in the agent losing control and rolling over.
The $\min$ aggregation prioritizes minimizing the risk of low rewards, leading to an overly conservative strategy.
Instead of efficient locomotion, the agent adopts passive or static behavior, often staying close to the ground to avoid unfavorable rewards.
This lack of exploration and controlled movement results in instability, ultimately causing the agent to collapse and terminate early due to height constraints.
Moreover, the $\fsum_{0.99} + \max_{0.99}$ aggregation encourages aggressive behavior by jointly optimizing cumulative and peak rewards.
The agent exhibits rapid, unstable locomotion, often pushing for immediate gains.
While this reduces stability, reward-related metrics remain high, indicating strong overall performance at the cost of greater energy use and inconsistency.
Finally, the $\fsum_{0.99} - \var$ aggregation prioritizes stability by penalizing reward fluctuations, leading to more controlled and steady locomotion.
The agent avoids aggressive actions and achieves longer episode durations.
However, while reducing variance enhances stability, it also limits the ability of agent to explore high-reward strategies, leading to robust but suboptimal overall performance.


\subsection{Portfolio environment}
\label{ssec:portfolio_details}

In our experiment, we trained agents using five different random seeds over a rolling 5-year window, with a total of 10 training periods.
Specifically, for each training period, training begins on January 1 of a given year and continues for five years, ending on December 31 of the fifth year.
Each training period starts one year after the previous one, resulting in overlapping but not identical training datasets.
Following the training phase, we evaluate the performance of agents in the subsequent year, immediately following the training period.
Finally, we assess their generalization performance in the test phase, which takes place in the year after the evaluation period.
This design allows us to systematically analyze the agents' performance across different temporal contexts while leveraging historical data in a structured and overlapping manner.


\paragraph{Implementation}

We conducted experiments using a modified version of the PPO \citep{schulman2017proximal} implementation from Stable-Baselines3 \citep{sb3}.


\paragraph{Hyperparameters}

We trained each agent for a total of $\num{7500000}$ time steps.
Compared to the default settings of PPO in Stable-Baseline3, we made several modifications to better suit our environment.
The learning rate followed a linear decay schedule, starting from $\num{3e-4}$ and gradually decreasing to $\num{1e-5}$ over the course of training.
We set the discount factor to $0.9$ and the GAE lambda to $0.9$ to reduce reliance on long-term returns, and used a slightly wider clipping range $0.25$ to allow for greater policy updates.
These adjustments were empirically tuned for improved stability and performance in our setting.


\paragraph{Computational resource}

Training a single agent takes approximately 1.5 hours on an NVIDIA RTX 2080 GPU, with the environment running in parallel on 10 CPU cores to accelerate data collection.


\paragraph{Results: Sharpe ratio comparison over time}

\begin{figure}
\centering
\includegraphics[width=0.8\linewidth]{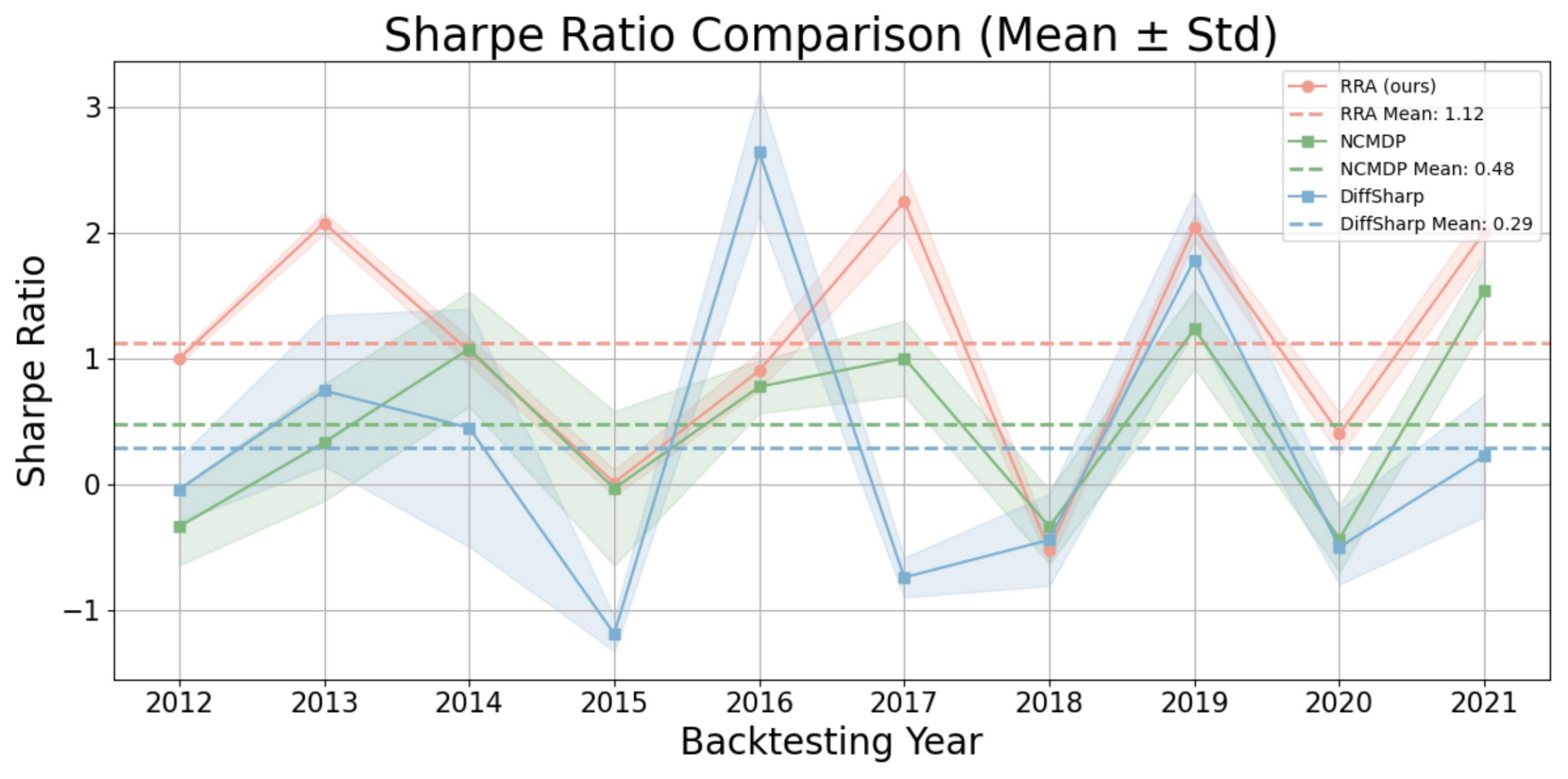}
\caption[Sharpe ratio]{%
A year-by-year comparison of Sharpe ratios obtained by different methods during the test phase across a rolling backtesting window from 2012 to 2021.
Each data point represents the mean performance across five different random seeds, with the shaded regions indicating one standard deviation to reflect variability.
The horizontal dashed lines represent the mean Sharpe ratio across all years for each method, providing a summary view of their long-term performance.
}
\label{fig:portfolio}
\end{figure}

Figure~\ref{fig:portfolio} presents a comparison of the Sharpe ratios achieved by three methods during the test phase over the ten-year backtesting period.
Each backtesting period refers to a historical test year following the training and validation phases, during which the strategy is evaluated on previously unseen data to assess its out-of-sample performance \citep{bailey2015probability}.
The red curve represents our method, which not only achieves the highest mean Sharpe ratio with relatively low variability, but also consistently delivers the best or highly competitive performance in $8$ out of the $10$ backtesting periods.
This consistency across random seeds and temporal splits underscores the practical generalization ability of our proposed method and its suitability for real-world financial decision-making.


\clearpage
\section{Discussion}
\label{app:discussion}
\paragraph{Reward function design vs.~aggregation strategies}

Changing the reward function and adjusting how rewards are aggregated are two complementary approaches to shaping agent behavior.
Rather than asserting the superiority of one approach over the other, we examine the trade-offs and situational advantages associated with each.

Reward function modification directly encodes task objectives into the per-step feedback signal received by the agent.
This approach is expressive and flexible, allowing designers to incorporate domain-specific preferences \citep{ng1999policy, hadfield2017off}, intermediate goals \citep{andrychowicz2017hindsight}, or constraints \citep{achiam2017constrained}.
However, designing an effective reward function often requires careful tuning, may introduce unintended incentives, and can suffer from reward misspecification, especially in environments with sparse or delayed feedback \citep{ng1999policy, ziebart2008maximum, hadfield2017off}.

In contrast to modifying the reward function itself, reward aggregation modification keeps the underlying reward signal fixed and instead alters how rewards are aggregated over time to define the training objective \citep{wang2020planning, cui2023reinforcement, veviurko2024max}.
This offers a structured way to influence long-term behavior without redefining the reward signal at each time step.
For instance, the $\max$ aggregation \citep{quah2006maximum, veviurko2024max} focuses on the highest reward in a trajectory, encouraging strategies that pursue the most valuable or high-potential actions, while the $\min$ aggregation \citep{cui2023reinforcement} emphasizes avoiding the worst-case outcomes, encouraging risk-averse strategies.
This approach is effective when the reward signal provides informative feedback, but the desired policy depends on how that feedback is interpreted over time.
However, limited or ambiguous reward signals may restrict the ability of any aggregation function to align with the intended behavioral goals.

In practice, modifying the reward and adjusting the aggregation function are not mutually exclusive and can be combined effectively.
The reward function provides the essential feedback for learning, while the aggregation method influences how this feedback is evaluated over time.
The choice of whether to modify one, the other, or both should be guided by the nature of the task, the clarity and expressiveness of the reward, and the behavioral patterns desired in the learned policy.


\paragraph{Limitations of sum-based objectives}

While standard RL typically defines the training objective as the sum of per-step rewards, this formulation tends to be effective under certain assumptions about the reward signal and task structure \citep{silver2021reward}.
First, it is generally better suited to tasks where the overall performance can be approximated by accumulating the reward at each time step.
In such cases, the total return should reflect meaningful progress over time.
Second, sum aggregation assumes that the timing of rewards is not a critical factor.
While discounted sums introduce a preference for earlier rewards, they still impose a fixed temporal structure.
Therefore, sum is suited to tasks where the timing of rewards is relatively neutral and consistent accumulation matters more than when specific rewards happen.
Finally, sum-based objectives are more likely to be effective when the reward function offers sufficient granularity, providing reliable feedback at each step to support training.

Despite its simplicity and widespread adoption, summing per-step rewards may be less effective in scenarios where its underlying assumptions are not fully satisfied.
In particular, many tasks do not neatly align with the structure implied by a standard sum-based objective.
As a result, the learned policy may be optimal under the sum semantics but misaligned with the intended behavioral goals.
For example, in safety-critical environments, aggregating rewards via summation might obscure low-reward outliers, as occasional high rewards could mask dangerous behaviors.
Similarly, in peak-oriented tasks where success depends on achieving exceptional performance at a specific moment, summation may diminish the significance of these peak events by averaging them with less important steps.
In such contexts, adjusting how rewards are aggregated over time may offer additional flexibility for aligning the learning objective with designer intent.

\end{document}